\documentclass {article} 
\usepackage{arxiv}
\usepackage{booktabs} 
\usepackage{algorithm}%
\usepackage{algpseudocode}%
\usepackage{hyperref}
\usepackage{amsmath}
\usepackage{xcolor}
\usepackage{xspace}
\usepackage{soul}
\newcommand{\ignore}[1]{}
\usepackage[textsize=tiny]{todonotes}
\usepackage{graphicx}
\usepackage{boldline}
\usepackage[export]{adjustbox}
\usepackage[caption=false]{subfig}
\usepackage[font=small]{caption}
\usepackage{adjustbox}
\usepackage{enumerate}
\usepackage{amsfonts}
\usepackage{colortbl}
\usepackage{float}
\usepackage{rotating}
\usepackage{setspace}
\title{A Hybrid Cooperative Co-evolution Algorithm Framework for Optimising Power Take Off and Placements of Wave Energy Converters}

\author{
  Mehdi Neshat \\
  Optimization and Logistics Group\\
  School of Computer Science\\
  The University of Adelaide\\
   Australia \\
  \texttt{mehdi.neshat@adelaide.edu.au} \\
   \And
 Bradley Alexander \\
  Optimization and Logistics Group\\
  School of Computer Science\\
  The University of Adelaide\\
   Australia \\
  \texttt{bradley.alexander@adelaide.edu.au} \\
  \And
 Markus Wagner \\
  Optimization and Logistics Group\\
  School of Computer Science\\
  The University of Adelaide\\
   Australia \\
  \texttt{markus.wagner@adelaide.edu.au} \\
  }

\begin{document}

\maketitle
\doublespacing
\begin{abstract}
Wave energy technologies have the potential 
to play a significant role in the supply of renewable 
energy on a world scale. 
One of the most promising designs for wave energy converters (WECs) are fully submerged buoys. In this 
work we explore the optimisation of WEC arrays 
consisting of a three-tether buoy model called CETO. 
Such arrays can be optimised for total energy 
output by adjusting both the relative
positions of buoys in farms and also the power-take-off (PTO) parameters for each buoy. The search space
for these parameters is complex and multi-modal. Moreover,
the evaluation of each parameter setting is computationally expensive -- limiting the number of full model evaluations that can be made. 
To handle this problem, we propose a new hybrid cooperative co-evolution algorithm (HCCA). HCCA consists of a symmetric local search plus Nelder-Mead and a cooperative co-evolution algorithm (CC) with a backtracking strategy for optimising the positions and PTO settings of WECs, respectively.
Moreover, a new adaptive scenario is proposed for tuning grey wolf optimiser (AGWO) hyper-parameter.  AGWO participates notably with other applied optimisers in HCCA.  For assessing the effectiveness of the proposed approach five popular Evolutionary Algorithms (EAs), four alternating optimisation methods and two modern hybrid ideas (LS-NM and SLS-NM-B) are carefully compared in four real wave situations (Adelaide, Tasmania, Sydney and Perth) with two wave farm sizes (4 and 16). According to the experimental outcomes, the hybrid cooperative framework exhibits better performance in terms of both runtime and quality of obtained solutions. 

\end{abstract}


\sloppy

\section{Introduction}
Renewable energy technologies make up an increasing proportion of new-build electricity generating worldwide~\cite{liebreich2013bloomberg}. 
Ocean wave energy is one very promising technology for contributing to growth in energy demand from renewable sources due to the high-energy densities of ocean environments, and high capacity factors wave energy converter (WEC) models~\cite{drew2009review}. 
It is envisaged that ocean wave energy could supply more than $70\%$ of the world’s whole energy demand \cite{yaobao2013principle}; but the current WECs technologies are not fully developed due to the technical engineering challenges of harnessing ocean wave power in harsh ocean environments.  

In this research, we apply a developed WEC simulator for evaluating the absorbed power of a wave farm consisting of CETO-6 model WEC converters~\cite{mann2011application}. 
CETO converters are spherical submerged three-tether buoys. These converters were first designed in 2007 by the Carnegie Clean Energy company \cite{mann2007ceto}.  
The energy output of WEC's of this design is dependent on a number of factors including, the relative positions of WEC's in an array; the power-take-off (PTO) settings on each buoy's tethers; and the long-term sea conditions of the wave-farm site. In this work, we optimise both the position and PTO parameters of simulated wave farms consisting of both 4 and 16 buoys in 4 real wave environments. 

Since computing the complex hydrodynamic interactions among converters is computationally costly, the evaluation of each WECs arrangement can take several minutes. Moreover, a combination of both search spaces WECs positions and PTOs creates a multimodal and large-scale optimisation problem. These challenges require the use of robust, low-cost global search heuristics customised to this problem domain.  
To date the best performing heuristics for this problem~\cite{Neshat:2019:HEA:3321707.3321806} have been hybrid optimisation methods that placed and refined buoys parameters one at a time. 

In this paper, we propose a new hybrid cooperative co-evolution algorithm (HCCA) for optimizing WEC array
positions and PTO parameters that builds on these previous approaches by:
The main contributions of the HCCA are:
\begin{enumerate}

\item Developing the applied simulator for evaluating the PTOs configurations per each wave frequency, which is more close to the real sea states. 

\item Extending the Grey Wolf Optimiser by a new adaptive mechanism (AGWO) for balancing exploration and exploitation. 

\item Embedding the AGWO within a Cooperative Co-evolution method (SLPSO + SaNSDE) for optimizing the PTO configuration of the wave farm. 
\item Employing a backtracking strategy to further optimise both position and PTO settings of the obtained array. 
\end{enumerate}

To evaluate this new algorithm, we compare HCCA to a comprehensive range of meta-heuristics for optimizing the total power output of a wave farm, including (1) five popular off-the-shelf EAs, (2) four cooperative optimisation ideas, and (3) three hybrid optimisation algorithms. We evaluate our using four real wave scenarios from the Southern coast of Australia (Perth, Sydney, Adelaide and Tasmania).
Each scenario embeds a detailed model of a wave environment including time-integrated distributions of wave-heights, periods and directions. The experimental results show that HCCA is able to 
significantly outperform other optimisation approaches with regard to both convergence speed and total absorbed power output.

The remainder of the paper is organized as follows. We provide an overview of related work in Section~\ref{sec:relatedwork}, and introduce the mathematical model for the WECs being studied in Section~\ref{sec:model}. The optimisation setup and the proposed optimisation methods are described in Sections~\ref{sec:opt} and~\ref{sec:method}, respectively. In Section~\ref{sec:experiments}, the experimental results are presented. We conclude with a summary and outline potential future work. 

\section{Related Work} \label{sec:relatedwork}

There have been a number of studies in optimising the power output a variety of WEC models. 
One initial study \cite{child2010optimal} optimised WEC positions for five buoys using both the Parabolic Intersection (PI) method and a GA. The study required a high number of function evaluations (37000). The wave environment modelled was highly simplified, with just one wave direction. 
A recent study by Ruiz et al.~\cite{ruiz2017layout} employed another simple wave scenario to compare a custom GA, CMA-ES \cite{hansen2006cma} and glow-worm optimisation (GSO) \cite{krishnanand2009glowworm} for optimizing the position of the buoys in a discrete grid. The investigation found that while the convergence rate of CMA-ES is faster than of the other two methods, it could not overcome both the GA and GSO, in terms of total power production. 
In other recent WEC position optimisation research, Wu et al.~\cite{wu2016fast} compared a 1+1EA and population-based evolutionary algorithm (CMA-ES) to optimise both 25 and 50 buoys using a simplified uni-directional irregular wave model. That paper revealed that the 1+1EA with a simple mutation operator is able to outperform CMA-ES. However, the performance achieved for both the 25 and 50-buoy layouts was low. 
Neshat et al.~\cite{neshat2018detailed,DBLP:journals/corr/abs-1904-09599,neshat2019adaptive} characterized a more complicated wave scenario (seven wave directions and 50 wave frequencies) with the intra-buoy effects and employed this knowledge to make a customized, single-objective hybrid heuristic (local search + Nelder-Mead). However, the wave model still used an artificial wave scenario and the proposed method did not tune PTO parameters. 

Another challenging aspect of maximizing the total power output of the wave farm is controlling the WECs' oscillations with respect to the incoming waves' frequency. This is because maximum efficiency will be achieved at resonance. However, maintaining a resonant condition can be challenging in real sea states with multiple different frequencies~\cite{falnes2002ocean}.

One way of achieving resonance is by configuring the power take-off (PTO) system of the WECs, either in online or offline settings.
For then online setting, Ding et al.~\cite {ding2016sea} implemented the maximum power point tracking (MPPT) control method for optimising the damping rate (dPTO) of one converter (CETO 6). The MPPT is a type of online-optimisation based on the gradient-ascent algorithm. The outcomes reported a high efficiency of the MPPT damping controller compared with a fixed-damping system, but the performance of MPPT was not assessed for layouts with more than one buoy.  
In later work Abdelkhalik et al. \cite{abdelkhalik2018optimization} utilised the hidden genes genetic algorithm (HGGA) to tune PTO parameters. 
While the proposed optimiser boosted the total energy produced, HGGA's efficiency was not compared to other modern EAs. 

Glass et al.~\cite{giassi2018layout} used a combination of a generic GA with an analytical multiple scattering method to optimise WECs parameters including buoy radius, draft, and converter damping. That work produced some 5 and 9-buoy layouts with constructive interactions in a simple (uni-directional) wave scenario. 
Silva et al. \cite{e2016hydrodynamic} compared the performance of a GA and COBYLA ~\cite{powell1994direct} (Constrained optimisation BY Linear Approximations) for maximizing the produced annual electrical output of one WEC with a U-shaped design inside an oscillating water column (UGEN) by adjusting PTO settings and the buoy's geometrical characteristics. The obtained results showed that COBYLA method converged to design with fewer evaluations. However, the GA produced a better solution overall, because COBYLA had converged to a local optimum.
In another recent study, hybridization of a customized local search with a Nelder-Mead algorithm and a refinement strategy (SLS-NM-B) was introduced~\cite{Neshat:2019:HEA:3321707.3321806} for optimizing both arrangement and PTOs parameters of WECs model (CETO). While the optimisation results represented a considerable power improvement of SLS-NM-B compared with other popular EAs, the optimisation of the PTOs settings were optimised in unison.

\section{Mathematical modelling for wave energy converters} \label{sec:model}

\subsection{Wave Resource}
\begin{figure}[t]
\subfloat[]{
\includegraphics[clip,width=0.49\columnwidth]{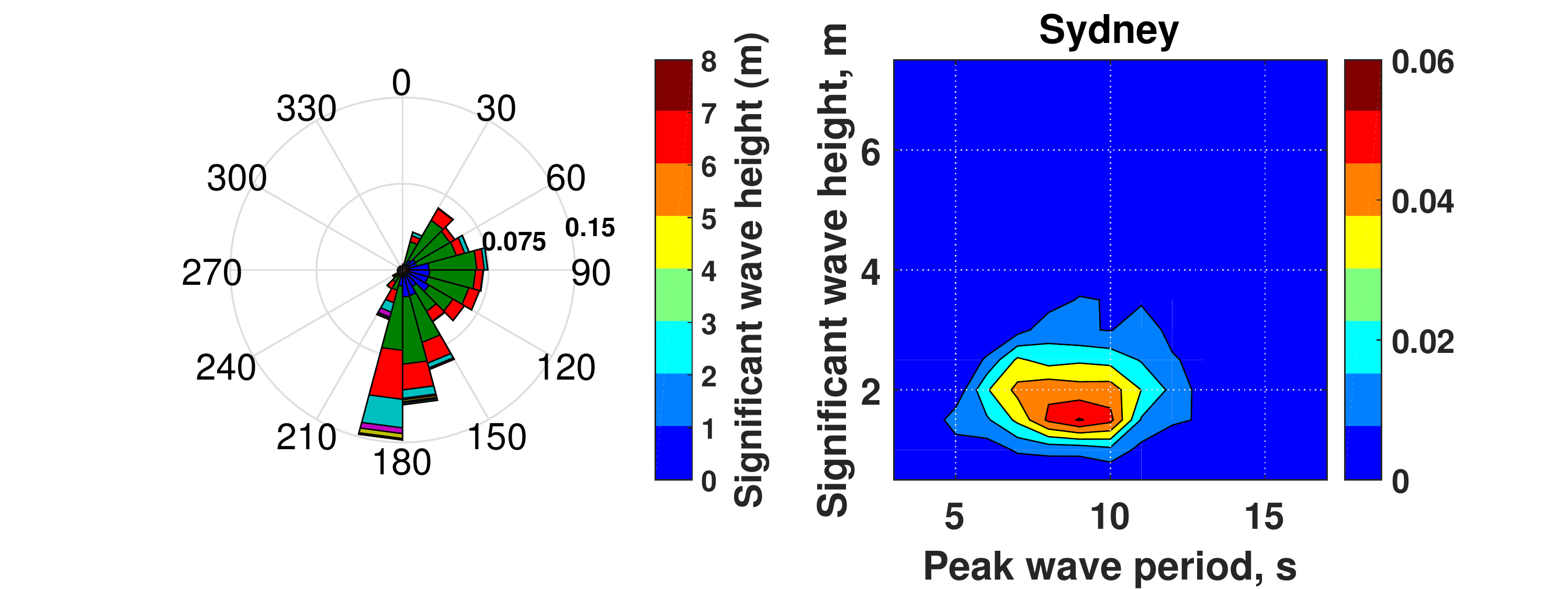}}
\subfloat[]{
\includegraphics[clip,width=0.49\columnwidth]{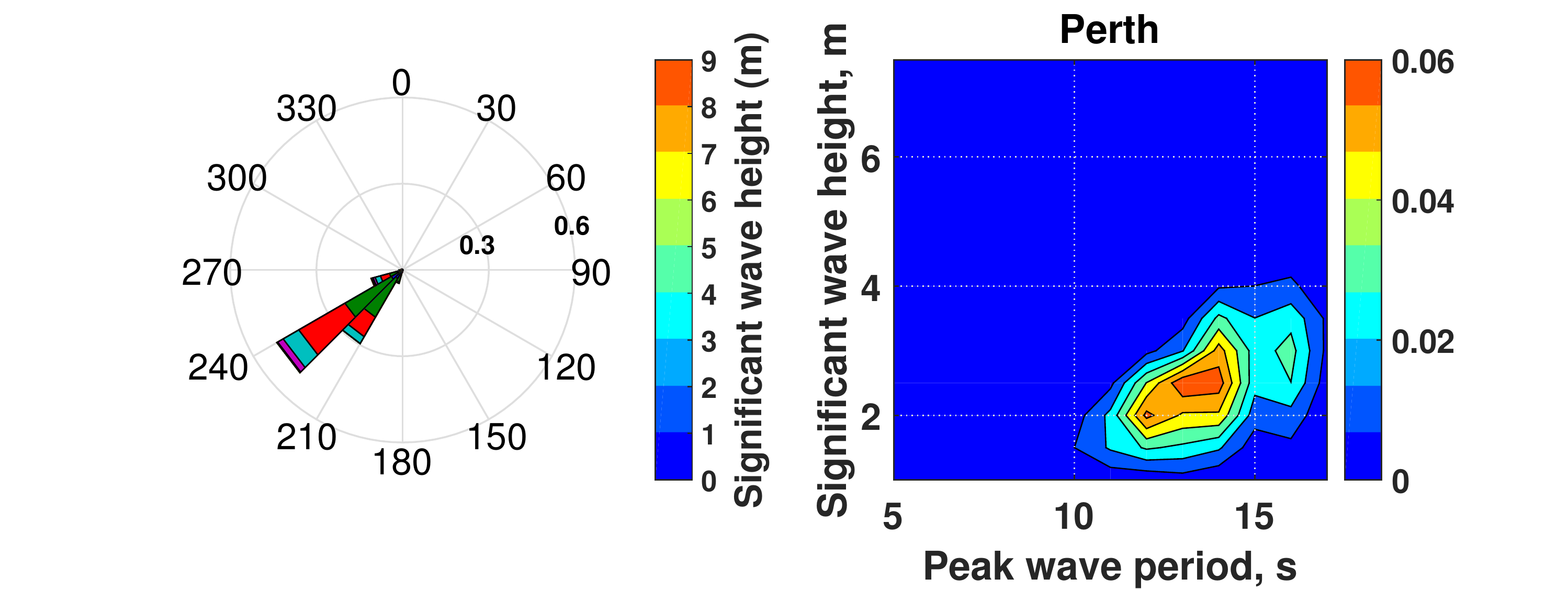}}\\
\subfloat[]{
\includegraphics[clip,width=0.49\columnwidth]{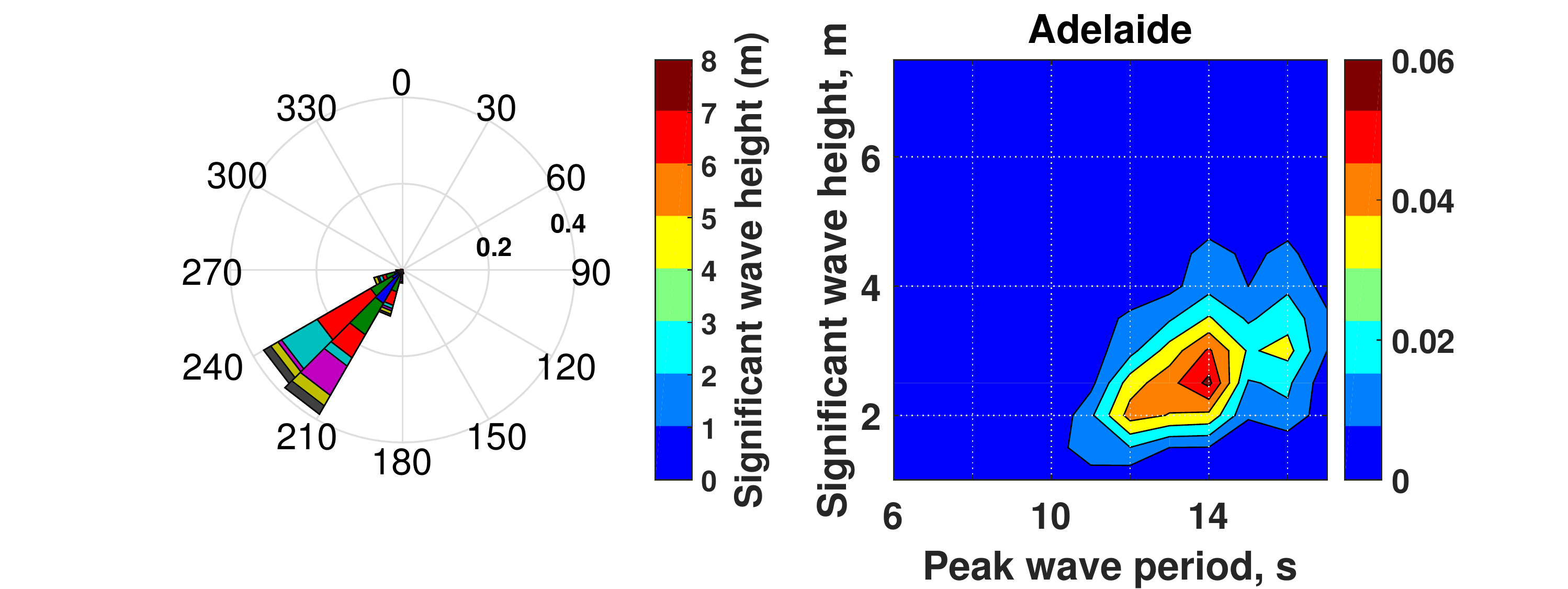}}
\subfloat[]{
\includegraphics[clip,width=0.49\columnwidth]{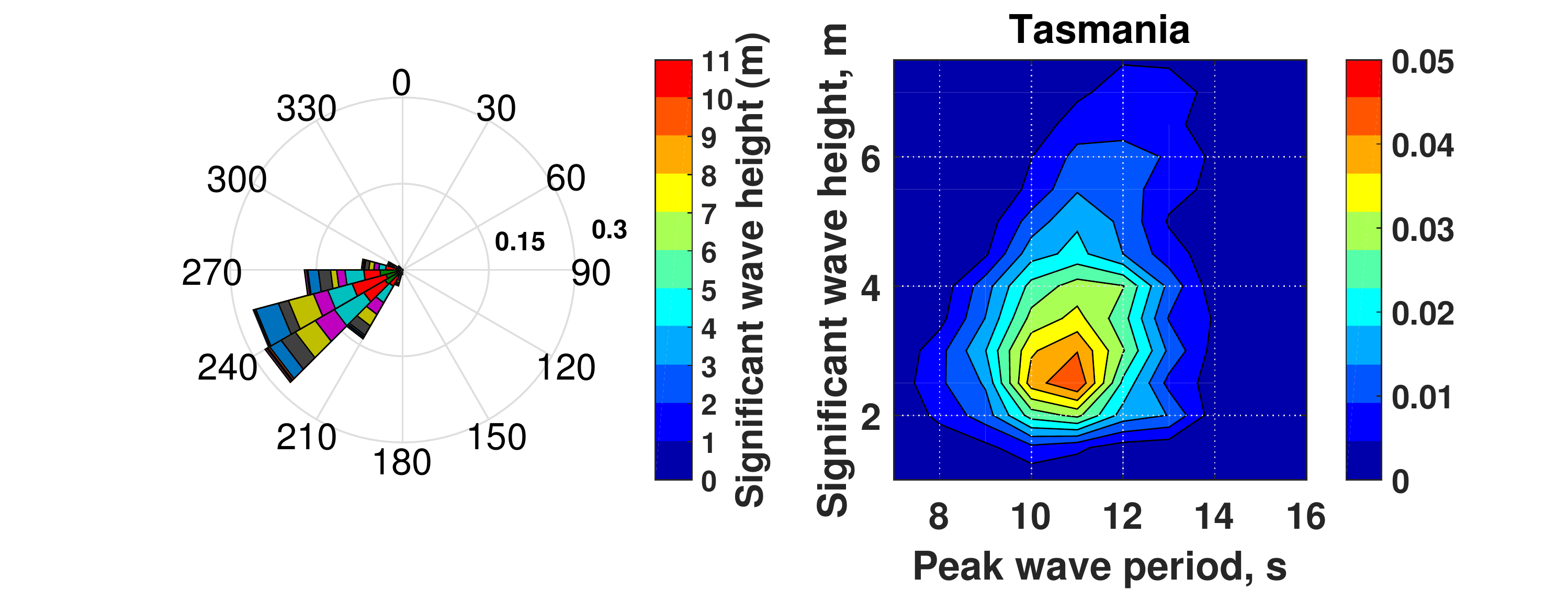}}
\caption{Scatter and wave rose diagrams for four wave energy sites in Australia: (a)~Sydney, (b)~Perth, (c)~Adelaide  and (d)~Tasmania. These are: the directional wave rose (left) and wave scatter diagram (right).}%
\label{fig:wave_direction}%
\end{figure}

According to the latest real wave data set from Australian Wave Energy Atlas \cite{CSIRO2016}, different four wave sites on the southern coast of Australia are studied in this paper including  Perth, Adelaide, Tasmania (southwest coast) and Sydney. Figure~\ref{fig:wave_direction} shows the directional wave rose and wave scatter diagram of these four wave scenarios. It is observed that the applied wave regimes differ with regard to the directional distributions and cumulative energy. This diversity provides four various search spaces for evaluating the performance of the optimisation methods accurately.  The applied model of the ocean wave is irregular directional waves embedding the Bretschneider spectrum \cite{bretschneider1952generation}. 

\subsection{Power Absorption Modelling }
The applied WEC in this research is a symmetric, spherical and fully submerged buoy (minimum depth of 8m at the top of the buoy)  with three-tethers. Each tether is attached to a mooring system installed on the seabed. The principal goal of the mooring system is to keep the floating structure in  position within specific tolerances in both under normal load conditions and severe storm load conditions. The assumed optimal angle of each tether is 55 degrees ~\cite{scruggs2013optimal}. The details of physical WECs characteristics, including submergence and ocean depth can be found in ~\cite{neshat2018detailed}.

For modelling the motion of a WEC in a frequency domain, three degrees (surge, sway and heave) of freedom are formulated based on linear wave theory. The applied hypotheses are the following: 
\begin{align}\label{eq:model}
\begin{split}
\hat{F}_{exc,\Sigma}&=\left(\left(M_{\Sigma}+A_{\sigma}(\omega)\right)j\omega+B_{\sigma}(\omega)-\frac{K_{pto,\Sigma}}{\omega}j+D_{pto,\Sigma}\right)\ddot{X}_{\Sigma}\\
{M} &= m{I}_{3N}\\
   {K}_{pto} &= K_{pto}{I}_{3N}\\
   {D}_{pto} &= D_{pto}{I}_{3N}
\end{split}
\end{align}
where ${F}_{exc}$ is the frequency-dependent vector of excitation forces and $M$ is known as a mass matrix (${I}_{3N}$ is the identity matrix with size $3N$). $N$ is the number of WECs in the layout, and the constant $3$ denotes the number of degrees of freedom. $\ddot{X}_{\Sigma}$ is a vector of body acceleration in the surge, heave and sway directions. The matrices ${A}$ and ${B}$ define the hydrodynamic added-mass and radiation damping coefficients, respectively. The PTO mechanism is modelled on an oscillating spring. The ${K}_{pto}$ and ${D}_{pto}$ are, respectively,  the stiffness of spring and damping PTO matrices. Each row of these matrices represents the settings for each buoy. For each buoy, there are $50$ individual ${K}_{pto}$ and ${D}_{pto}$ parameter settings representing a tuned response to $50$ different ocean wave frequencies. For modelling the hydrodynamic interaction between submerged buoys, a semi-analytical solution is given in \cite{wu1995radiation}.

For calculating the power output of an entire WEC
array, Equation \ref{total-power} computes  the mean all WEC's power harnessed in a regular wave frequency environment:
$\omega$, amplitude, and wave angle $\beta$.
\begin{equation}\label{total-power}
P_{\Sigma}= \frac{1}{4}\left(\hat{F^*}_{exc,\Sigma}\ddot{X}_{\Sigma}+\ddot{X^*}_{\Sigma}\hat{F}_{exc,\Sigma}\right)-\frac{1}{2}\ddot{X^*}_{\Sigma}B\ddot{X^*}_{\Sigma}
\end{equation}
While we are able to calculate the total power of the wave farm in Equation \ref{total-power}, it is very computationally expensive, and the computational cost rises quadratically with the number of buoys. Note that, where there is constructive interaction between converters, the total power output can grow super--linearly with the number of buoys.

\section{Optimisation Setup}\label{sec:opt}
The formulation of the optimisation problem to 
maximise the power output of a WEC array is:
\[
  P_{\Sigma}^* = \mbox{\em argmax}_{\mathit{X,Y,K_{pto},D_{pto}}} P_{\Sigma}(\mathit{X,Y,K_{pto},D_{pto}})
\]
\noindent where $P_{\Sigma}(\mathit{X,Y,K_{pto},D_{pto}})$ shows the annual average power produced for given WEC locations and PTO settings in a 2-D coordinate system at $x$-positions: $\mathit{X}=[x_1,\ldots,x_N]$, $y$-positions: $\mathit{Y}=[y_1,\ldots,y_N]$ and Power Take-off configurations including $\mathit{K_{pto}}=\{[B_{k_1}^1,\ldots,B_{k_{50}}^1],\ldots,[B_{k_1}^N,\ldots,B_{k_{50}}^N]\}$ and $\mathit{D_{pto}}=\{[B_{d_1}^1,\ldots,B_{d_{50}}^1],\ldots,[B_{d_1}^N,\ldots,B_{d_{50}}^N]\}$ . where $B$ is the $i$th buoy here and $N \in \{4,16\}$. it is assumed that all WECs are placed the same depth (5 metres) in ocean with the uniform depth of $30$ metres. 

\paragraph{Constraints}
In this work, there are three  constraints, including constraints of farm boundaries, safe distance constraints between generators and constraints on PTOs variables.  In terms of farm boundaries the positions of each buoy $(x_i,y_i)$ in the wave farm is  restricted to a square search space $S=[x_l,x_u]\times[y_l,y_u]$: where $x_l=y_l=0 ~and~x_u=y_u=\sqrt{N * 20000}\,m^2$. The minimum safety distance constraint, to allow for shipping egress,  is set to 50 meters. The PTO constraints are on spring damping PTO coefficients of $d_l=5\times10^4, d_u=4\times10^5$ and $k_l=1, k_u=5.5\times10^5$. Where a candidate solution satisfies  all the constraint functions, it is marked as a feasible layout. For handing both boundary constraints (position and PTOs),  infeasible solutions are forced to the most adjacent feasible design.  For the safety distance constraint, a steep penalty function is used:
\[
\begin{array}{ll}
\mbox{\em{Sum}}_{\mbox{\em dist}}  = & \sum_{i=1}^{N-1}\sum_{j=i+1}^{N} 
(\mbox{\em{dist}}((x_i,y_i),(x_j,y_j))-50), \\
&\mbox{if } \mbox{\em{dist}}((x_i,y_i),(x_j,y_j))<50 \mbox{else 0}
\end{array}
\]
where $sum_{dist}$ is the sum of violations of the safe distance between buoys. The Euclidean distance between both buoys $i$th and $j$th is denoted by $dist((x_i,y_i),(x_j,y_j))$. The penalty value to the total power absorbed of the wave farm is calculated by $(\mbox{\em{Sum}}_{\mbox{\em{dist}}}+1)^{20}$. The penalty strongly encourages selecting feasible layouts during the optimisation process.
 
\paragraph{Computational Resources}
The optimisation approaches studies here are evaluated and compared for both  4 and 16 WEC arrays for four real wave scenarios. For comparing all proposed methods in a realistic design setting, a time budget criterion of three days is set for optimisation method trial on an HPC supercomputer with a 2.4GHz Intel 6148 processor running 12 processes in parallel with 128GB of RAM. On this platform, this mode of parallelisation usually accommodates more than ten times speedup. Note that the implementations of the proposed optimisation methods are written to so as to exploit the parallel processing capabilities of the platform maximally. 
The software platform used for running the function evaluations and the optimisation algorithms is MATLAB R2018.


\section{Optimisation Methods}\label{sec:method}
In this research three different broad optimisation strategies are employed for maximizing the total absorbed power output of 4 and 16-buoy layouts in this research. The first approach applies search algorithms to all decision variables simultaneously. 
These variables include all the $x$ and $y$ buoy position and all of the PTO parameters. For $16$ buoys this approach requires that over $1632$ variables are optimised all at once.
The second broad approach is to apply 
cooperative methods~\cite{bezdek2003convergence}, which alternate between the optimisation of position and PTO parameters.  
The third strategy, used in \cite{neshat2018detailed, Neshat:2019:HEA:3321707.3321806},  places buoys sequentially (one-at-a-time). Under this strategy, the performance of three hybrid methods are evaluated and compared: $LS-NM$~\cite{neshat2018detailed}, $SLS-NM-B$~\cite{Neshat:2019:HEA:3321707.3321806}, and a new hybrid cooperative EA (HCCA). The details of the algorithms evaluated for each strategy are summarised in Table \ref{details:OP}.  
\begin{table*} 
\caption{Review of the proposed framework methods employed in this paper. All approaches are restricted to the same computational budget constraint. parallelism can be categorised into two groups as per-individual or per-frequency according to the individuals number in the population.
 }
\label{details:OP}
\scalebox{0.9}{
\begin{tabular}{|p{3.6cm}|p{2.2cm}|p{11cm}|}
\hlineB{4} 
\textbf{Abbreviation} & \textbf{parallelizm} &  \multicolumn{1}{|c|}{\textbf{Description}} \\ \hline\hline
\hlineB{4} 
\multicolumn{3}{ |c| }{\textbf{All-at-once methods (Positions+PTO parameters)}}\\
\hlineB{4}

\textbf{$CMA-ES$ }& per-individual  & CMA-ES~\cite{hansen2006cma} all dimensions, $\mu=4 + int(3 * log(N_{var}))$ , $\sigma=0.3*Area$ \\ \hline

\textbf{$DE$} & per-individual & Differential evolution~\cite{storn1997differential}, $\mu=50$, $F=0.5$, $\mbox{$P_{cr}=0.5$} $ \\ \hline

\textbf{$PSO$}   &   per-individual  & Particle Swarm optimisation~\cite{eberhart1995new}. with $\mu$= 50,~$c_1=1.5, c_2=2,\omega=1$ (linearly decreased) \\ \hline

\textbf{$GWO$}   &   per-individual  & Grey Wolf Optimiser~\cite{mirjalili2014grey}.  with $\mu$= 50, $\alpha=2$ (linearly decreased to zero)   \\ \hline

\textbf{$AGWO$}   &   per-individual  & Adaptive Grey Wolf optimiser. Where $\mu=50$, $\alpha=2$ will be adaptively updated,  $CN_m^{Max}$=0.3, $CN_m^{Min}=10^{-6}$, $C_f=0.7$    \\ \hline

\textbf{$NM$} & per-frequency & Nelder-Mead search \cite{lagarias1998convergence} is run in all dimensions iteratively $MaxFunEvals=100$  \\ \hline

\hlineB{4} 
\multicolumn{3}{ |c| }{\textbf{Cooperative Evolutionary Ideas }}\\
\hlineB{4}
\textbf{$(2+2)CMA-ES + NM$} & per-individual $\&$ frequency  & CMA-ES ($\mu=\lambda=2$) cooperates with Nelder-mead where the position optimisation is done by CMA-ES and Nelder-Mead adjusts the spring-damping coefficients of all buoys in the round robin fashion. \\ \hline

\textbf{$(1+1)EA + NM$} & per-frequency  & Cooperative strategy of 1+1EA (all position dimensions, $P_{Mu}=\frac{1}{N}$) with linearly decreasing mutation step size ($\sigma$) per generation~\cite{eiben2007parameter} at 100 iterations and then Nelder-Mead tries to optimise the PTO parameters in all dimensions.  \\ \hline

\textbf{$AGWO + NM$} & per-individual $\&$ frequency  & Adaptive GWO is in charge of optimizing the PTO settings of the buoys. Afterward, the position configuration of the best candidate is optimised by Nelder-Mead search. This cooperative process is repeated until the time budget runs out.  \\ \hline

\textbf{$CCOS$} & per-individual  & Cooperative Co-evolution of $SLPSO$ and $SaNSDE$ with Online Optimiser Selection~\cite{sun2018cooperative}. Where $\mu=50$, $\mathbb{A}=2$, $\mathbb{C}=N \times 2$.  \\ \hline

\textbf{$SLPSO_{II}$} & per-individual  & Double Social Learning Particle Swarm optimisation, Setup for $SLPSO_{II}$ from~\cite{sun2018cooperative}. $\mu=50$, $\mathbb{A}=2$, $\mathbb{C}=N \times 2$ \\ \hline

\textbf{$SaNSDE_{II}$} & per-individual  & Double Self-adaptive Neighborhood Search Differential Evolution \cite{sun2018cooperative}. $\mu=50$, $F$, $P_{cr}$ and $\rho$ are initialized at 0.5, but  updated adaptively. $\mathbb{A}=2$, $\mathbb{C}=N \times 2$ \\ \hline

\hlineB{4} 
\multicolumn{3}{ |c| }{\textbf{Hybrid optimisation methods (one-at-a- time) }}\\
\hlineB{4}
$LS+NM$  & per-frequency & Repeated Local Sampling + Nelder Mead search~\cite{neshat2018detailed} buoys are placed at normally distributed random offset ($\sigma=100m$) from previous buoy and then ($MaxSam=512$) the best candidate sample is chosen. Next, the PTO parameters of chosen sample are enhanced by Nelder-Mead search.   \\ \hline

$SLS+NM+B$& per-frequency &  Symmetric Local Sampling + Nelder-Mead + Backtracking~\cite{Neshat:2019:HEA:3321707.3321806}. The new buoy is locally placed by a symmetric search approach. Next, both configurations (positions and PTOs) are adjusted by Nelder-Mead iteratively. Finally, the backtracking strategy modifies least-well performing buoy's locations and PTOs.      \\ \hline

$HCCA$ & per-individual $\&$ frequency   & Hybrid Cooperative Evolution Algorithm. SLS sets an initial location for each new buoy; Nelder-Mead optimises the buoy's position; then adjusts the PTOs. The process iterates until all buoys are placed. Backtracking is then applied to improves the positions and PTO settings for some buoys which have the lowest absorbed power. \\
\hlineB{4}
\end{tabular}
}
\end{table*}
\subsection{Evolutionary Algorithms (All-at-once)}\label{sec:allatonce}
\label{subsec:EAs}
In these experiments, five popular EAs and a new adaptive variant of GWO are used to optimise all dimensions simultaneously. These EAs are: (1) covariance matrix adaptation evolutionary-strategy (CMA-ES)~\cite{hansen2006cma}, 
(2) Differential Evolution (DE)~\cite{storn1997differential}, 
(3) Particle Swarm optimisation (PSO)`\cite{eberhart1995new}, (4) Grey Wolf optimiser (GWO)~\cite{mirjalili2014grey} and (5) Nelder-Mead simplex direct search (NM) \cite{lagarias1998convergence}  is combined with a mutation operator (Nelder-Mead+Mutation). Furthermore, we introduce a new variant of GWO called the adaptive grey wolf optimiser (AGWO).

\subsubsection{Adaptive Grey Wolf optimiser (AGWO)}
The adaptive grey wolf optimiser is a new variant
of the grey wolf optimiser that tunes hyper-parameter
settings to improve performance in this search domain.

\paragraph{Overview of grey wolf optimiser (GWO)}
The GWO algorithm~\cite{mirjalili2014grey}  is categorized as a bio-inspired stochastic method that mimics grey wolves hunting behaviours in a pack. In the population, there are four classes of responsibility: the alpha wolf is responsible for leading the pack members. Following positions are allocated to beta and delta wolves; these assist the alpha in decision making.  The remainder of the pack is called omegas -- these help sample the search space. GWO simulates some aspects of the hunting process including 1) searching for the prey (optimum), 2) encircling the prey, 3) hunting and 4) attacking the prey.

\paragraph{Encircling the prey}
\begin{align}\label{eq:diff}
\begin{split}
\vec{D} &=|\vec{C}.\vec{X_p}(t)-\vec{X}(t)|\\
\end{split}
\end{align}
\begin{align}\label{eq:move}
\begin{split}
\vec{X}(t+1) &=\vec{X_p}(t)-\vec{A}.\vec{D}
\end{split}
\end{align}
where $\vec{D}$ describes the interval among the prey location $\vec{X_ p}$ and a member of the pack $\vec{X} $ in the current iteration ($t$). Additionally, There are two  coefficient vectors ($\vec{A}$ and~$\vec{C}$) for controlling the behaviours of the exploration and exploitation, which can be computed by Equations \ref{eq:coeffA}~and ~\ref{eq:coeffC}:
\begin{equation}\label{eq:coeffA}
\vec{A}=2.\vec{a}.\vec{r_1}-\vec{a} \to 0\le a \le 2
\end{equation}
\begin{equation}\label{eq:coeffa}
a=2-iter.(\frac{2}{Max_{iter}})
\end{equation}
\begin{equation}\label{eq:coeffC}
\vec{C}=2.\vec{r_2}
\end{equation}
where $a$ is linearly decreased from 2 to 0 during the optimisation process. And also two random variables $r_1$ and $r_2$ are generated between 0 and 1. 

\paragraph{Hunting}
For having a successful exploration in the search space, the search agents (solutions) positions are updated based on the knowledge of three best-sampled candidates (alpha, beta and delta). This is because we assume a prior that a nearby optimum can be found among these best. The position update formulas (Equations \ref{eq:allmove}, \ref{eq:move2} and \ref{eq:diff2}) are as follows. 
\begin{equation}\label{eq:allmove}
\vec{X}(t+1)=\frac{\vec{X_1}+\vec{X_2}+\vec{X_3}}{3}
\end{equation}
\begin{align}\label{eq:move2}
\begin{split}
\vec{X_1} &=\vec{X_\alpha}(t)-\vec{A_1}.\vec{D_\alpha}\\
\vec{X_2} &=\vec{X_\beta}(t)-\vec{A_2}.\vec{D_\beta}\\
\vec{X_3} &=\vec{X_\delta}(t)-\vec{A_3}.\vec{D_\delta}
\end{split}
\end{align}
\begin{align}\label{eq:diff2}
\begin{split}
\vec{D_\alpha} &=|\vec{C_1}.\vec{X_\alpha}-\vec{X}|\\
\vec{D_\beta} &=|\vec{C_2}.\vec{X_\beta}-\vec{X}|\\
\vec{D_\delta} &=|\vec{C_3}.\vec{X_\delta}-\vec{X}|
\end{split}
\end{align}
\paragraph{Attacking the prey (exploitation)}
The hunting manner is followed by attacking the prey and converging to the optimum positions. This can be achieved mathematically by reducing the $a$ variable from 2 to 0 gradually.  It is observed that when $|\vec{A}|<1$ search agents are forced to attack the prey that is like a local search (exploitation process). Inversely, where $|\vec{A}|>1$ leads to a global search (divergence) or exploration process. 

\begin{figure}
\centering
\subfloat[]{
\includegraphics[clip,width=0.49\columnwidth]{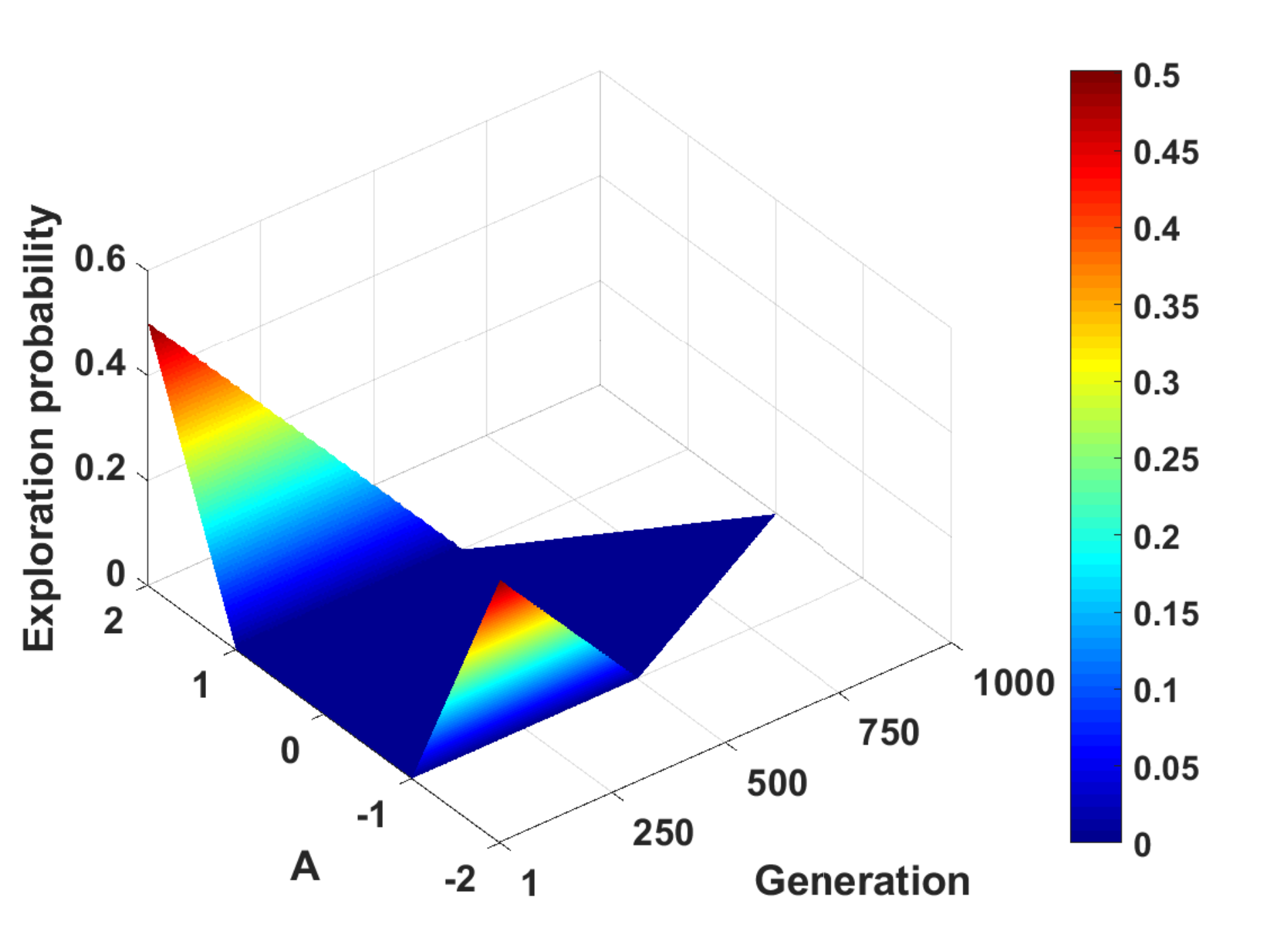}}
\subfloat[]{
\includegraphics[clip,width=0.49\columnwidth]{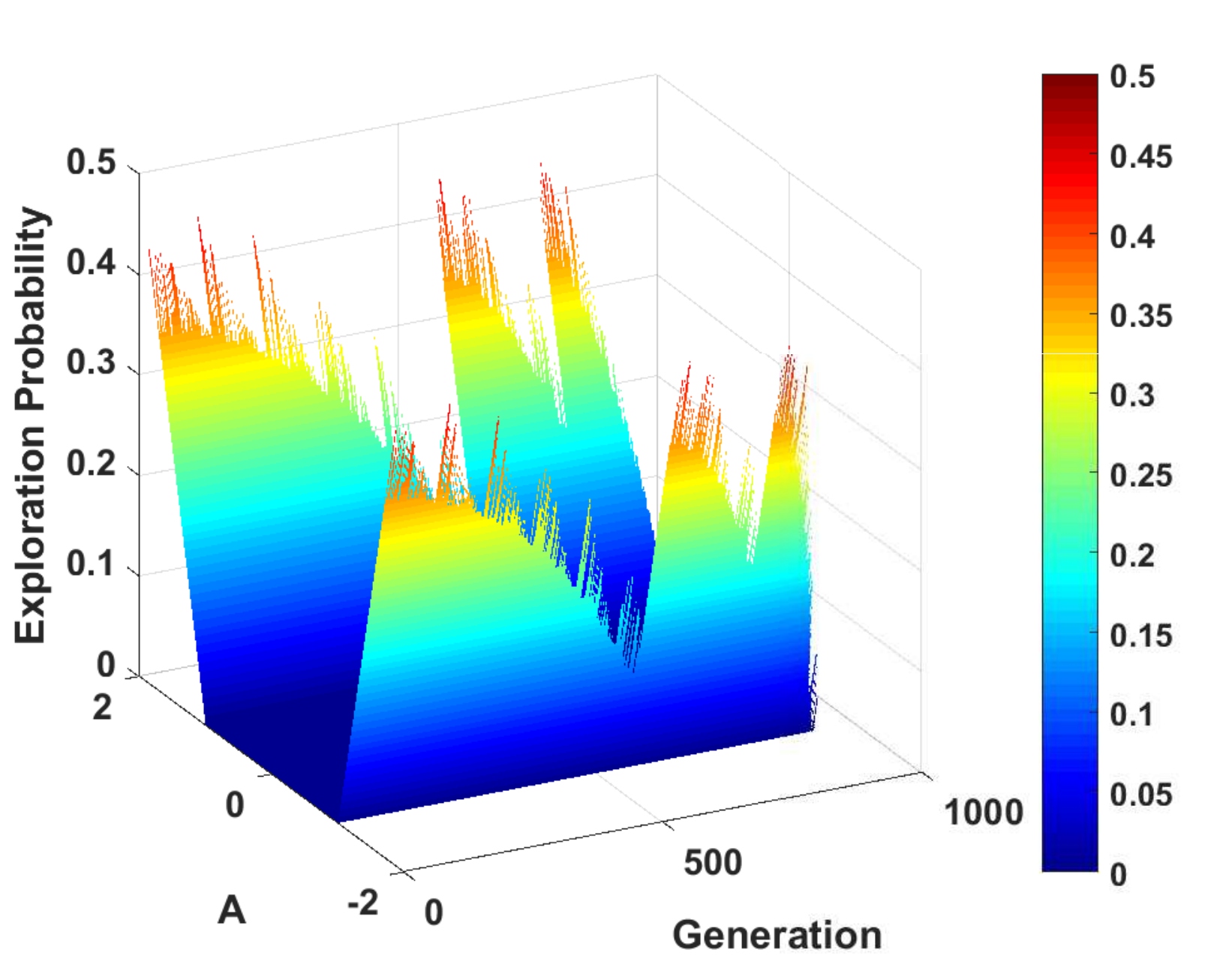}}\\
\subfloat[]{
\includegraphics[clip,width=0.49\columnwidth]{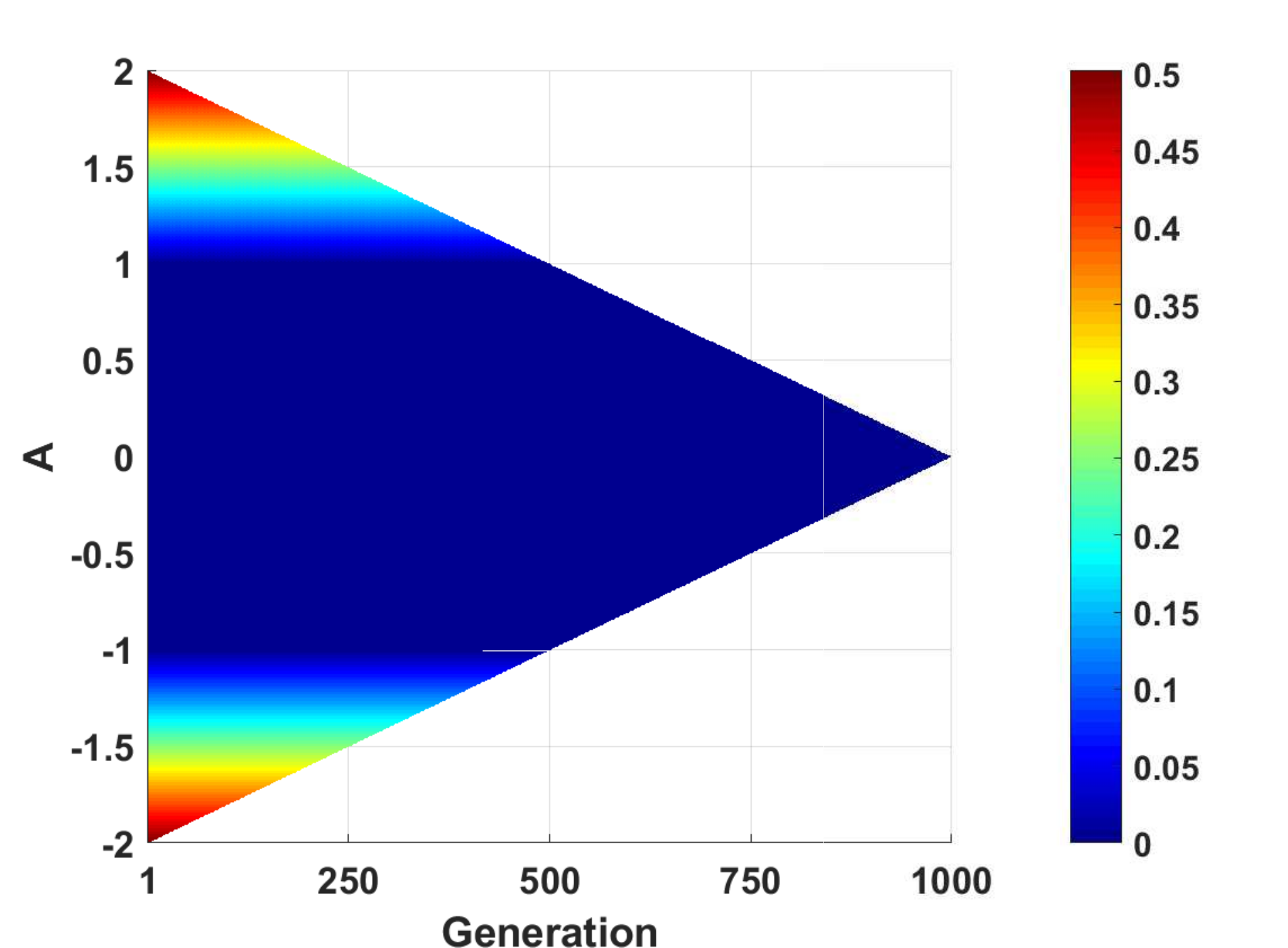}}
\subfloat[]{
\includegraphics[clip,width=0.49\columnwidth]{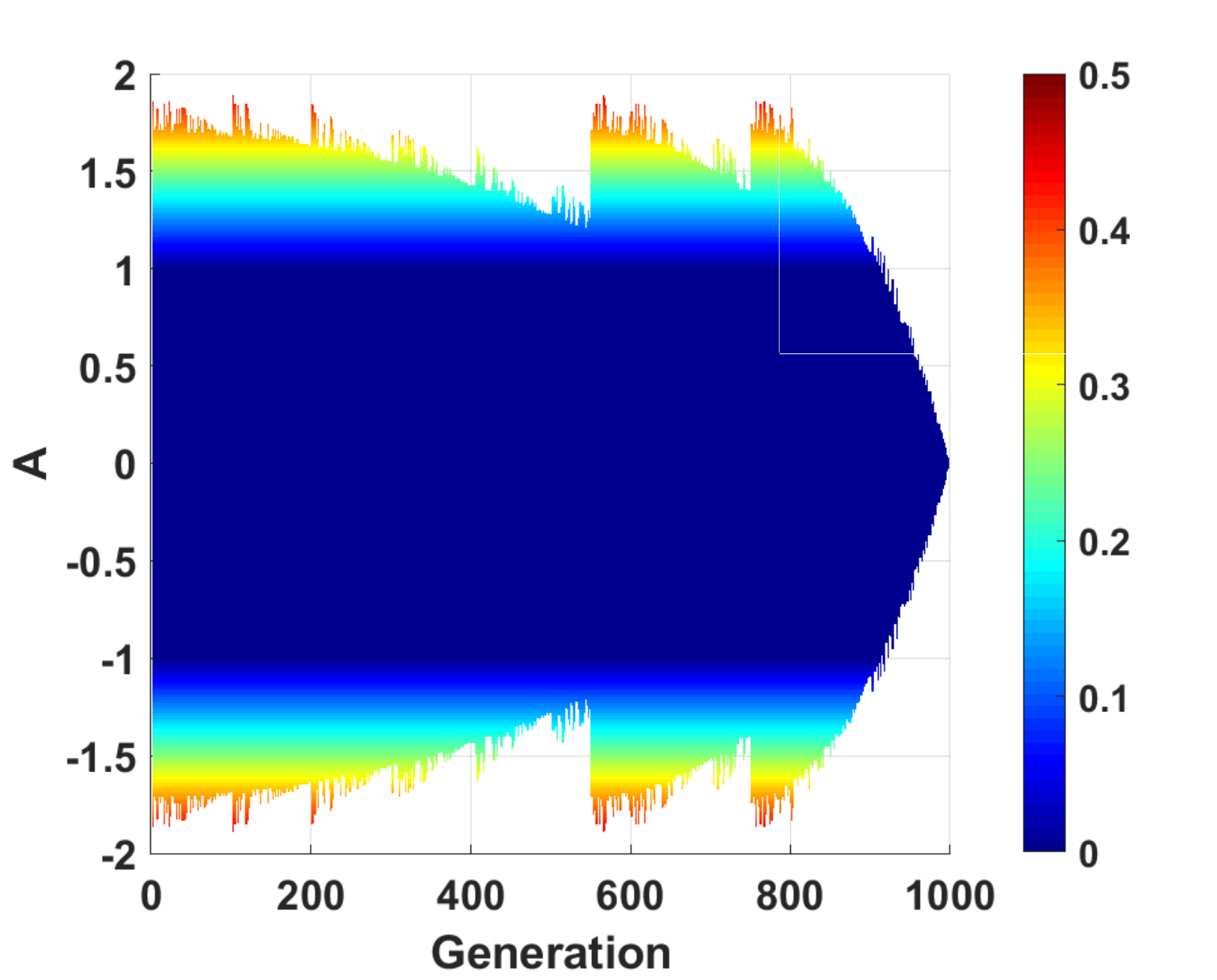}}
\caption{(a and c)The probability of original GWO exploration per generation (3D and 2D). (b and d) one example of the proposed adaptation mechanism for the control vector ($a$). These figures show the AGWO exploration probability per generation (3D and 2D). }%
\label{fig:control_variable}%
\end{figure}

\paragraph{Adaptive Grey Wolf optimiser (AGWO)}
One of the most critical parameters of GWO is $\vec{A}$ because it can adjust both diversification ($|\vec{A}| > 1$) and intensification ($|\vec{A}| < 1$) of the search process. According to Equation \ref{eq:coeffA}, the vector of $\vec{A}$ values can be between $-a$ and $a$ ($\vec{A}\in[-2a,2a]$), where $a$ is reduced linearly during the optimisation from 2 to 0. It means that the probability of exploration ($|\vec{A}| > 1$) at the initial iteration is 0.5 and will be linearly decreased until 0 in the middle of the search process. On the other hand,  the exploitation probability in the first iteration is 0.5 that is similar to exploration probability (giving a balanced-heuristic setting at the start); however,  the exploitation probability is gone up to 1 where half of the iterations are devoted ($iter=Max_{iter}/2$). Significantly, in the remaining iterations ($Max_{iter}/2$),  exploitation probability is 1, but exploration probability is 0 without any change. This issue is one reason why GWO is faced with premature convergence in some cases. Figure~\ref{fig:control_variable} (a and c) show this unbalanced search behavior. To overcome this shortcoming, a number of different mechanisms have been suggested.

\begin{figure}
\centering

\subfloat[]{
\includegraphics[clip,width=0.49\columnwidth]{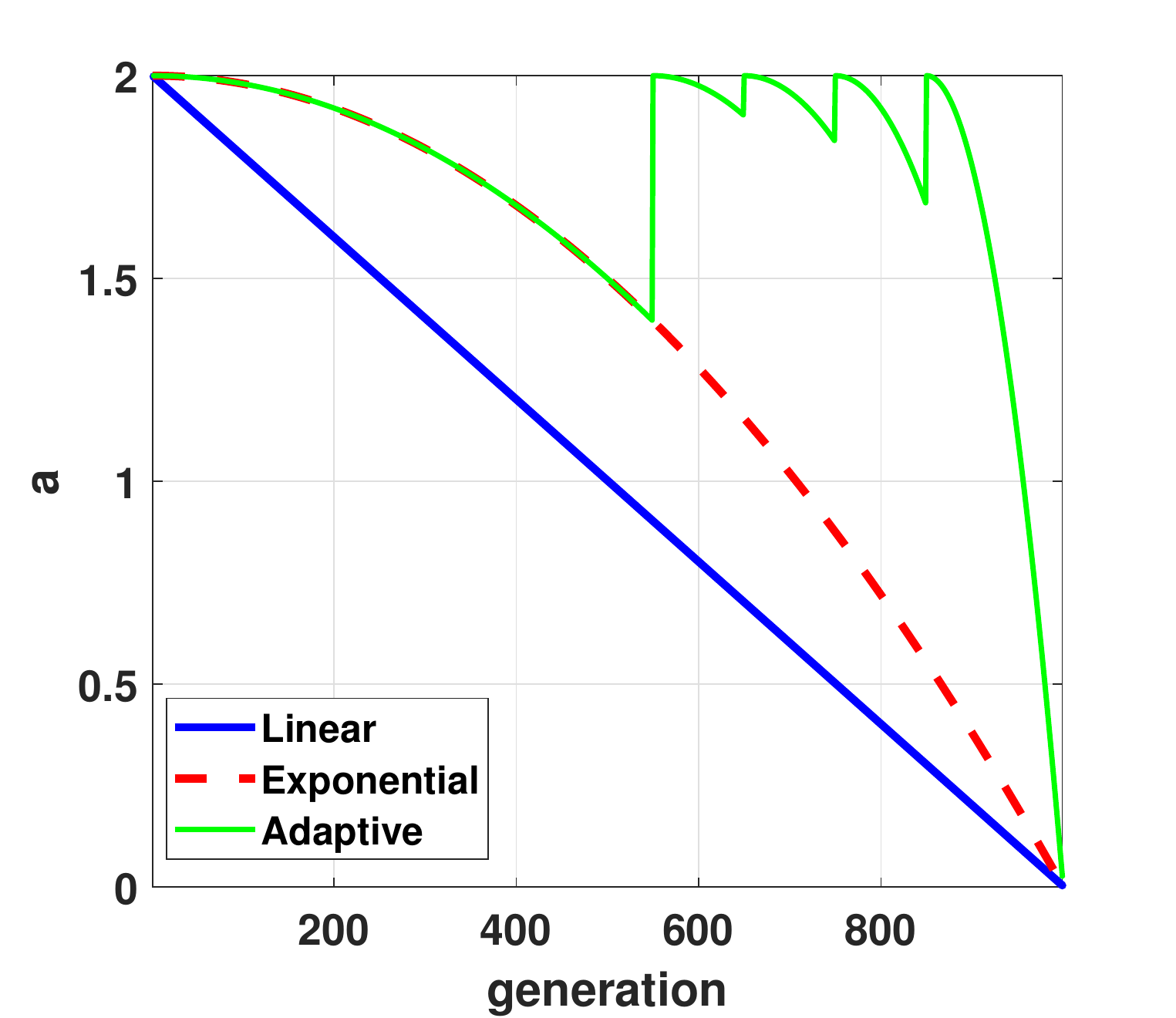}}
\subfloat[]{
\includegraphics[clip,width=0.49\columnwidth]{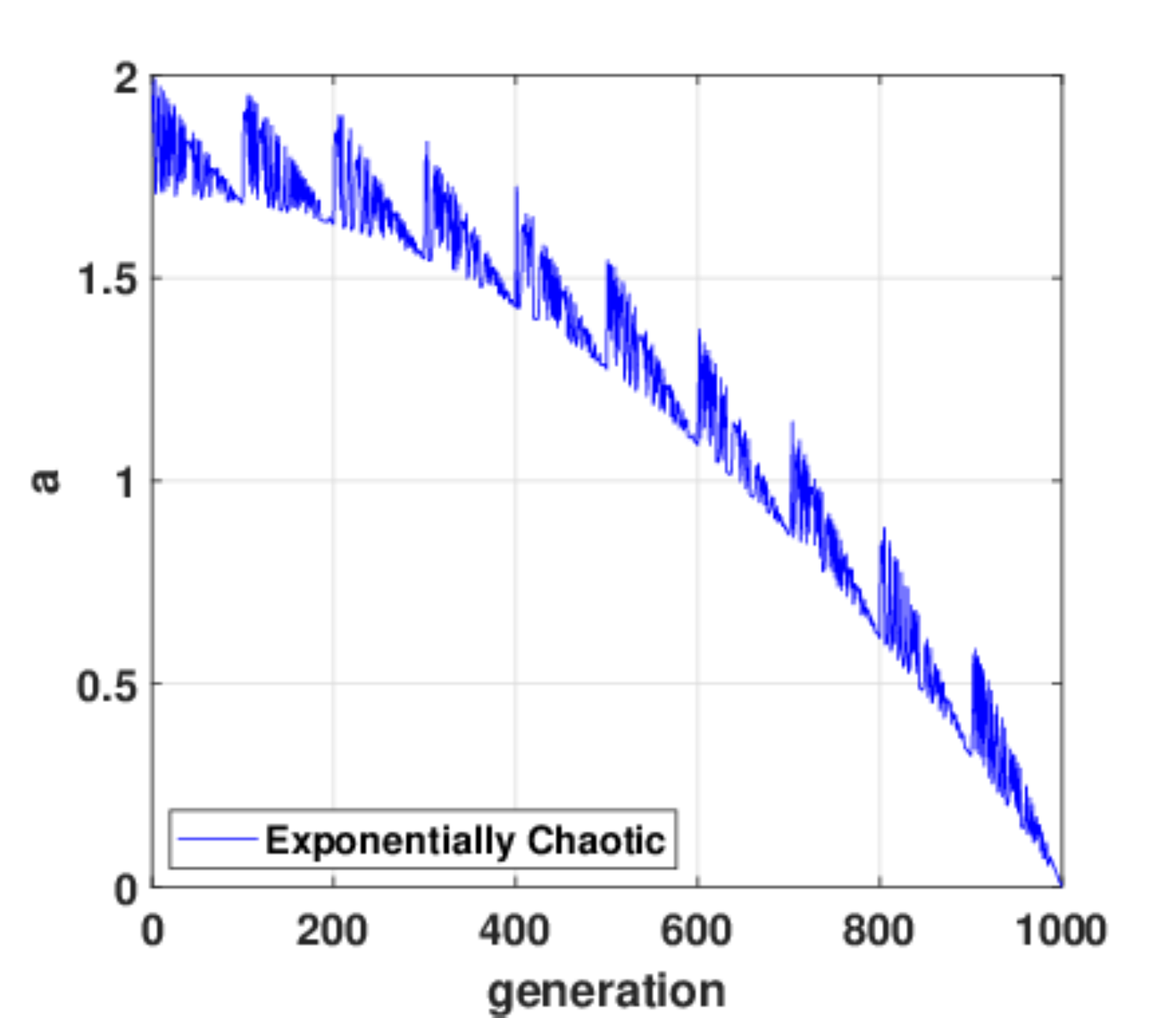}}\\
\subfloat[]{
\includegraphics[clip,width=0.49\columnwidth]{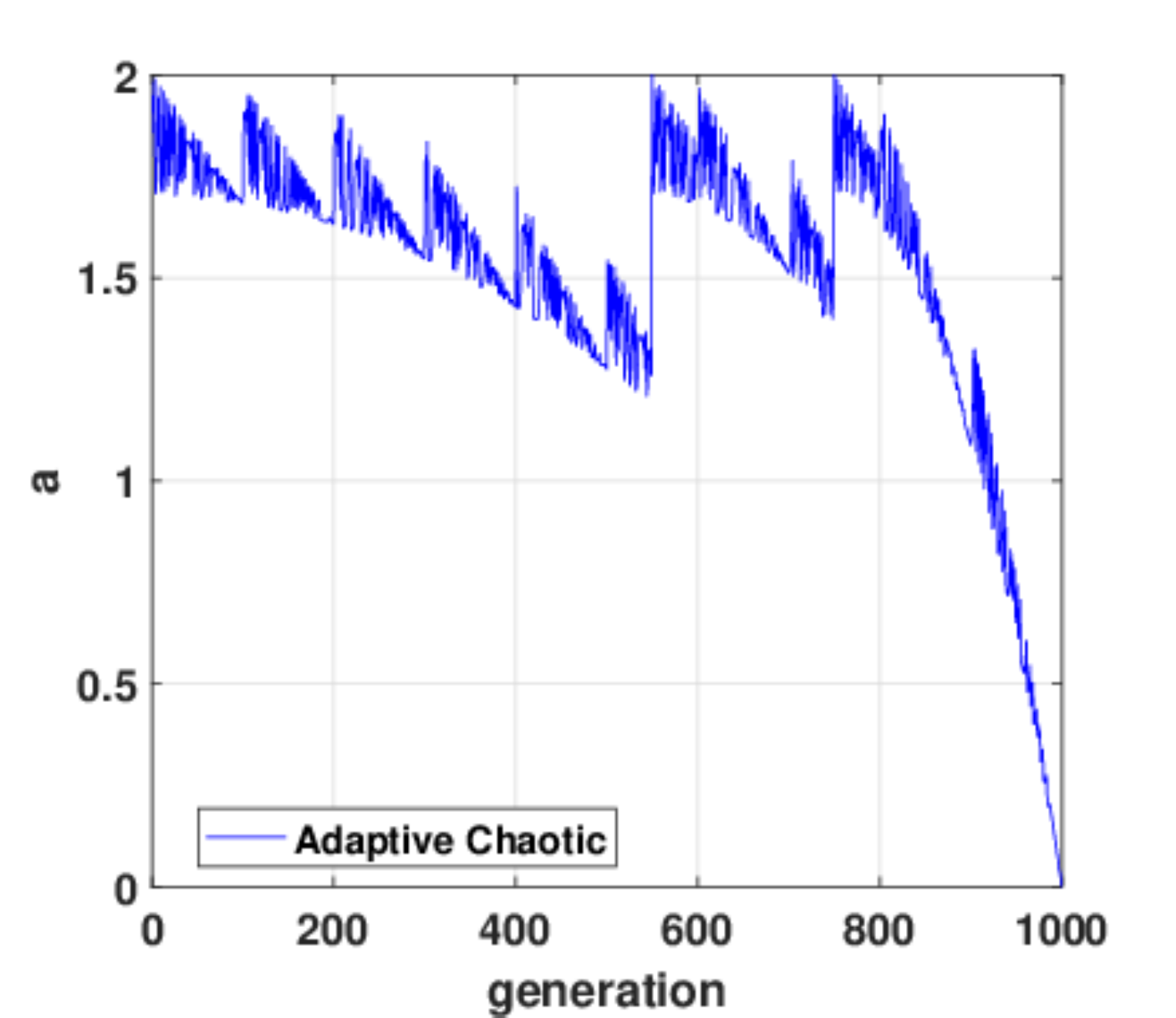}}
\subfloat[]{
\includegraphics[clip,width=0.49\columnwidth]{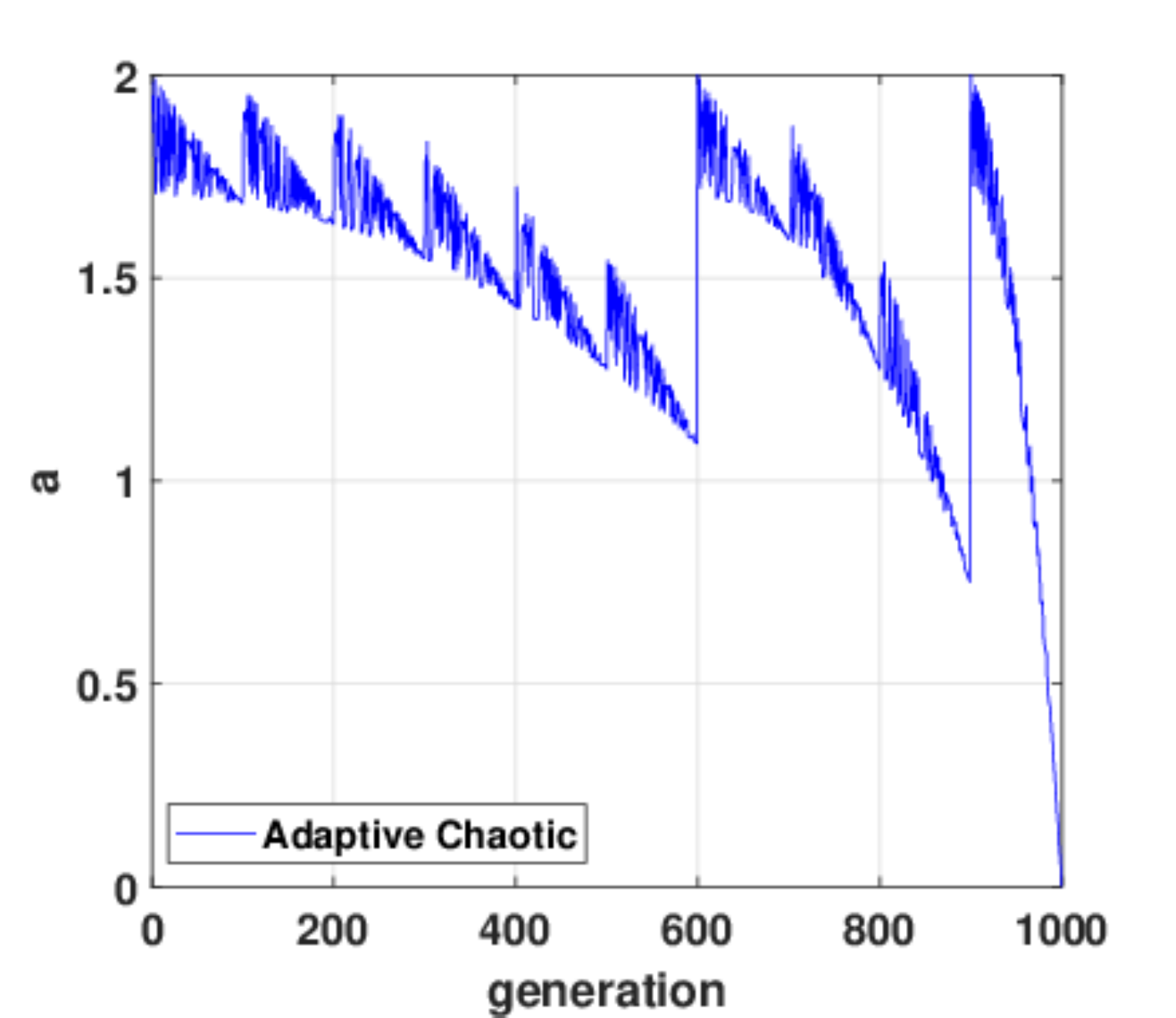}}
\caption{(a) the original strategy of control variable $a$  decreases linearly\cite{mirjalili2014grey}(blue line) and in \cite{mittal2016modified} it polynomially decreases (red line). This decay process can be adapted based on the optimisation process achievements (green line) which is introduced by this research. (b) shows a combination of polynomial and chaotic behavior of the control variable. This chaotic behavior is normalized and periodic. Graphs (c) and (d) show decay plots that combine both adaptive and chaotic decay. The use of adaptation and its combination with chaotic decay is new to the AGWO algorithm described here. }
\label{fig:chaotic_linear}%
\end{figure}

Mittal et al.~\cite{mittal2016modified} proposed an improved version for updating $\vec{a}$ in (mGWO) which decayed more slowly to improve exploration. Figure~\ref{fig:chaotic_linear}(a) represents this slower decay function. However, in this static mechanism after $70\%$ of the iterations, the value of $a$ has still decayed below $1$. A similar modification was introduced by Long et al.~\cite{long2017modified} in their Improved Grey Wolf optimiser (IGWO). More recently, Saxena et al. ~\cite{saxena2019beta} scaled the decay function using a $\beta$-chaotic sequence to allow for faster oscillation between exploration and exploitation phases during the parameter decay process. 

In previous research, various approaches were recommended for adjusting the $a$ parameter, but these ideas did not pay attention to the GWO performance during the optimisation. In this paper, we propose an adaptive mechanism for updating the control variable $a$ of GWO (AGWO). In this way, the optimisation performance is observed, and after a pre-determined period of $\rho$ iterations, where the best-found solution does not overcome the alpha particle, the control parameter should be incremented back to $2$. Moreover, a chaotic distribution is implanted with mapping by a normalize function for obtaining a great balance between exploration and exploitation.  The main AGWO contributions  can be seen in the following:       
\begin{enumerate}
\item Generating and combining a chaotic sequence with the control parameter ($a$) in each iteration. For achieving the best performance, ten various chaotic maps are applied and compared. Table \ref{table:chaoticmaps} shows the applied these chaotic maps in the adaptive idea. 
\item Using the normalization function periodically for distributing the chaotic sequence between upper and lower bias. The mathematical formulation of the function can be represented by Equation \ref{eq:CN}. 
\begin{equation}\label{eq:CN}
N_{m_{iter}}=CN_m^{Max}-(\frac{CN_m^{Max}-CN_m^{Min}}{Max_{iter_N}})\times iter_N
\end{equation}
Where the maximum and minimum values of the normalization function are $CN_m^{Max}=0.3$ and $CN_m^{Min}=10^{-6}$, respectively. And also $Max_{iter_N}$ is the maximum iterations in each period. Therefore, the normalized chaotic values ($CC_{iter}$) can be produced by Equation \ref{eq:CC}:
\begin{equation}\label{eq:CC}
CC_{iter}= N_{m_{iter}} * Cf
\end{equation}
Where $Cf$ is generated by the applied chaotic map. One example of the adaptive chaotic mechanism is presented in Figure~\ref{fig:control_variable}(b,d). The generated value of the chaotic sequence is embedded in control vector $\vec{a}$  and is presented in following equation: 
\begin{equation}\label{eq:Ca}
a= (2-CN_m^{Max})-(iter_c^2\times\frac{2-CN_m^{Max}}{Max_{iter_c}^2})+CC_{iter}
\end{equation}

\item  Introducing an adaptive mechanism for updating the control vector when the optimisation results are not satisfied for $\rho$ iterations.  When search stagnates in this way, the control vector is reset to $2$, and then the decay slope of the control vector is adjusted to a sharper gradient. This results in a switch from exploitation ($|\vec{a}|< 1$) to exploration when search stagnates. 
\end{enumerate}

Figure \ref{fig:chaotic_linear} demonstrates various mechanisms for updating the control vector include linear and polynomial ideas (a), and three samples of the new adaptive chaotic method (b,c and d).    

To sum up, AGWO is a combination of the ideas which consist of 1) an adaptive updating mechanism for tuning the $\vec{a}$ that depending on current search performance. Consequently, this feature of AGWO facilitates a balance between  exploration and exploitation processes throughout the entire search. 2) a chaotic sequence coefficient which scales the normalization function and further helps prevent the premature convergence (avoidance of local minima).  The pseudo-code of the  AGWO is presented in Algorithm \ref{alg:AGWO}. 

\begin{figure}
\centering

\subfloat[]{
\includegraphics[clip,width=0.49\columnwidth]{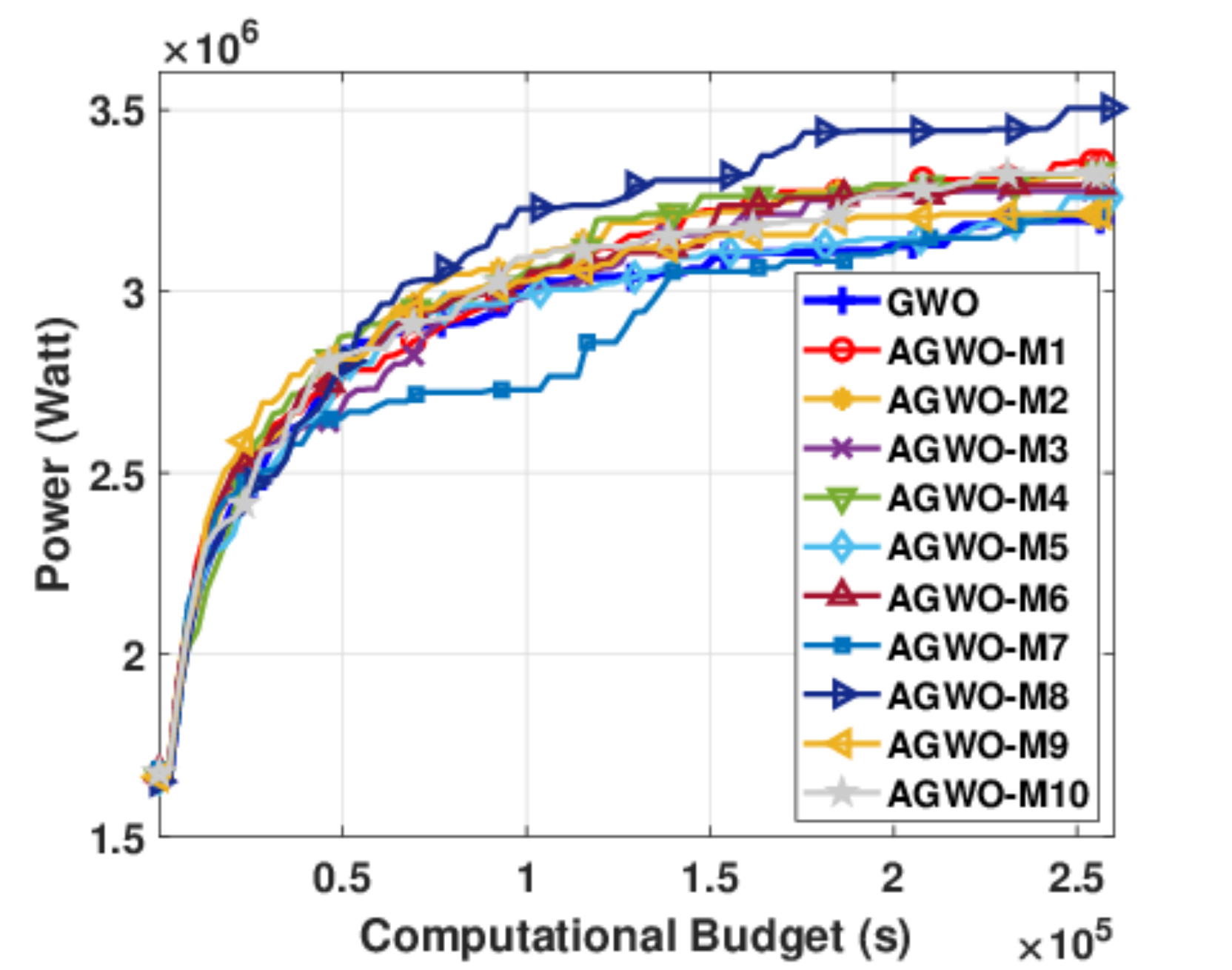}}
\subfloat[]{
\includegraphics[clip,width=0.49\columnwidth]{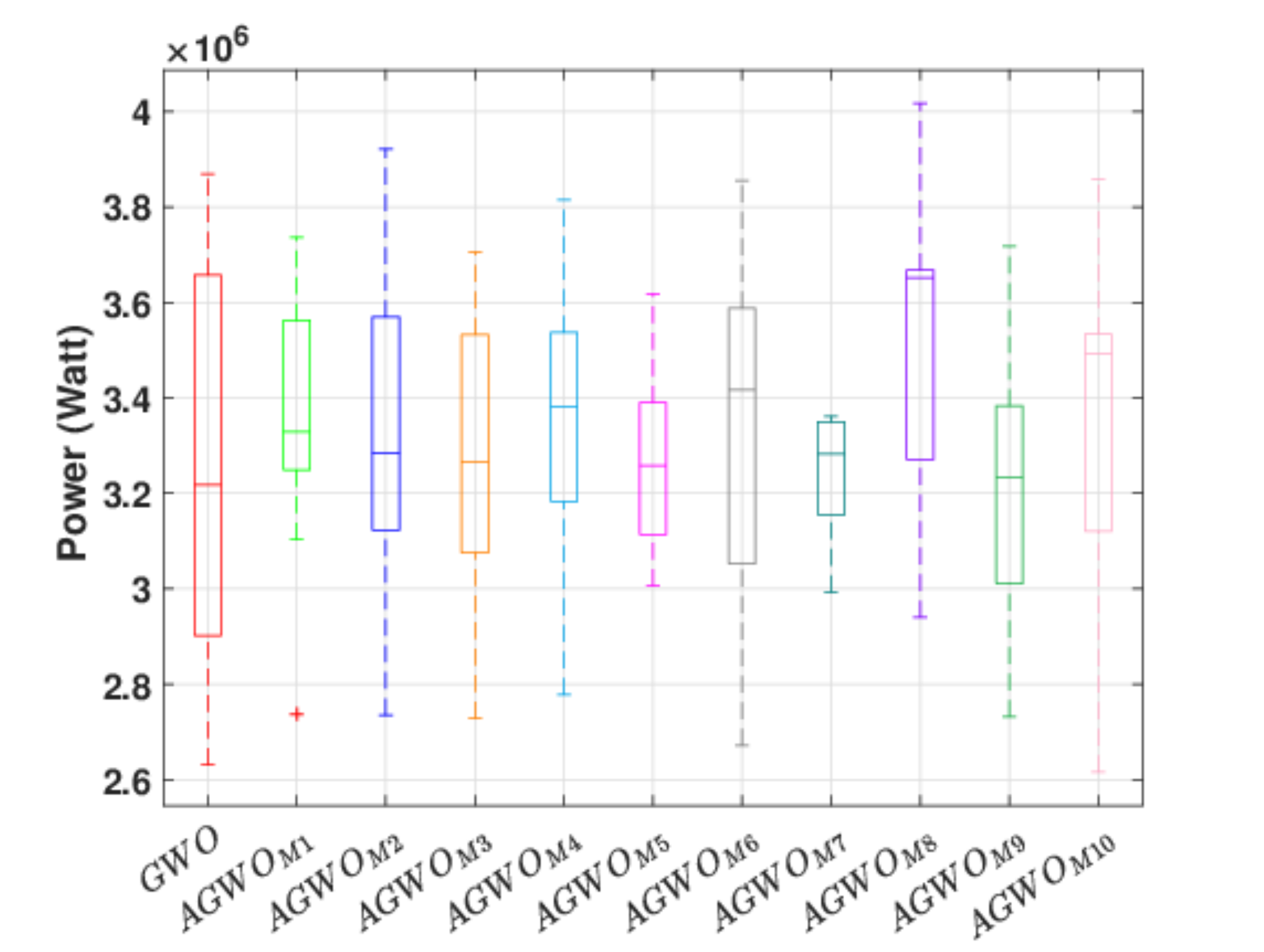}}

\caption{The convergence (a) and quality (b) comparison of the 10 different chaotic maps (M1-M10~Table \ref{table:chaoticmaps} ) combined with AGWO performance for 16-buoy layouts in Perth wave model. (10 independent runs for each configuration.) }%
\label{fig:chaoticmaps}%
\end{figure}

In order to measure and test the impact of various chaotic maps on the AGWO performance, a set of well-known chaotic maps \cite{kaur2018chaotic, saremi2014biogeography, wang2014chaotic,li2015modified} are applied. Table \ref{table:chaoticmaps} shows the details of these maps ($M_1,M_2,...,M_{10}$) and the summaries of AGWO parameters settings are listed in Table \ref{details:OP}. The parallelization of search is done per individual, and a 16-buoy layout size of the Perth wave model is selected as a case study. The results of Figure~\ref{fig:chaoticmaps} and Table~\ref{table:maps-results} are reported over ten independent runs. It can be seen that applying the chaotic maps with the adaptive strategy results in improved performance for GWO. In comparison to GWO results, the best performance is produced by the  $M_8(Singer)$  map with better convergence speed and the average total power outputs improved by by 3.83\% and 7.95\%, respectively. Based on this performance, we use $M_8(Singer)$ for the chaotic map in the following experiments. 

\begin{algorithm}
\small
\caption{$\mathit{Adaptive\, Grey\, Wolf\, optimiser\, (AGWO)}$}\label{alg:AGWO}
\begin{algorithmic}[1]
\Procedure{AGWO}{}
 \State $\mathit{size}=\sqrt{N*20000}$  \Comment{Farm size}
 \State $\mathit{Np}=50$  \Comment{Population size}
 \State $\mathbf{S}=\{\langle x_1,y_1,B_{k_1}^1,...,B_{k_{50}}^1,B_{d_1}^1,...,B_{d_{50}}^1 \rangle,\ldots$\\
  $,\langle x_{Np},y_{Np},B_{k_1}^{Np},...,B_{k_{50}}^{Np},B_{d_1}^{Np},...,B_{d_{50}}^{Np} \rangle\}$ \Comment{Initial Population}

\State $\mathit{energy}=Eval([S_{1},S_{2},\ldots,S_{Np}])$ \Comment{Evaluate Layouts}
\State $\mathbf{Initialize}$  parameters $\mathit{a, A, C, CN_m^{Max}, CN_m^{Min}}, \rho~and~ \mathit{Max_{iter_N}} $
\State $\mathit{X_\alpha}$=\em{The best layout from} $\langle S_{1},\ldots,S_{Np}\rangle$
\Comment{Find three best layouts}
\State $\mathit{X_\beta}$={\em{The second best layout from}}
$\langle S_{1},\ldots,S_{Np-1}\rangle$
\State $\mathit{X_\delta}$={\em{The third best layout from}} 
$\langle S_{1},\ldots,S_{Np-2}\rangle$

\While{{\em{stillTime()}}}
 
  \For{ $i$ in $[1,..,\mathit{Np}]$ }
  \State \em{Update} $S_i$ \em{by Equation~\ref{eq:allmove}}
  \If{ {\em{$S_i$ is not feasible}} }
       \State  $\mathit{S_i}=\mathit{Repair(S_i)}$ \Comment{ replacing by nearby feasible solution}
  \EndIf
     \EndFor 
     
     \State $\mathit{energy}=Eval([S_{1},S_{2},\ldots,S_{Np}])$ \Comment{Evaluate Layouts}
     \State $\mathit{BestEnergy_{iter}}$=\em{Max}($\mathit{energy}$)
     \If{rem($\mathit{iter},\rho$)=0~\&~ $\mathit{BestEnergy_{iter}<f(X_\alpha})$}
     \State  $\mathit{a=2, iter_N=1~and~iter_c=1}$ \Comment{Reset control variables}
     \State
     \Comment{Reset iteration of normalization and chaotic sequence} 
     \Else
      \State \em{Update}~$N_{m_{iter}},CC_{iter}$~and~$Cf$ \em{by Equation~\ref{eq:CN},\ref{eq:CC}}
     \State \em{Update}~$a,A$~and~$C$ \em{by Equation~\ref{eq:Ca},\ref{eq:coeffC},\ref{eq:coeffa}}
     \EndIf
  \State \em{Update}~$X_\alpha, X_\beta$~and~$X_\delta$
     \EndWhile 
\State \textbf{return} $\mathit{S},\mathit{energy}$  \Comment{Final Layout}
\EndProcedure
\end{algorithmic}
\end{algorithm}

\subsection{Cooperative optimisation methods}\label{sec:alternate}
Wave farm parameter (Position+PTO settings) optimisation has a very high dimensionality which makes it a challenging search problem. One option for dealing with this issue is to divide the decision variables into two subsets: WEC positions and PTO settings. This decreases the problem dimension and provides a more homogeneous search space.  Four cooperative optimisation techniques are proposed and compared including a new combination of AGWO and the Nelder-Mead, hybrid of (2+2)CMAES and Nelder-Mead \cite{Neshat:2019:HEA:3321707.3321806}, and a combination of a 1+1EA and Nelder-Mead \cite{Neshat:2019:HEA:3321707.3321806}, and, finally, the CCOS algorithm introduced in \cite{sun2018cooperative}. Details of these algorithms are as follows:

\subsubsection{AGWO + Nelder-Mead}(AGWO-NM)
As GWO is designed as an unconstrained meta-heuristic idea, it is not able to handle the constraint of WECs distances (safe distance) easily. However, GWO can be a fast and effective unconstrained optimisation method. In this way, a combination of AGWO and Nelder-Mead is proposed that AGWO adjusts the PTO configurations of WECs to achieve the highest power output and then NM is used for optimizing the arrangement of buoys. This optimisation process is run iteratively using the same computational budget until the runtime (three days) runs out.     

\subsubsection{Cooperative Co-evolution with Online optimiser Selection: CCOS
}
The Cooperative Co-evolution with an online mechanism for selecting the suitable optimiser (CCOS) introduced by Sun et al. \cite{sun2018cooperative}.  The CCOS consists of two general parts: decomposition and optimisation. In the first stage, a robust recursive algorithm ~\cite{sun2017recursive} is used to group parameters into subsets based on how they correlated during optimisation. These subsets are recursively decomposed according to the strength of parameters interactions.
The algorithm is able to decompose an $n$-dimensional problem using $\mathcal{O}(n\log{}n)$ steps. During the optimisation phase, two state-of-the-art adaptive optimisers are employed;  the social learning particle swarm optimiser (SLPSO~\cite{cheng2015social}), and self-adaptive differential evolution with neighbourhood search  (SaNSDE~\cite{yang2008self}). The main contributions of SaNSDE are 1) incorporating the search biases of distributions, Cauchy and Gaussian operators. SaNSDE takes into account the trade-off between small and large mutation step sizes; 2) all control parameters of SaNSDE are self-adapted based on statistical performance tracking during the optimisation process. 
Moreover, to assess the CCOS  algorithm (SLPSO+SaNSDE) thoroughly, we compared  CCOS's performance against the performances of a double SLPSO ($SLPSO_{II}$) and SaNSDE ($\mathit{SaNSDE_{II}}$). This evaluation helps isolate the impact of using these optimisers cooperatively. 

\subsection{Hybrid optimisation algorithms }\label{sec:hybrid}
In the earlier work, a practical WECs optimisation idea was developed~\cite{neshat2018detailed} called Local Search + Nelder Mead (LS-NM); that showed using a local sampling by a normal distribution in the previous buoy's neighbourhood (outside of the safe distance) combined with greedy selection could produce high-performing layouts. Such one-at-a-time placement is a fast optimisation strategy. However, tuning the position of the placed buoys required a considerable computational budget. This work also did not consider other WEC optimisation parameters such as PTO settings. 
 
 More recently ~\cite{Neshat:2019:HEA:3321707.3321806} proposed an improved heuristic (SLS-NM-B) for placing the new WEC one-at-a-time and tuning PTOs settings. As local sampling in LS-NM is done without strong regard to  useful priors of direction and distance, a repaired step is needed to modify the current position. SLS-NM-B represented a symmetric local search with the deterministic directions and bounded search space. Moreover, a backtracking strategy was introduced for improving the WEC parameters including both position and PTO settings (with the 
 latter being tuned in unison for each buoy). Nevertheless, SLS-NM-B was not designed to handle the high dimensional search problem that arises when all PTO frequency response settings are allowed to move independently. This is because the  Nelder-Mead optimiser that is applied for tuning the PTO parameters converges extremely slowly in the high-dimensional search space~\cite{han2006effect}.
 Moreover, it has not been proven that the PTO parameter space in this problem is uni-modal and so the downhill search heuristics such as Nelder-Mead may not be suitable for global optimisation. 

\begin{algorithm}
\small
\caption{$\mathit{HCCA}$}\label{alg:HCCA}
\begin{algorithmic}[1]
\Procedure{Hybrid Cooperative Co-evolution Algorithm}{}\\
 \textbf{Initialization}
 \State $\mathit{size}=\sqrt{N*20000}$  \Comment{Farm size and $N$ is buoy number}
 \State $\mathbb{A}_{ii}= \{1 \le ii \le |\mathbb{A}| \}$ \Comment{candidate optimisers}
  \State $\mathit{angle=\{0,45,90,\ldots,315\}}$ \Comment{symmetric samples angle}
 \State $\mathit{iters}=Size([angle])$ \Comment{Number of symmetric samples}
\State $\mathbb{S}=\{\langle x_1,y_1,B_{k_1}^1,...,B_{k_{50}}^1,B_{d_1}^1,...,B_{d_{50}}^1 \rangle,\ldots$
\State $\ldots,\langle x_N,y_N,B_{k_1}^N,...,B_{k_{50}}^N,B_{d_1}^N,...,B_{d_{50}}^N \rangle\}$ \Comment{Positions\&PTOs}
 \State $\langle \mathbb{S}_1,\mathbb{S}_2,...,\mathbb{S}_N \rangle=\mathit{Decompose(\mathbb{S})}$ \Comment{Decomposing $\mathbb{S}$ per buoy} 
 \State 
$
\begin{cases}
\mathbb{S}_1=\{\langle x_1,y_1 \rangle,\langle B_{k_1}^1,\ldots,B_{k_{50}}^1,B_{d_1}^1,\ldots,B_{d_{50}}^1 \rangle\}=\bot & \\
\mathbb{S}_2=\{\langle x_2,y_2 \rangle,\langle B_{k_1}^1,\ldots,B_{k_{50}}^2,B_{d_1}^2,\ldots,B_{d_{50}}^2 \rangle\}=\bot & \\
\ldots\\
\mathbb{S}_N=\{\langle x_N,y_N \rangle,\langle B_{k_1}^N,\ldots,B_{k_{50}}^N,B_{d_1}^N,\ldots,B_{d_{50}}^N \rangle\}=\bot & \\
                  \end{cases}
$
 \State $\mathit{U_{\mathbb{A}_{ii},\mathbb{S}_i}}=0$ \Comment{Initialize the accumulated contributions of optimisers}
\State $\mathit{S}_1=\{\langle size,0\rangle,\langle \vec{r_1}\times Max_{k},\vec{r_2}\times Max_{d}\rangle\}$  \Comment{initialize first buoy  }


 \If{$\mathit{i}=1 $} \Comment{optimise first buoy PTOs by the optimisers }
      \For{$ii$ in $|\mathbb{A}|$} \Comment{Calculate contribution}
\State $\langle \mathit{I_{(\mathbb{A}_{ii},\mathbb{S}_i)}},\mathit{Energy}\rangle$={\em{optimise}}~$(\mathit{\mathbb{S}_{i_{PTOs}}, \mathbb{A}_{ii}})$ 
\State $\mathit{U_{(\mathbb{A}_{ii},\mathbb{S}_i)}}$=($\mathit{\hat{U}_{(\mathbb{A}_{ii},\mathbb{S}_i)}}$ + $\mathit{I_{(\mathbb{A}_{ii},\mathbb{S}_i)}}$ ) / 2 \Comment{Accumulate contribution}
  \EndFor
    \State $\mathit{\mathbb{S}_{{i+1}_{PTOs}}}= \mathit{\mathbb{S}_{{i}_{PTOs}}}$  
  \EndIf
\State $\mathit{BestIndex=Max(\mathit{U_{(\mathbb{A}_{ii},\mathbb{S}_i)}}\to 1 \le ii \le |\mathbb{A}|)}$

 \For{ $i$ in $[2,..,N]$ } $\mathit{bestEnergy}=0;$

 \For{$j$ in $[1,..,\mathit{iters}]$}

\State $(Sample_{j},\mathit{energy_{j}})$={\em{SymmetricSample}}$(\mathit{angle_j},\mathbb{S}_{(i-1)})$ 
   
    \If{ {\em{$Sample_j$ is feasible $\&$ $energy_j$ $>$ bestEnergy}} }
       \State  $\mathit{tPos}=\mathit{Sample_j}$ \Comment{ Temporary buoy position}
      \State $\mathit{bestEnergy}=\mathit{energy_j}$
      \State $\mathit{bestAngle}=\mathit{j}$
  \EndIf
  
  \EndFor
  
   \State $(Es_1,Es_2)$={\em{SymmetricSample}}$(\mathit{bestAngle\pm15},\mathbb{S}_{(i-1)})$ 
   
   \State $(\mathbb{S}_{(i)},\mathit{energy})$={\em{FindbestS}}$(\mathit{tPos},Es_1,Es_2)$ 
   
   \State \textbf{\em{PTO settings Optimisation}}
   \State $\langle \mathit{I_{(\mathbb{A}_{BestIndex},\mathbb{S}_i)}},\mathit{Energy}\rangle$={\em{optimise}}~$(\mathit{\mathbb{S}_{i_{PTOs}}, \mathbb{A}_{BestIndex}})$ 
   
\State $\mathit{U_{(\mathbb{A}_{BestIndex},\mathbb{S}_i)}}$=($\mathit{\hat{U}_{(\mathbb{A}_{BestIndex},\mathbb{S}_i)}}$ + $\mathit{I_{(\mathbb{A}_{BestIndex},\mathbb{S}_i)}}$ ) / 2
\State $\mathit{BestIndex=Max(\mathit{U_{(\mathbb{A}_{ii},\mathbb{S}_i)}}\to 1 \le ii \le |\mathbb{A}|)}$
\State $\mathit{\mathbb{S}_{{i+1}_{PTOs}}}= \mathit{\mathbb{S}_{{i}_{PTOs}}}$  

   \State \textbf{\em{Position Optimisation}}
  \State $(\mathbb{S}_{i},\mathit{Energy})$={\em{Nelder-Mead}}$(\mathbb{S}_{i_{\mathit{Position}}})$ 
\EndFor 
\State \Comment{Call BackTracking procedure}
\State $\langle \mathbb{S}, \mathit{Energy} \rangle$=\textit{BackTracking} $(\mathbb{S},\mathbb{A}_{\mathit{BestIndex}})$  
\EndProcedure
\end{algorithmic}
\end{algorithm}

\subsubsection{Hybrid Cooperative Co-evolution algorithm (HCCA)}
One of the most effective strategies for solving the large-scale optimisation problems is Cooperative Co-evolution (CC) framework~\cite{yang2008large}. In CC, the general idea is dividing the decision variables into some components (decomposition) and employing one or more optimisers in a round-robin fashion (in biased or unbiased mode) for optimizing the sub-problems. In this paper, as a combination of WECs placements and PTOs settings forms a large number of decision variables ($N\times 102$) with a complex search space, we propose a new hybrid Cooperative Co-evolution (HCCA) method. The steps of the proposed hybrid algorithm are described in more details as follows.    

\paragraph{Decomposition:}
In the decomposition phase, we apply a knowledge-based approach according to the significant WECs hydrodynamic rule \cite{borgarino2012impact}. The rule is that both PTO parameters (damping coefficient ($dPTO$) and spring stiffness ($kPTO$)) of each converter should be optimised together. Therefore, the problem is decomposed into two sub-problems for each WEC, including PTO settings ($\langle B_{k_1}^i,...,B_{k_{50}}^i,B_{d_1}^i,...,B_{d_{50}}^i \rangle, 100D $) and position ($\langle x_i,y_i \rangle,2D$).  

\paragraph{optimisation:}
The HCCA optimisation phase is comprised of optimisation phases for the two-parameter groups listed above. For the buoy position parameter-group, the hybrid systematic neighbourhood search from(SLS-NM)~\cite{Neshat:2019:HEA:3321707.3321806} is applied by first uniformly sampling in search sectors whose boundaries are informed by an initial 2-buoy power landscape analysis. After this, a Nelder-Mead search is used to improve the best-sampled positions. 

In the second group of the optimisation, we propose a Cooperative Co-evolution idea for adjusting the PTOs settings that is a large-scale optimisation problem ($N \times 100$).  This CC framework is composed of three modern and efficient optimisers, SLPSO \cite{cheng2015social}, SaNSDE \cite{yang2008self} and a new proposed adaptive grey wolf optimiser (AGWO). The SLPSO is a competitive optimiser \cite{sun2018cooperative} for working in the context of CC because 1) it is computationally efficient, 2) needs no complicated fine-tuning of the control parameters, 3) has a high exploitation ability and convergence speed and 4) has worked well solving  dimensional optimisation problems. However, converging to a local optimum can be a problem 
encountered with SLPSO. Consequently, for developing the CC framework, combine this with another optimiser with a high capability of the exploration. The SaNSDE  optimiser has considerable capacity for exploration and has been broadly applied in the CC domain \cite{omidvar2017dg2}. The third optimiser used here in CC framework is AGWO which is the new GWO variant described earlier. 
During optimisation these three optimisers share the 
same population and and solve the components 
collaboratively. 
The Pseudo-code of the proposed HCCA algorithm for solving the WEC optimisation problem is shown in Algorithm \ref{alg:HCCA}.

\paragraph{Backtracking:}
After initial placement and PTO optimisation 
by the CC framework above a customized backtracking
 optimisation algorithm (BOA) 
 The BOA refines both buoy positions and PTO parameters.
 For positions, 
 the buoys with the lowest power output are selected and then NM is applied for optimizing the positions one at a time. For PTO parameters an optimiser is selected the best prior optimiser performance during the first search phase. The selected optimiser is then used to tune all of PTOs settings of the layout in all-at-once global search. 
 The pseudo-code of the backtracking approach is given in Algorithm \ref{alg:B}.

\begin{figure}
\centering
\includegraphics[width=\textwidth]
{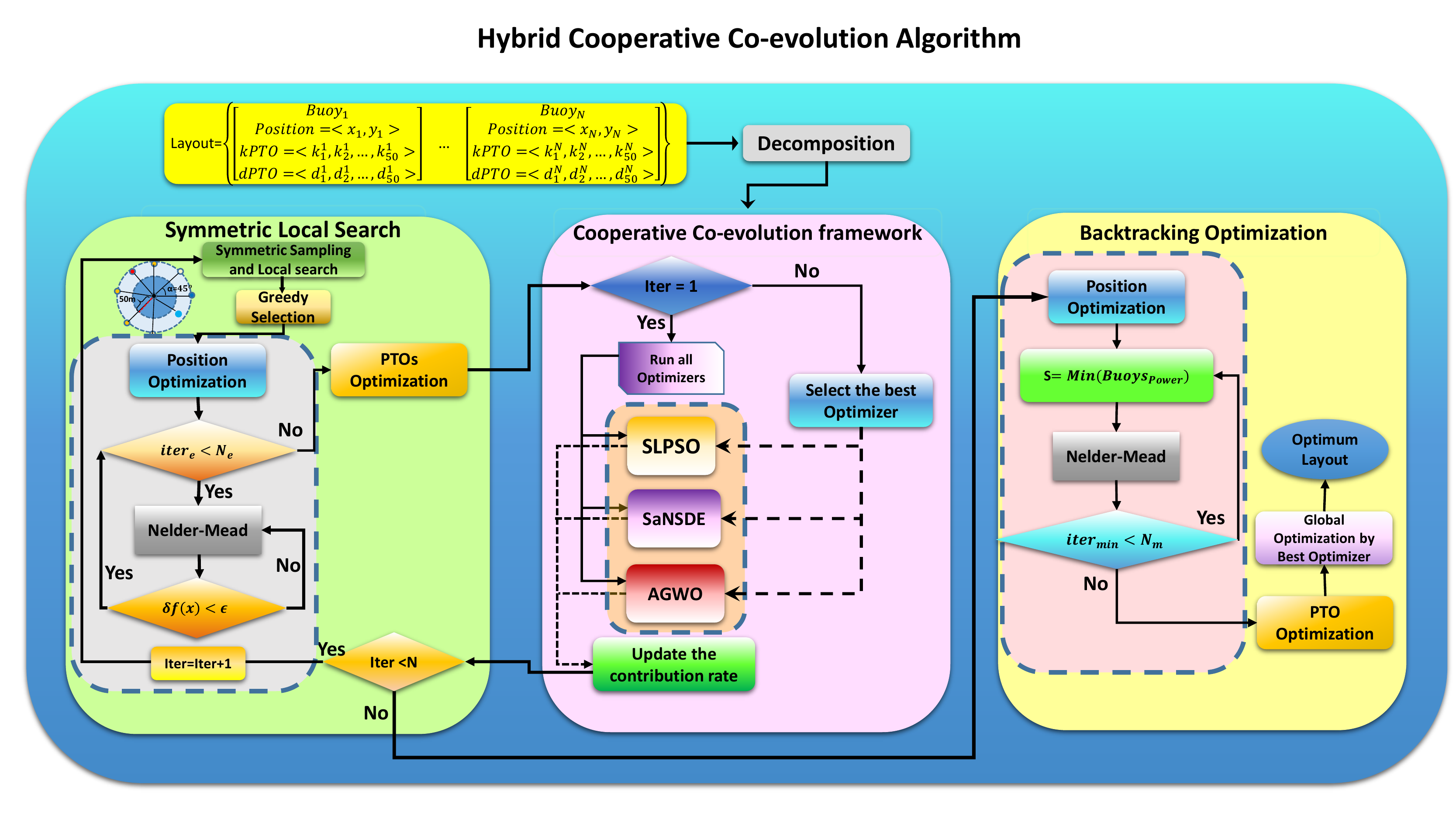}
\caption{Outline of the Hybrid Cooperative Co-evolution Algorithm (HCCA). $N,N_e$ and $N_m$ are the maximum buoy number in the layout, the maximum evaluation number of Nelder-Mead and the buoy numbers for refining their positions. $\epsilon$ is the Nelder-Mead function tolerance and stopping criterion.    }\label{fig:HCCA_land}
\end{figure}

\begin{figure}
\centering
\includegraphics[width=0.7\textwidth]
{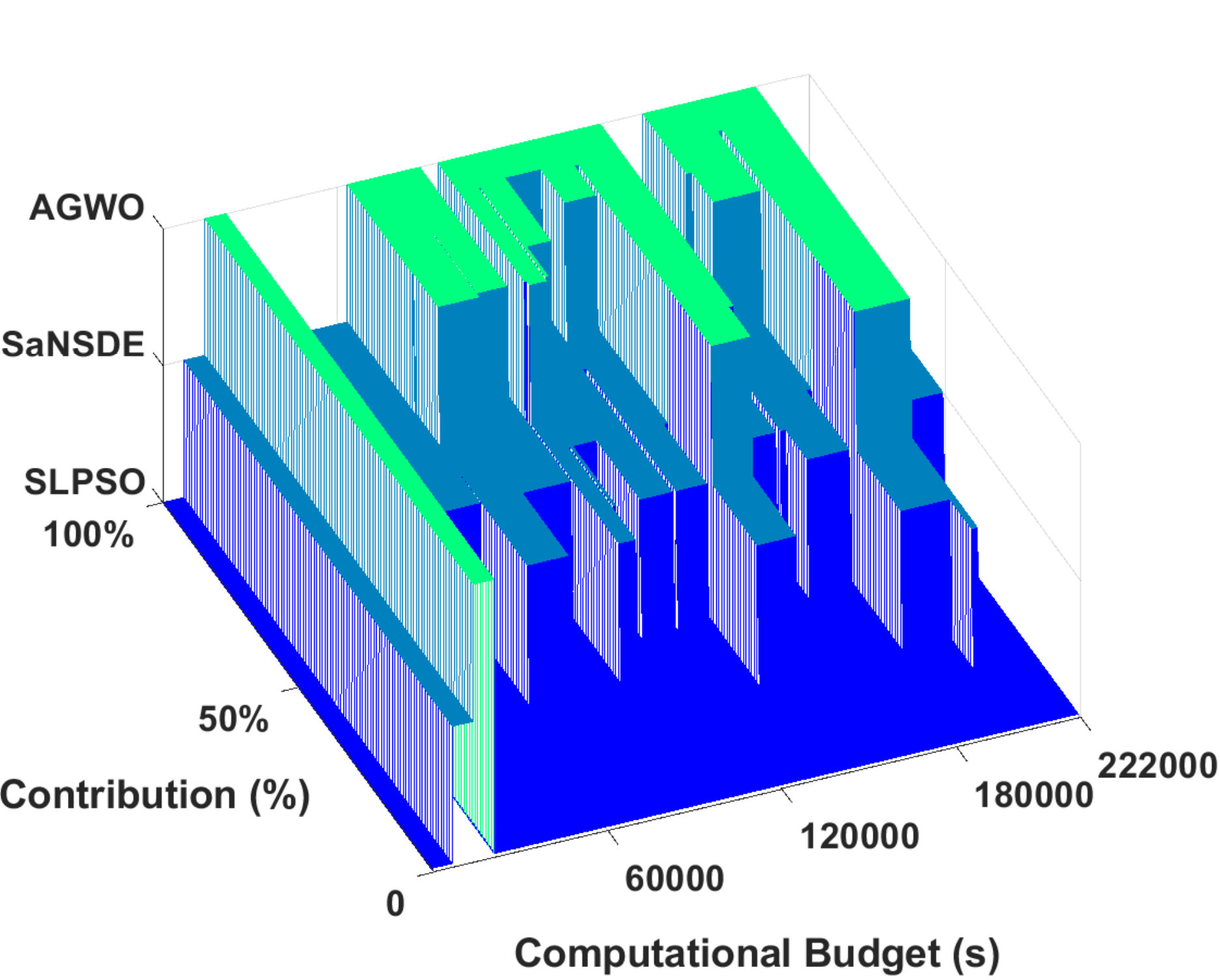}
\caption{The contribution percentage of HCCA optimisers (SLPSO, SaNSDE and AGWO) in the optimisation process when used to optimise the PTOs settings of 16-buoy layout in Perth wave Scenario.   }\label{fig:HCCA_contribution}
\end{figure}

Figure \ref{fig:HCCA_land} provides a graphical view of the proposed hybrid optimisation framework. In the first cycle, after placing the first buoy in a predefined location (recommended by \cite{Neshat:2019:HEA:3321707.3321806}), the three optimisers ($\mathbb{A}$) are employed to resolve PTOs settings ($\mathbb{S}_{j_{\mathtt{PTOs}}}$). Each optimiser is given the same computational budget.
Next, each optimiser's contribution is computed as a fitness improvement ($I_{(\mathbb{A}_{i},\mathbb{S}_j)}$):

\begin{equation}\label{eq:contribution}
I_{(\mathbb{A}_{i},\mathbb{S}_{j_{\mathtt{PTOs}}})}= \frac{f(\mathbb{\acute{S}}_{j_{\mathtt{PTOs}}})-f(\mathbb{S}_{j_{\mathtt{PTOs}}})}{f(\mathbb{\acute{S}}_{j_{\mathtt{PTOs}}})} ~~ i\in \{1 \le i \le |\mathbb{A}| \}, ~~ j\in \{1 \le j \le N \}
\end{equation}
where $f(\mathbb{\acute{S}}_{j_{\mathtt{PTOs}}})$ and $f(\mathbb{{S}}_{j_{\mathtt{PTOs}}})$ show the power of the layout obtained before and after employing $i^{th}$ optimiser in one cycle. The fitness improvement is a measure of the optimiser's ability to adjust the $j^{th}$ buoys PTOs settings. For updating the fitness improvement  for each optimiser during the whole optimisation process, an accumulated contribution variable is used~\cite{sun2018cooperative}. This performance tracking is encoded in Equation \ref{eq:accumulate}.  
\begin{equation}\label{eq:accumulate}
U_{(\mathbb{A}_{i},\mathbb{S}_{j_{\mathtt{PTOs}}})}= \frac{\acute{U}_{(\mathbb{A}_{i},\mathbb{S}_{j_{\mathtt{PTOs}}})} + I_{(\mathbb{A}_{i},\mathbb{S}_{j_{\mathtt{PTOs}}})}}{2} 
\end{equation}
where $\acute{U}_{(\mathbb{A}_{i},\mathbb{S}_{j_{\mathtt{PTOs}}})}$ tracks each optimiser's ($\mathbb{A}_{i}$) accumulated contributions. In the first cycle, the $\acute{U}_{(\mathbb{A}_{i},\mathbb{S}_{j_{\mathtt{PTOs}}})}$ is initialized to $0$. The accumulated contribution ${U}_{(\mathbb{A}_{i},\mathbb{S}_{j_{\mathtt{PTOs}}})}$  is the average of all fitness profits for each optimiser from previous cycles. In the next iteration, the best optimiser accumulated contribution ($\mathit{BestIndex=Max(\mathit{U_{(\mathbb{A}_{i},\mathbb{S}_{j_\mathtt{PTOs}})}}\to 1 \le i \le |\mathbb{A}|)}$) will be selected for optimizing the PTOs settings of the next placed generator. Figure \ref{fig:HCCA_contribution} presents the average contribution of each applied optimiser for tuning the PTOs configuration of 16-buoy layouts in the Perth wave model. It can be seen that after the initial cycle  the percentage contribution of SLPSO and SaNSDE is more than AGWO; however, for the last generations, AGWO's contribution is larger than that of the other optimisers.

After placing and optimizing the first buoy attributes, For each following buoy placement, eight symmetric local samples (SLS-NM) are done in different angles with the same resolution at $45^{o}$. However, the samples distances from the previously placed buoy are uniformly based on the bounded radial distance of between $50$ (safe distance) and $50+R'$. The best solution is chosen among all created symmetric samples. The infeasible solutions are ignored, and if all symmetric samples are infeasible, a feasible solution is performed by uniform random sampling. Also, then Nelder-Mead practices optimizing the position of the last-placed buoy. These both position and PTOs settings optimisation processes continue until the last buoy of the wave farm is placed and optimised. In the following, the backtracking strategy is run until the computational budget is depleted.

\begin{table}[]
\centering
\scalebox{0.7}{
\begin{tabular}{l|l|l|l|l|l|l|l|l|l|l|l}
\hlineB{5}
\multicolumn{12}{ c }{\textbf{\begin{large}Perth wave scenario (16-buoy)\end{large}}}\\ \\ \hlineB{5}
    & GWO       & $\mathit{AGWO_{M_1}}$      &  $\mathit{AGWO_{M_2}}$       &  $\mathit{AGWO_{M_3}}$       &  $\mathit{AGWO_{M_4}}$       &  $\mathit{AGWO_{M_5}}$       &  $\mathit{AGWO_{M_6}}$       &  $\mathit{AGWO_{M_7}}$       &  $\mathit{AGWO_{M_8}}$       &  $\mathit{AGWO_{M_9}}$       &  $\mathit{AGWO_{M_{10}}}$      \\ 
\hlineB{5}
\texttt{\textbf{Max}}  & 3869188 &  3736690& 3921497 & 3704994 & 3815293& 3618343& 3855858 & 3361619 & \textbf{4017436} & 3717341 & 3858006 \\ \hlineB{2}

\texttt{\textbf{Min}}  & 2631382 &  2737420 & 2735143  & 2728993 & 2778637 & 3006472 & 2671713  & 2993289 & 2940571 & 2731852 & 2616268  \\ \hlineB{2} 

\texttt{\textbf{Mean}} & 3258013&  3361803& 3324371 & 3276833 & 3335201& 3258500& 3324113 & 3216207 & \textbf{3517184} & 3212008 & 3331664 \\ \hlineB{2}

\texttt{\textbf{Median}} & 3218467 &  3328703 & 3285224  & 3265160 & 3382717 & 3235225 & 3417449 & 3230982 & \textbf{3664481}  & 3234115 & 3493016 \\ \hlineB{2}

\texttt{\textbf{STD}} & 428448 &  278820  & 328343& 280856 & 288771& 199893 & 385533 & 142654 & 322678 & 305389 & 350371 \\\hlineB{5}
\end{tabular}
}
\caption{Results of 10 chaotic maps on the case study of Perth wave model on AGWO}
\label{table:maps-results}
\end{table}

\section{Experiments}\label{sec:experiments}
This section starts with a small landscape study of the PTO parameter settings for a single buoy for the Perth wave scenario followed by 
a sensitivity analysis of the best-obtained 16-buoy layouts position. After that, we present the optimisation outcomes of the experiments comparing the effectiveness of the proposed algorithms explained above applied to WEC positions and PTOs settings under four real wave scenarios. In order to characterise the scalability of the frameworks, we apply them to both the 4-buoy ($=4\times102D$), and 16-buoy ($=16\times102D$) have been evaluated in the experiments. Finally, the algorithm convergence rates and 
the quality of the obtained solutions are also compared. 

\begin{figure}[tb]
\centering
\includegraphics[width=\textwidth]
{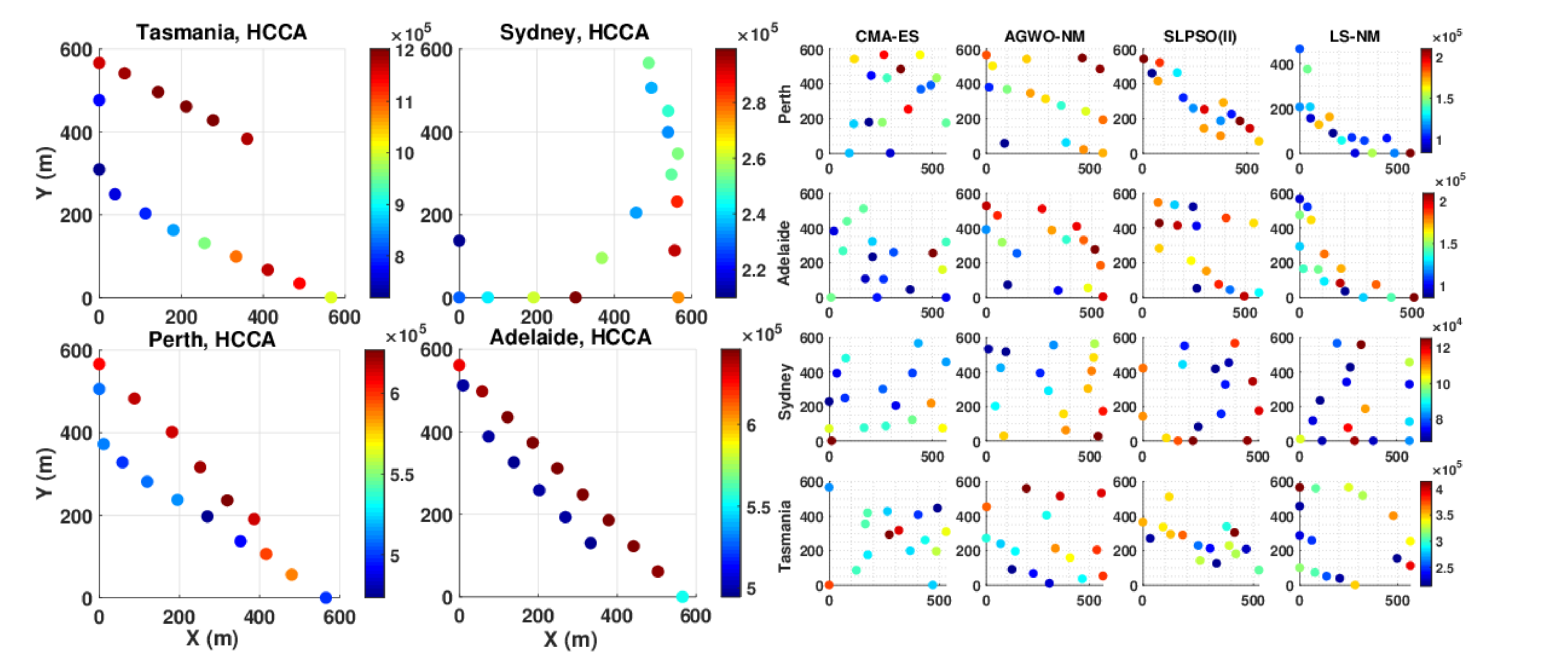}
\caption{The best-found 16-buoy layouts arrangement of the four real wave scenarios based on Table \ref{table:allresults16}. }\label{fig:allbest}
\end{figure}

\begin{table}[]
\centering
\scalebox{0.99}{
\begin{tabular}{l|l|l|l}
\hlineB{5}
 \textbf{NO.}&
\textbf{Name}&
\textbf{Chaotic Map}&
\textbf{Range} 
  \\          \hlineB{4}
                  
\texttt{\textbf{1}} &Chebyshev  & $x_{i+1}=cos(i cos^{-1}(x_i))$ &  (-1,1)       \\ 
 \hlineB{2}
\texttt{\textbf{2}} &Circle  & $x_{i+1}=mod(x_i+b-(\frac{a}{2\pi}) sin(2\pi x_i),1),a=0.5~and~ b=0.2$ &  (0,1)     \\
\hlineB{2}
\texttt{\textbf{3}} &Gauss/mouse  & 
$
 x_{i+1} = 
  \begin{cases} 
   1                          &  x_i = 0 \\
   \frac{1}{mod(x_i,1)}       & \text{otherwise}
  \end{cases}
  $
& (0,1)    \\
\hlineB{2}
\texttt{\textbf{4}} &Iterative  & $x_{i+1}=sin(\frac{a\pi}{x_i}),~a=0.7
$ &  (-1,1)        \\
\hlineB{2}
\texttt{\textbf{5}} &Logistic  & $x_{i+1}=a x_i(1-x_i),~a=4$ &   (0,1)      \\
\hlineB{2}
\texttt{\textbf{6}} &Piecewise  &
$
 x_{i+1} = 
  \begin{cases} 
   \frac{x_i}{P}              &  0\le x_i < P \\
   \frac{x_i-P}{0.5-P}                  & P\le x_i < 0.5  \\
    \frac{1-P-x_i}{0.5-P}                  & 0.5\le x_i < 1-P \\
   \frac{1-x_i}{P}       &  1-P\le x_i < 1
  \end{cases}
  , P=0.4
  $
&   (0,1)    \\
\hlineB{2}
\texttt{\textbf{7}} &Sine  & $x_{i+1}=\frac{a}{4}sin(\pi x_i),~a=4$ &   (0,1)   \\
\hlineB{2}
\texttt{\textbf{8}} &Singer  & $x_{i+1}=\mu(7.86x_i-23.31x_i^2+28.75x_i^3-13.302875x_i^4)~~,\mu=1.07$  &   (0,1)  \\
\hlineB{2}
\texttt{\textbf{9}} &Sinusoidal  & $x_{i+1}=ax_i^2 sin(\pi x_i)~~,a=2.3$ &    (0,1)    \\
\hlineB{2}
\texttt{\textbf{10}}&Tent  &
$
 x_{i+1} = 
  \begin{cases} 
   \frac{x_i}{0.7}                          &  x_i < 0.7 \\
   \frac{10}{3}(1-x_i)       & x_i\ge 0.7
  \end{cases}
  $
&   (0,1)  \\
\hlineB{2}
\hlineB{4}
\end{tabular}
}
\caption{The applied chaotic maps from \cite{saremi2014biogeography}.}
\label{table:chaoticmaps}
\end{table}

\subsection{Landscape analysis}
\subsubsection{PTOs settings analysis}
\begin{figure}[h]
\centering
\includegraphics[width=\textwidth]
{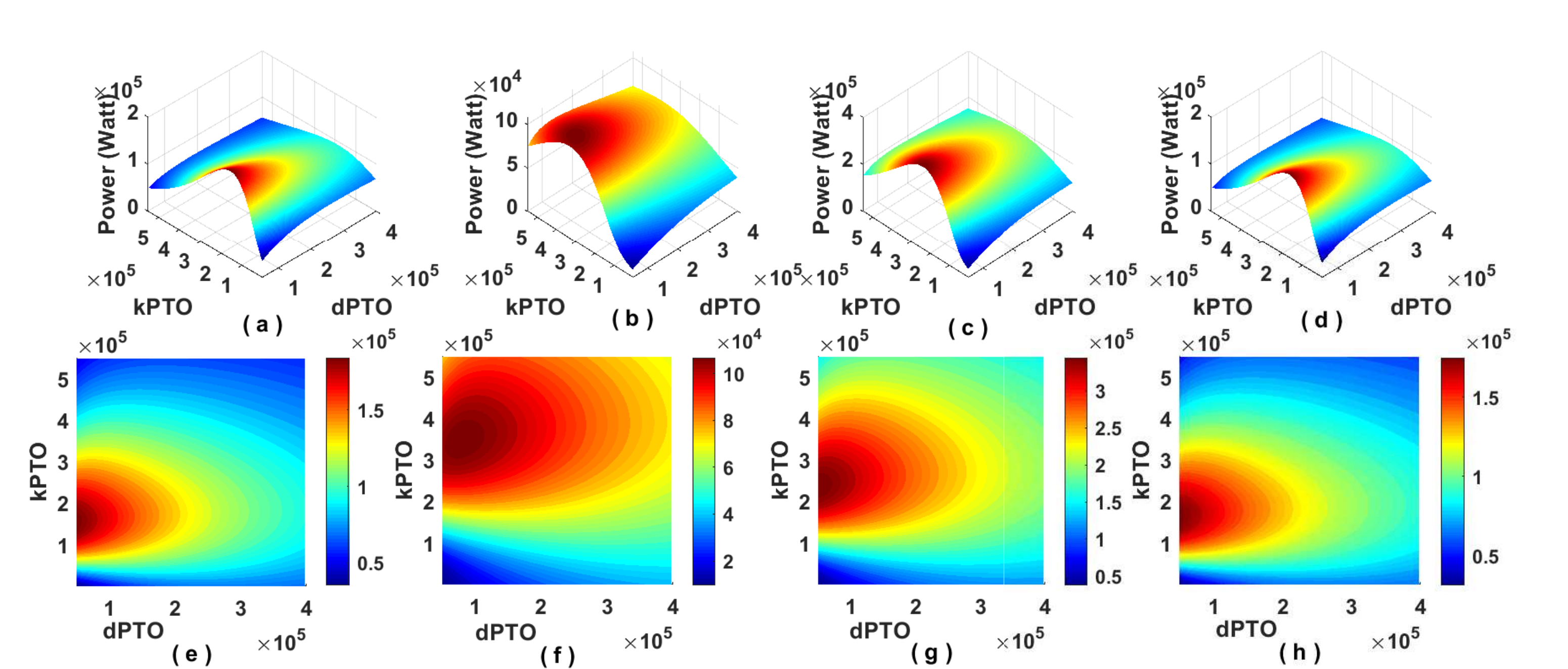}
\caption{PTOs settings power landscape analysis of four real wave scenarios (Adelaide(a,e),  Sydney(b,f), Tasmania (c,g) and Perth (d,h)) for one buoy layout.  We assume the most straightforward experiment of PTOs settings with the same value for all 50 wave frequencies. The spring-damping PTO configuration step size is 2500. Note that the real PTOs configurations search space is multi-modal and can be assigned by different values per each wave frequency. }\label{fig:PTO-same}
\end{figure}

In recent work \cite{Neshat:2019:HEA:3321707.3321806}, the impact of PTO parameter optimisation where these control parameters are kept the same for all wave frequencies for each buoy, was investigated and presented. This work found that tuning the PTOs parameters can be effective in optimising the total absorbed power of WECs (CETO model) in both Perth and Sydney wave climates by $4.48\%$ and $2.42\%$ , respectively.
Figure~\ref{fig:PTO-same} illustrates the PTO power landscape of one buoy with a simple grid search for tuning the damping-spring variables, where settings are kept the same for all wave frequencies. 

\begin{figure}[tb]
\centering
\includegraphics[width=\textwidth]
{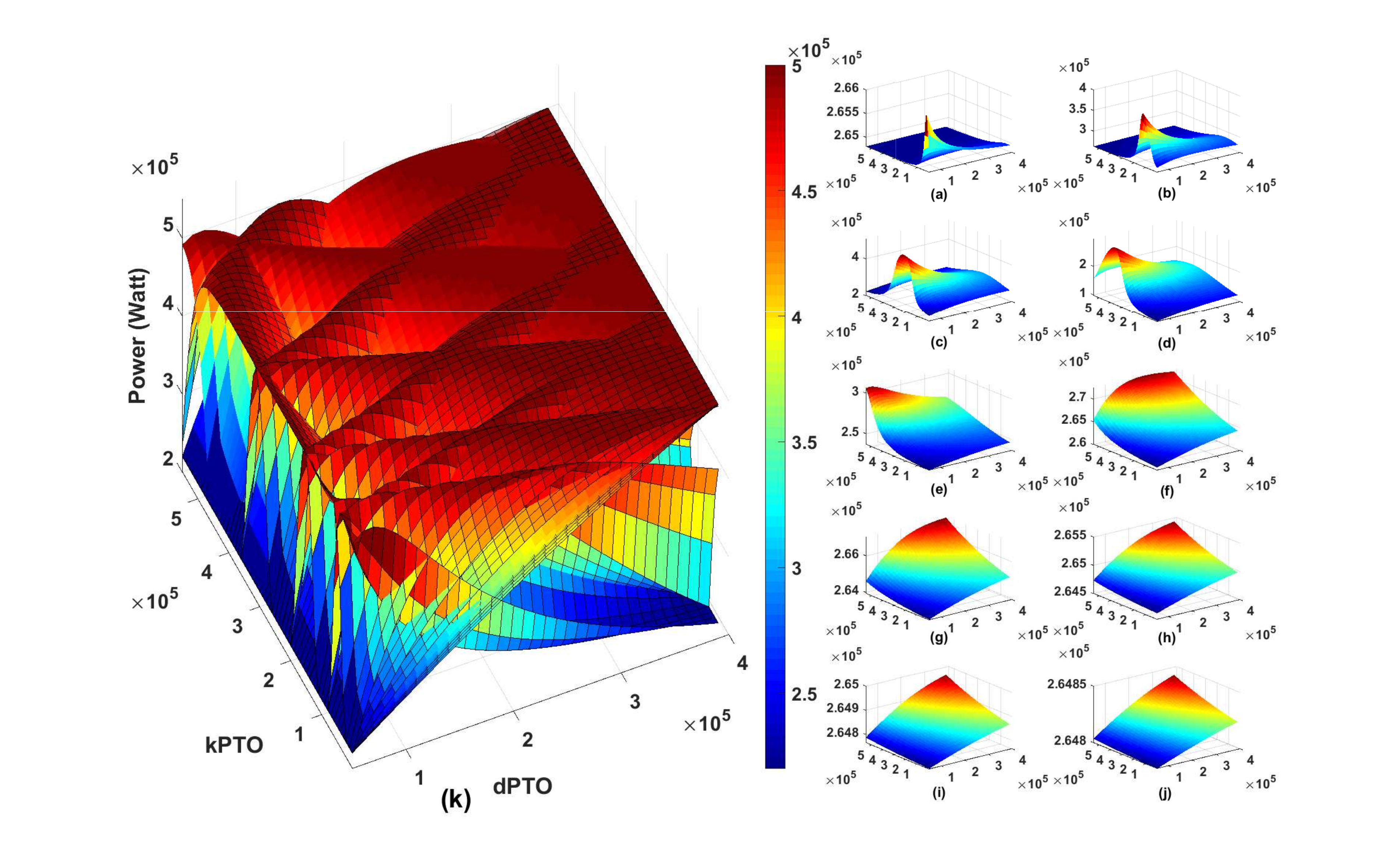}
\caption{The simplified power landscape of one buoy where PTO parameters are evaluated in ten sequential five-wave frequency groups. Figure (a) demonstrates the PTO (damping-spring parameters) power landscape of one buoy when we assume k and d parameters for $f_1,f_2,...,f_5$ are the same and other 45 wave frequencies are set by the predefined value ($dPTO=97412, kPTO= 407510$). Other figures follow the same pattern for instance Figure (b) represents the performance of a simple grid search ($Step=10000$) for plotting the power landscape of $f_6,f_7,...,f_{10}$ of tuned PTO parameters. In the left side, Figure (k) shows a normalized overlapping
of the surfaces of all ten landscapes in one graph for depicting the complexity level of the search space.    }\label{fig:PTO_landscape}
\end{figure}

However, in the real sea states, WECs control parameters (PTOs) should be tuned for each wave frequency. By tuning these parameters independently for each buoy, it is possible to extract more power.
We allow the PTO settings for all frequencies to be used for all proposed optimisation methods in this paper. For visualising the potential impact of PTO parameter optimisation for each wave frequency, a simplified experiment is done. Since the dimensionality of this problem is high ($2\times50$ for a single buoy), we divide the 50 frequencies into 10 groups. Each group includes five sequential frequencies, and we constrain them to have the same PTOs parameters. The 45 wave frequencies in each group are assigned by the  manufacturer's PTOs defaults ($k=407510$ and $d=97412$) \cite{neshat2018detailed}. 
Figure~\ref{fig:PTO_landscape} shows the modified PTOs optimisation power landscape of 10 groups for one buoy in Perth wave model, and for mixing all ten surfaces at one 3D figure, a normalised version of all landscapes is plotted (left figure) as a multi-layer 3D plot. We can see that this simplified search space of just one buoy PTOs is multi-modal and complex to search\footnote{Because this diagram is a low-dimensional projection from a higher-dimensional landscape it can't be automatically assumed that the higher-dimensional landscape for PTO optimisation at least, is also multi-modal. Note that previous work has shown that the buoy-positioning landscape is multi-modal, but exploring the multi-modality or otherwise of the entire higher-dimensional search landscape for PTO settings is future work.}. It is also of note that, even in this constrained search environment, there is a 3-fold improvement in extracted energy compared to previous studies~\cite{neshat2018detailed,Neshat:2019:HEA:3321707.3321806}. 

\begin{figure}[tb]
\centering
\includegraphics[width=0.7\textwidth]
{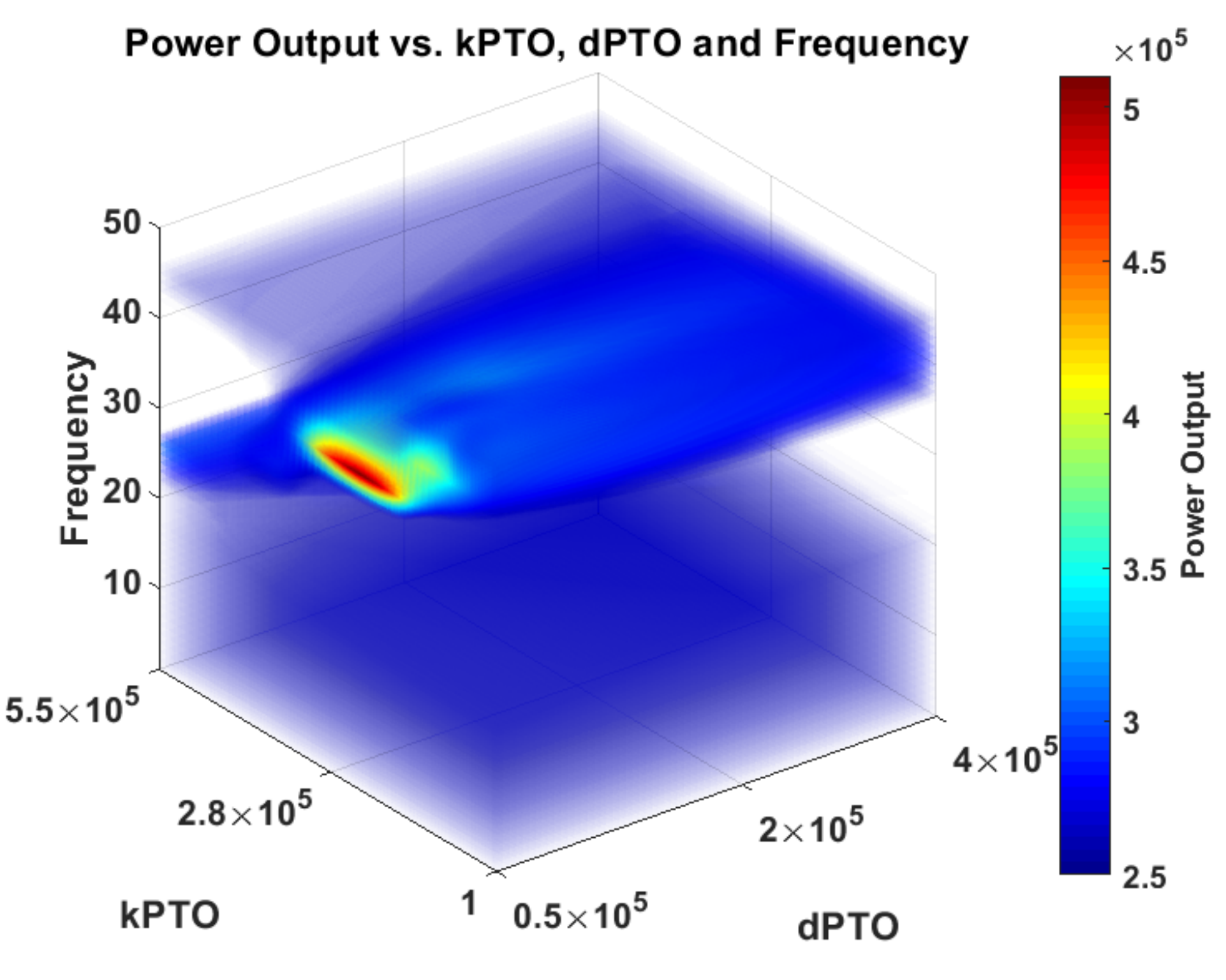}
\caption{A 4D view PTO power landscape for one buoy in the Perth wave model.}\label{fig:PTO_4D}
\end{figure}

Moreover, to provide an alternative visualization of this experiment, a 4D power landscape is plotted. Figure \ref{fig:PTO_4D} presents a trade-off of damping ($dPTO$), spring ($kPTO$), wave frequency and absorbed power. We can see that a specific range of frequencies with tuned values of PTO settings can produce more power. Note however the figure is plotted for one fixed buoy without the complex details of hydrodynamic interactions between buoys in the wave farm. A-priori, it is expected that introducing more buoys will produce interactions that will increase the complexity of this landscape further.

\begin{figure}
\centering

\subfloat[]{
\includegraphics[clip,width=0.8\columnwidth]{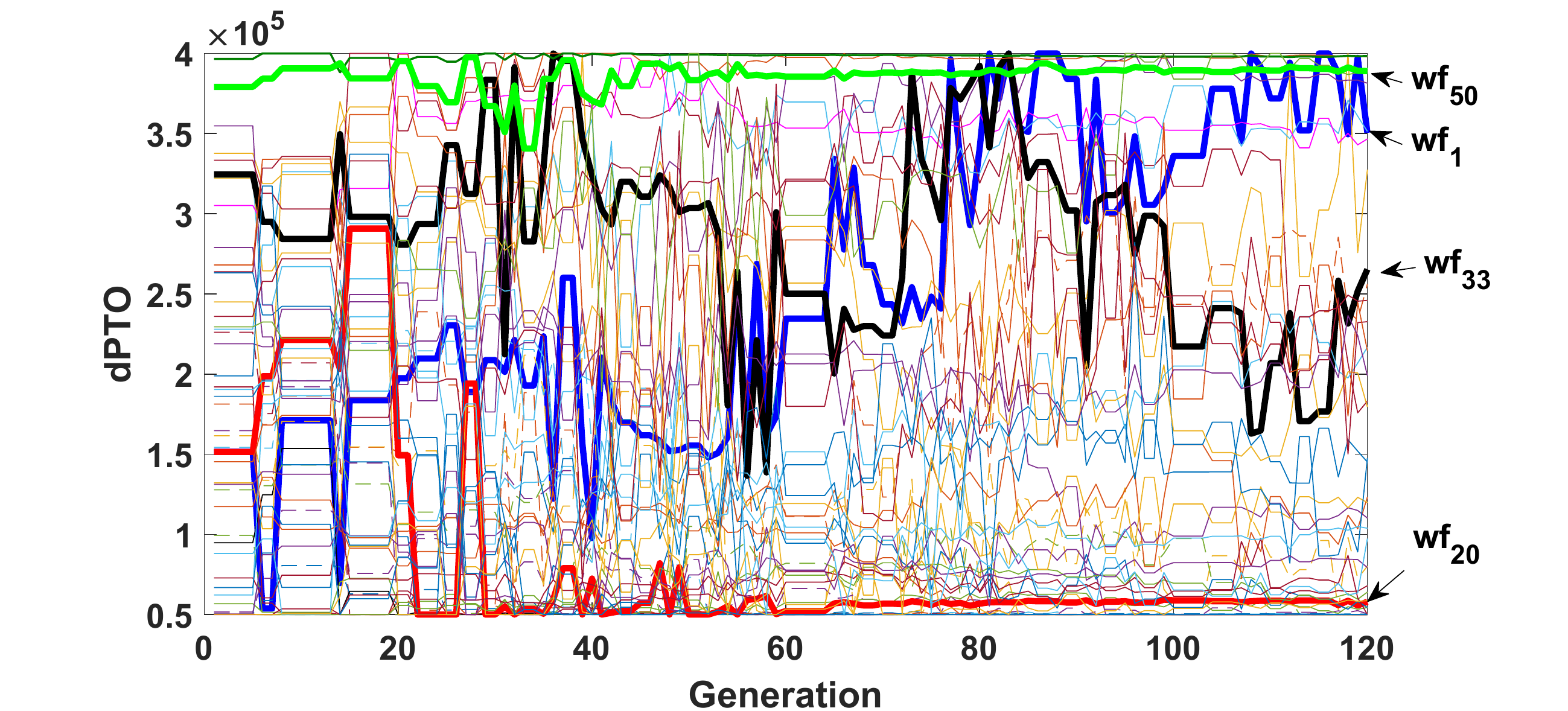}}\\
\subfloat[]{
\includegraphics[clip,width=0.8\columnwidth]{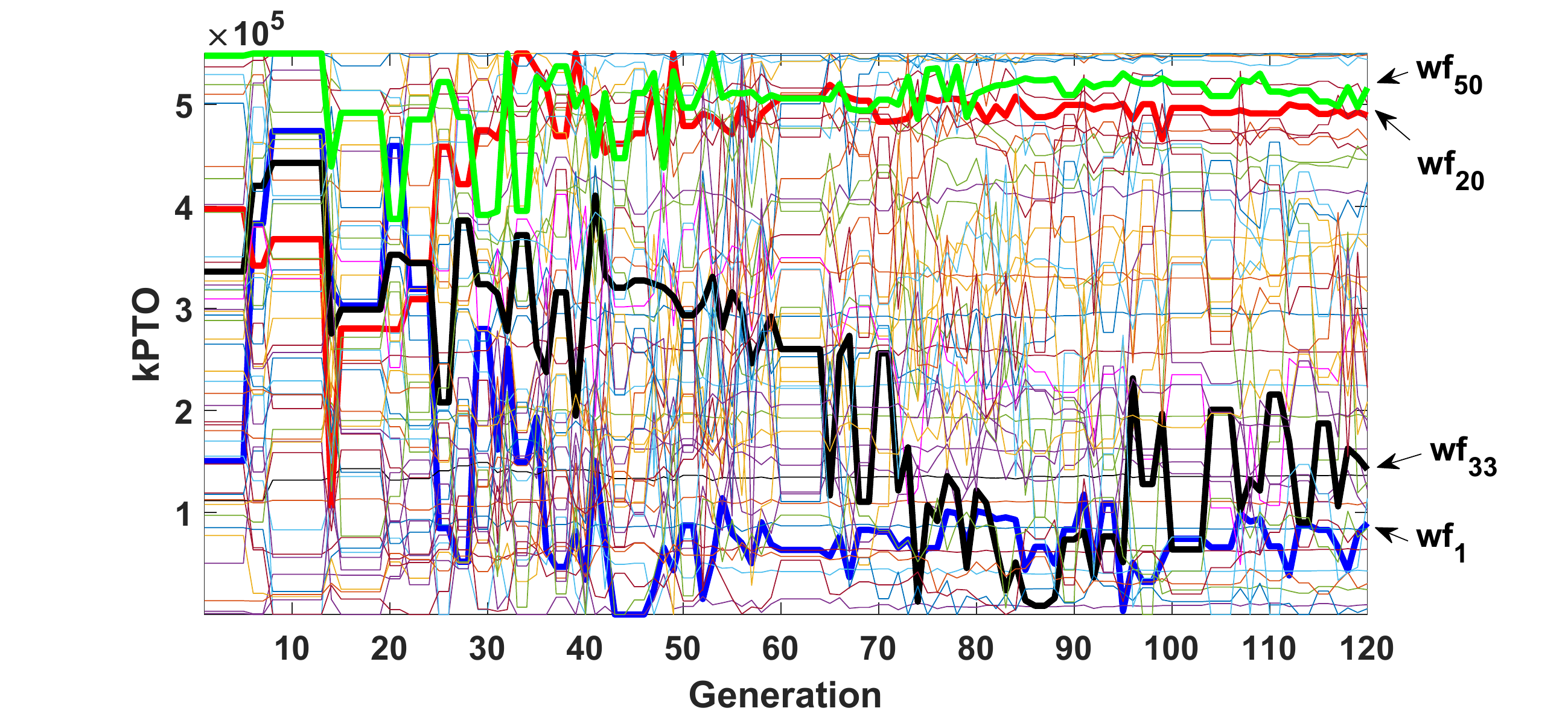}}\\
\subfloat[]{
\includegraphics[clip,width=0.8\columnwidth]{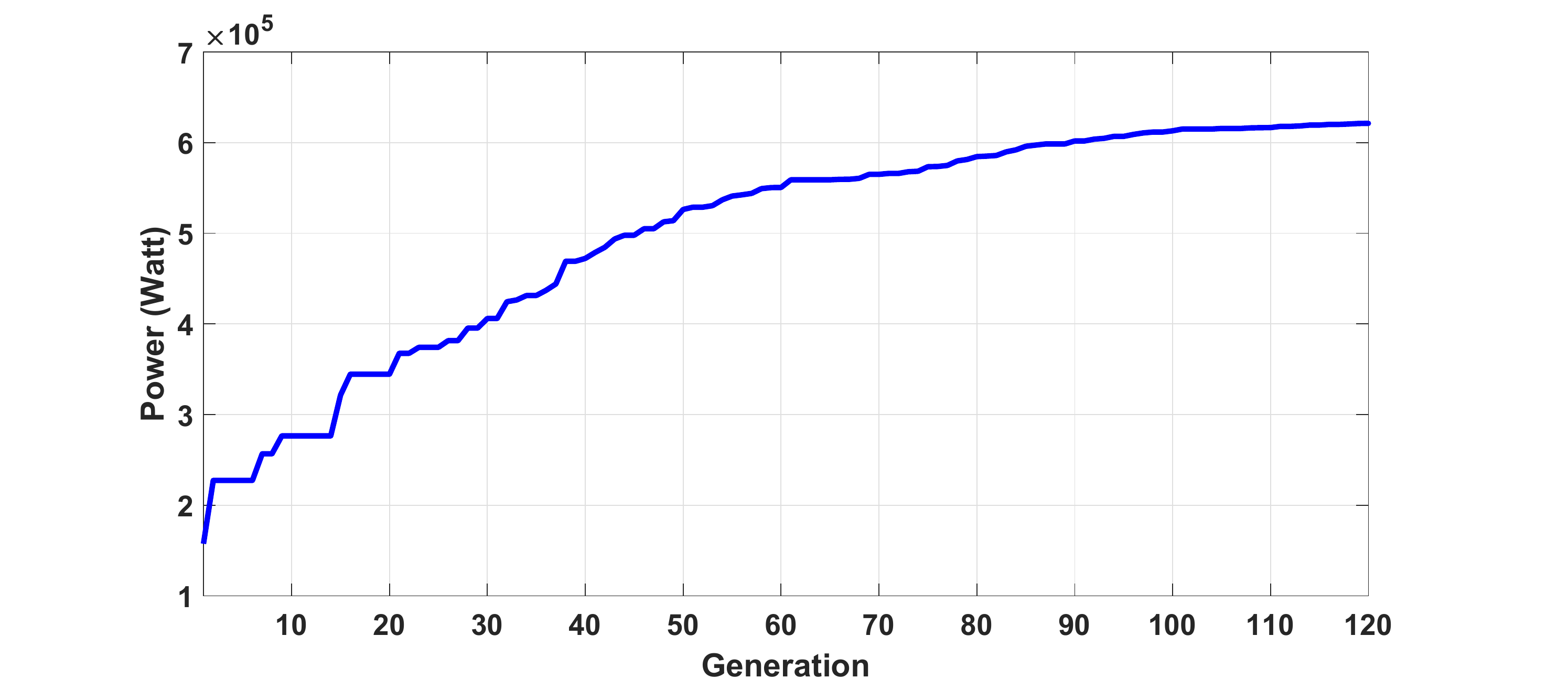}}

\caption{Damping-Spring (dPTO (a) and kPTO (b)) optimisation of one buoy by SLPSO for each wave frequency ($f_1,f_2,...,f_{50}$). The optimisation process of some chosen wave frequencies lines are marked in bold to highlight their trajectories.   }%
\label{fig:PTO-optimization}%
\end{figure}

The optimisation process of one buoy PTOs based on the 50 wave frequencies by SLPSO can be observed in Figure \ref{fig:PTO-optimization}. The PTO values fluctuate several times for each wave frequency and finally, converge to particular values. 

\subsubsection{Position analysis}
\begin{figure}
\centering
\subfloat[]{
\includegraphics[clip,width=0.49\columnwidth]{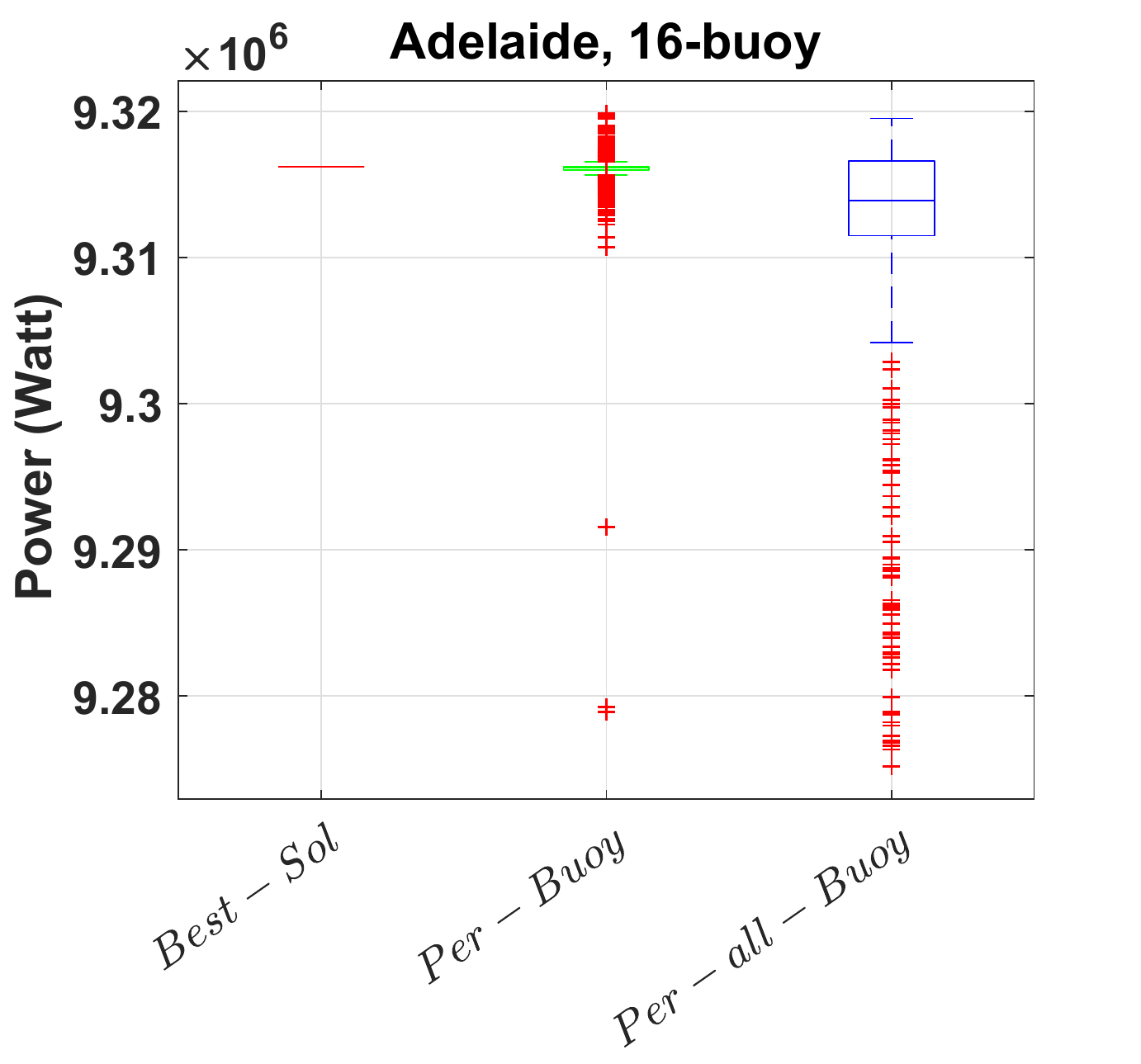}}
\subfloat[]{
\includegraphics[clip,width=0.49\columnwidth]{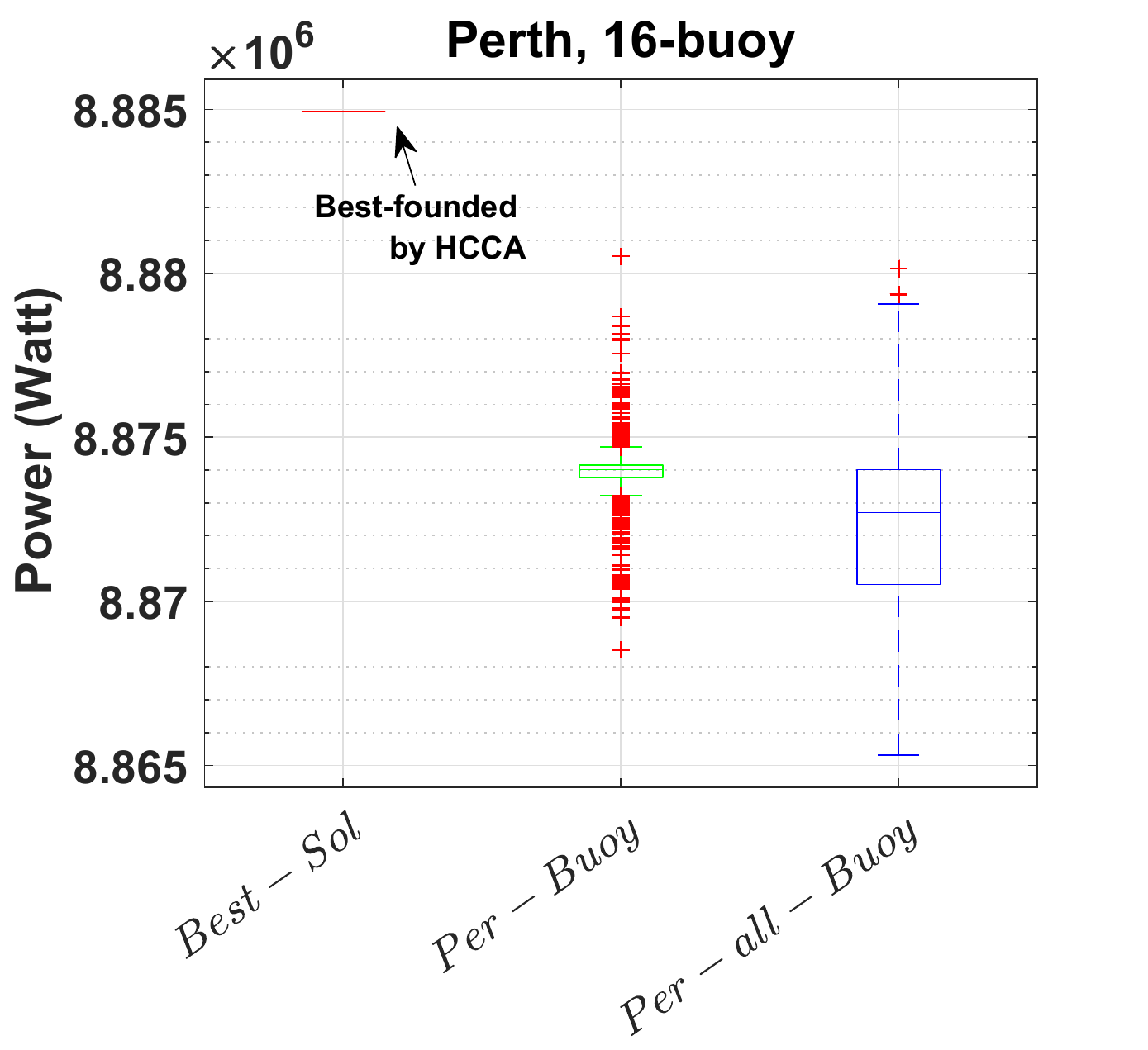}}\\
\subfloat[]{
\includegraphics[clip,width=0.49\columnwidth]{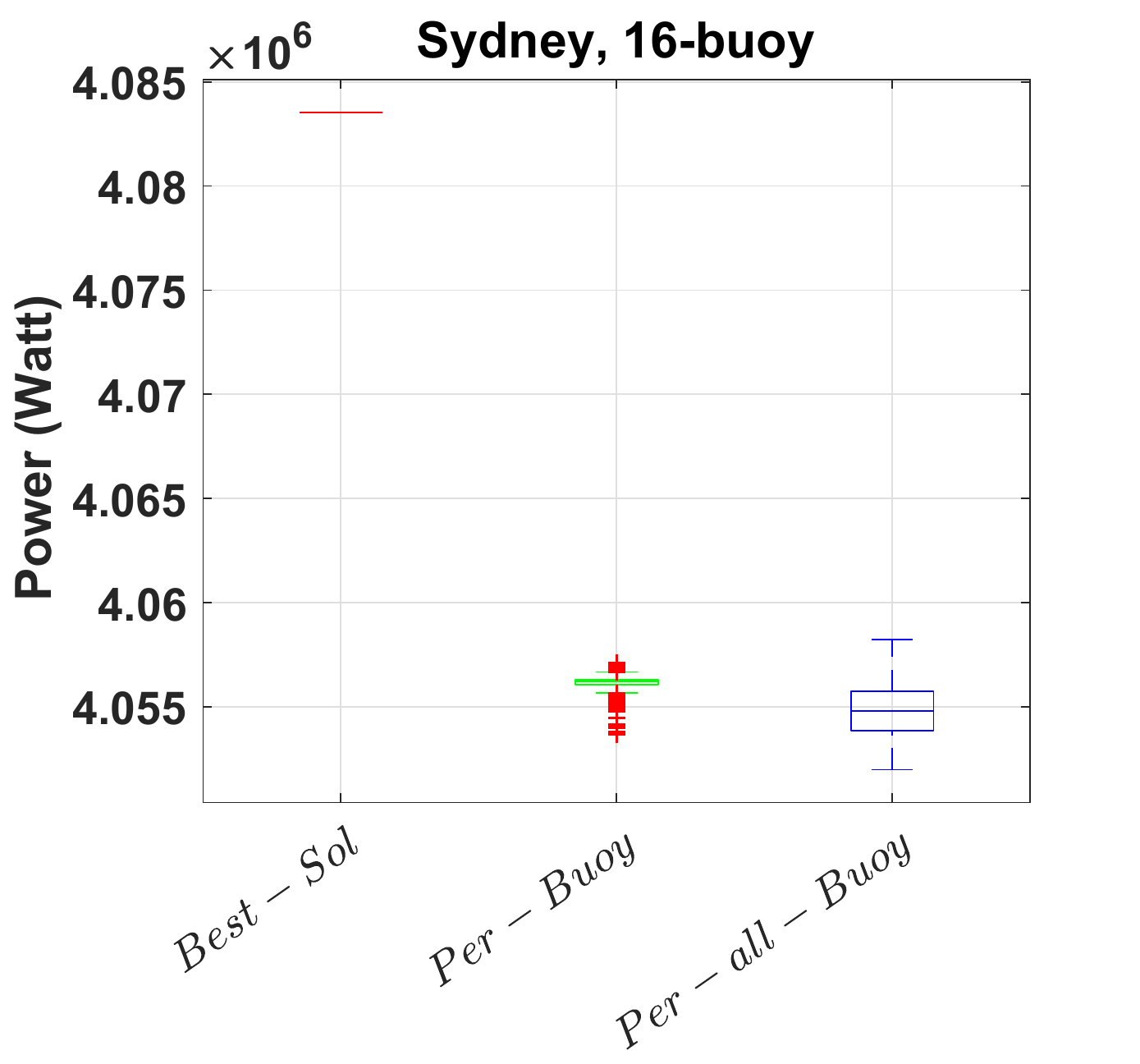}}
\subfloat[]{
\includegraphics[clip,width=0.49\columnwidth]{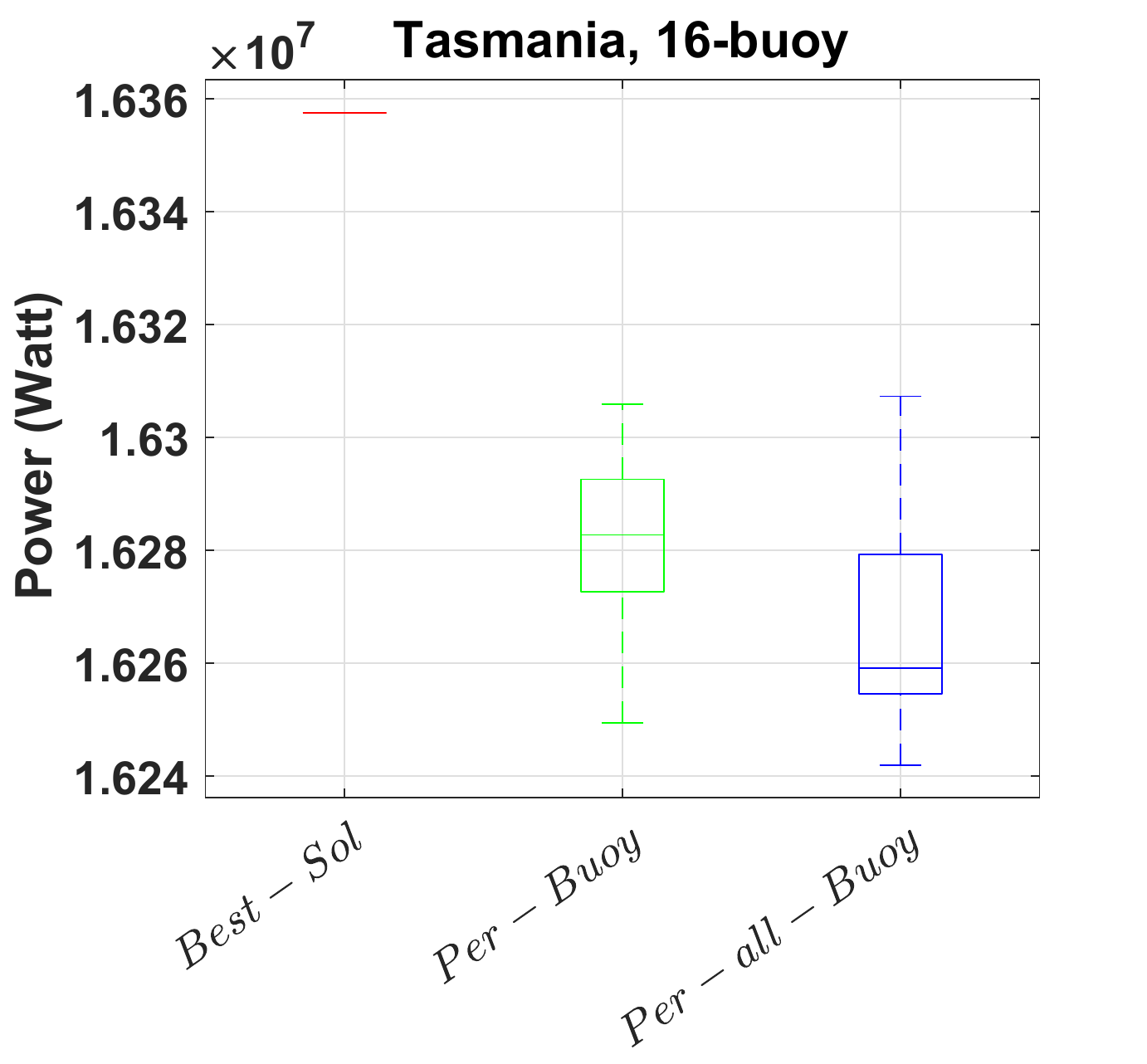}}\\
\caption{The position perturbation experimental results 16-buoy in 4 real wave models per each buoy and all buoys.}%
\label{fig:Perboxplot_16}%
\end{figure}

For evaluating the position sensitivity of the best-found 16-buoy arrangement in four real wave models, a practical experiment is done. In the first step, we perturb each generator's position by a  random variable with a normal distribution ($\mu=(x_i,y_i)$ and $\sigma=1m$) 100 times. Secondly, we perturb all buoys position by this strategy. Figure \ref{fig:Perboxplot_16} demonstrates the results of both perturbation experiments and the best 16-buoy layout power. We can see that this practical analysis is able to improve the total power output of Adelaide wave site by 0.04\%. This minor modification reveals that the proposed optimisation method (HCCA) can converge to a, locally, near-optimal configuration within the limited computational budget appropriately (in terms of position optimisation).

Moreover, for the other three wave scenarios (Sydney, Tasmania, and Perth), the position analysis experiments cannot find a better configuration than the HCCA optimisation results. The perturbation loss, respectively, for the best 16-buoy layouts power in Sydney, Tasmania and Perth wave farm are $0.67\%, 0.49\%$ and $0.12\%$ on average.  According to the results, the power outputs for the best-found layouts
are relatively insensitive to small perturbations in buoy position -- this is a good outcome in that small errors in buoy placement in a real environment are unlikely to have a major impact on power output. 
\subsection{optimisation Experiments}
 In this part, we summarize the experimental results from optimizing the layout and PTO parameters from 4-buoy arrays and then 16-buoy arrays. The 16-buoy experiments are expected to be more challenging due to the larger number of parameters and a much larger number of buoy interactions. 

\subsubsection{4-buoy layout results}
\begin{figure}
\centering
\subfloat[]{
\includegraphics[clip,width=0.49\columnwidth]{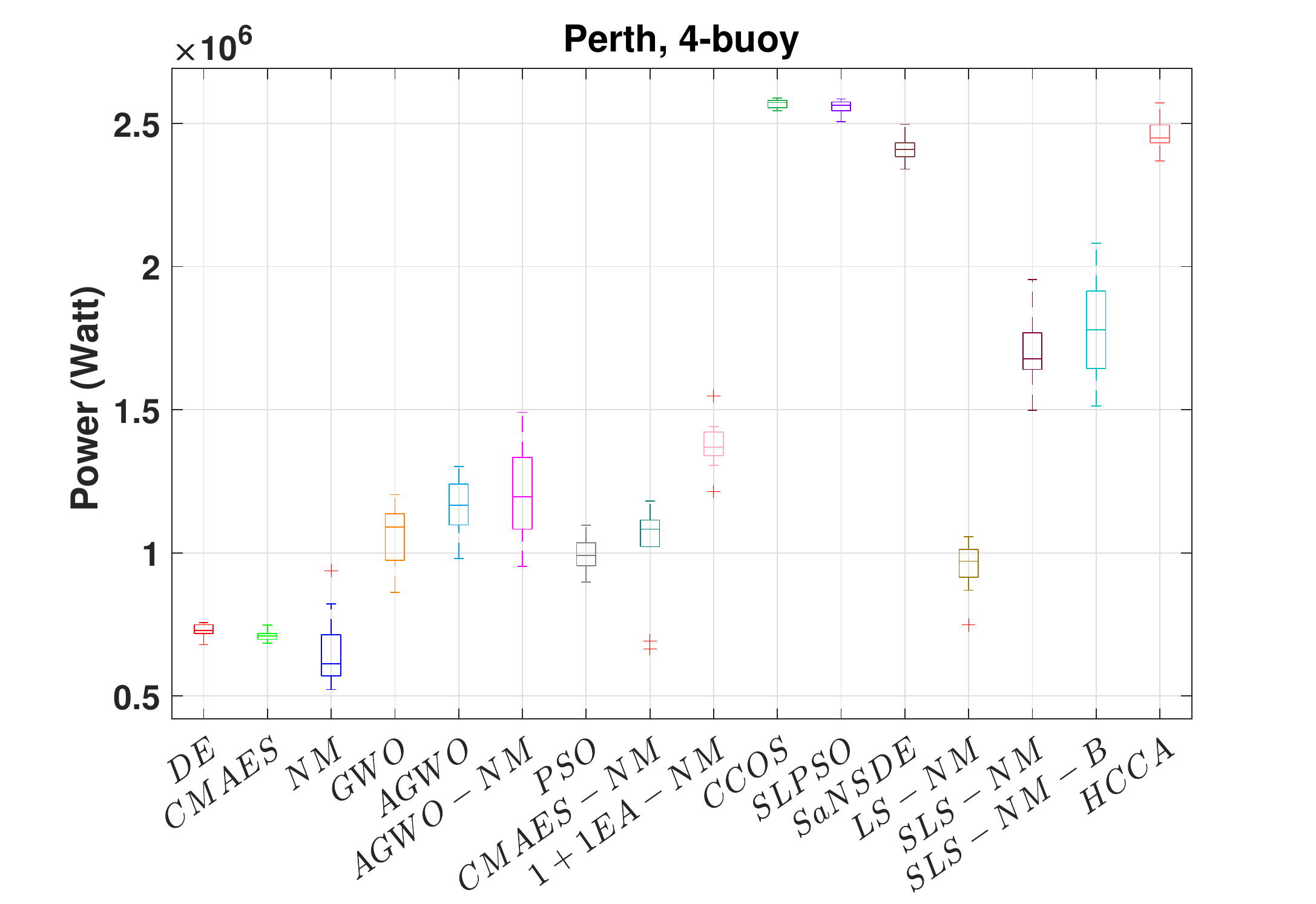}}
\subfloat[]{
\includegraphics[clip,width=0.49\columnwidth]{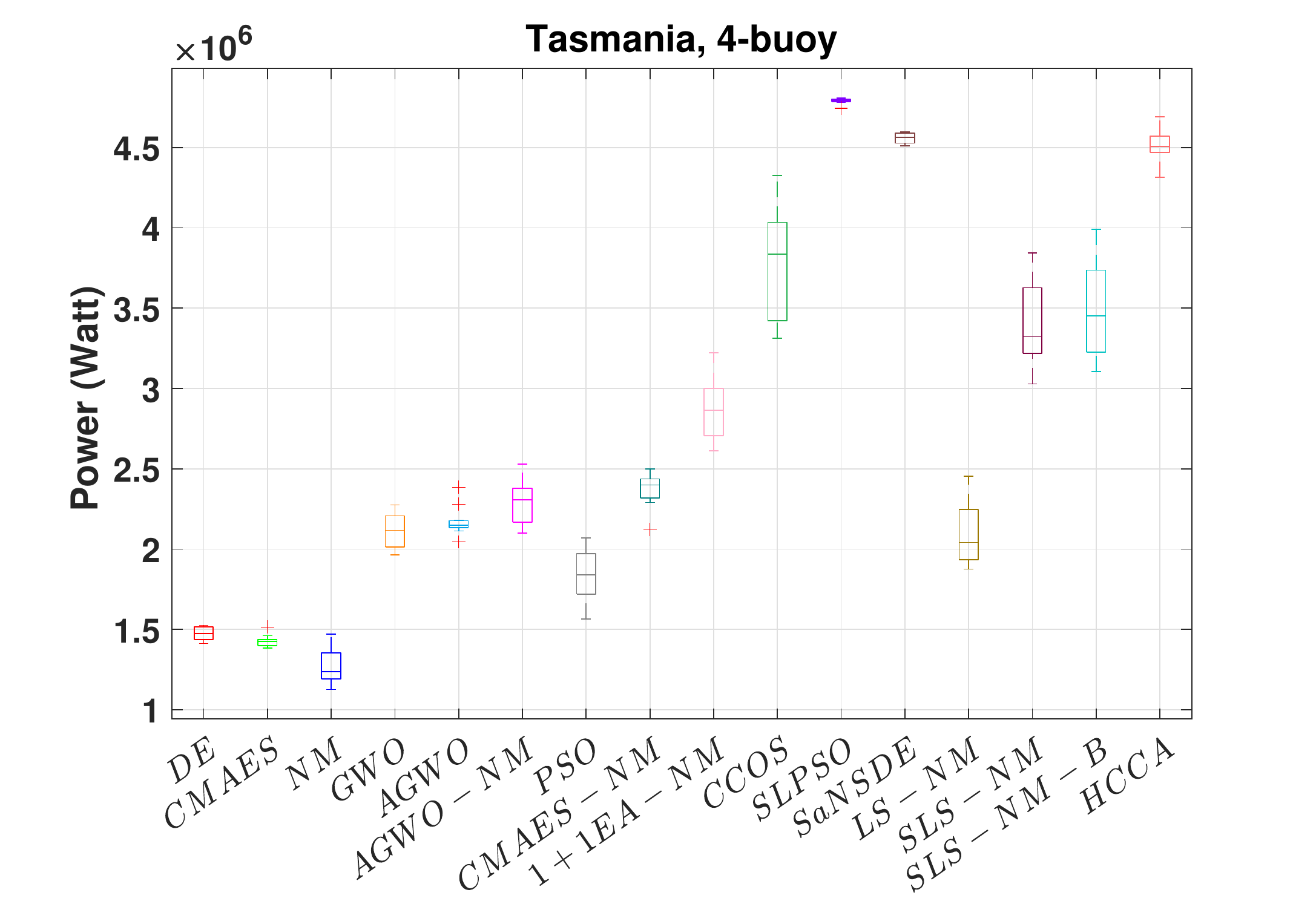}}\\
\subfloat[]{
\includegraphics[clip,width=0.49\columnwidth]{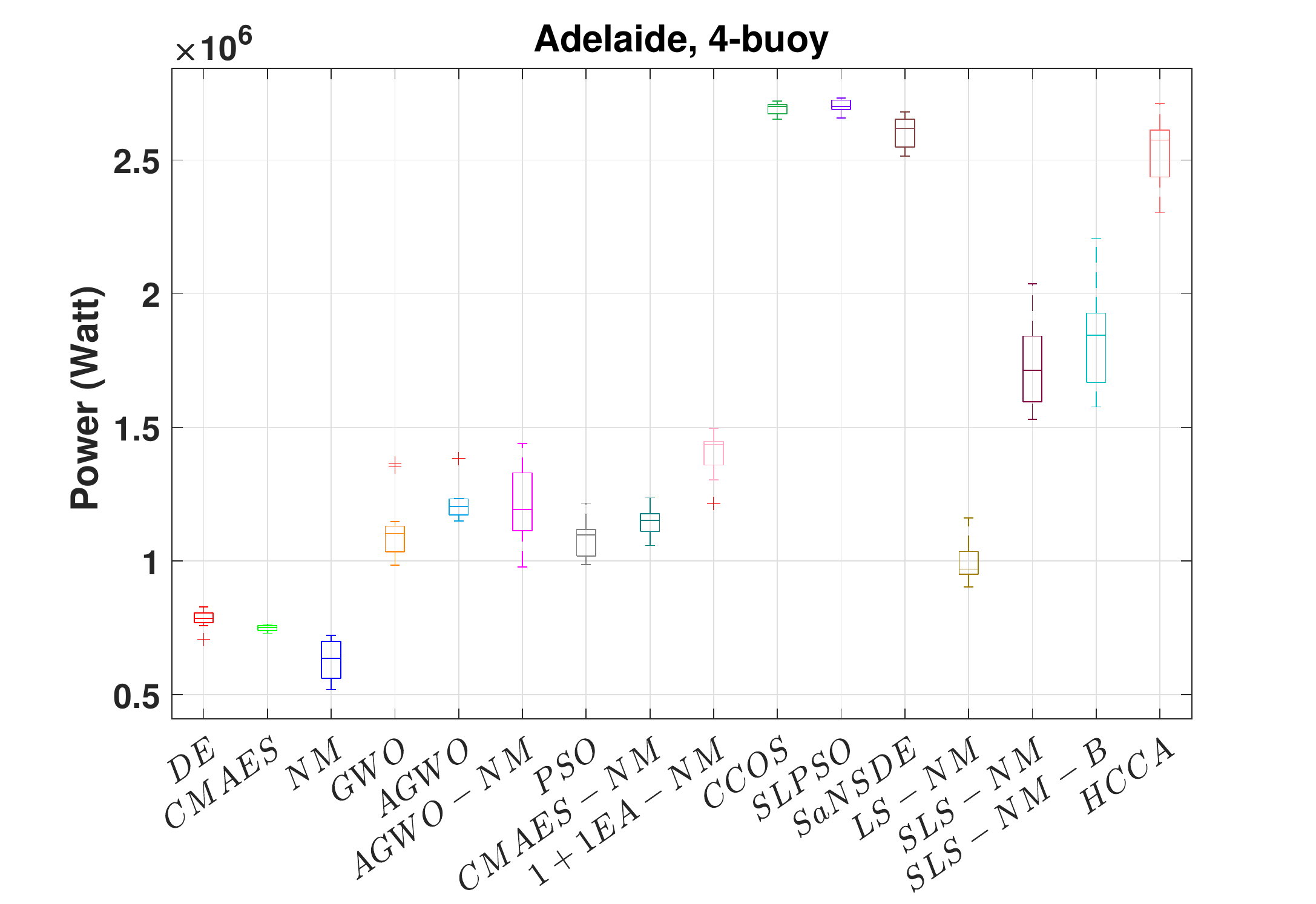}}
\subfloat[]{
\includegraphics[clip,width=0.49\columnwidth]{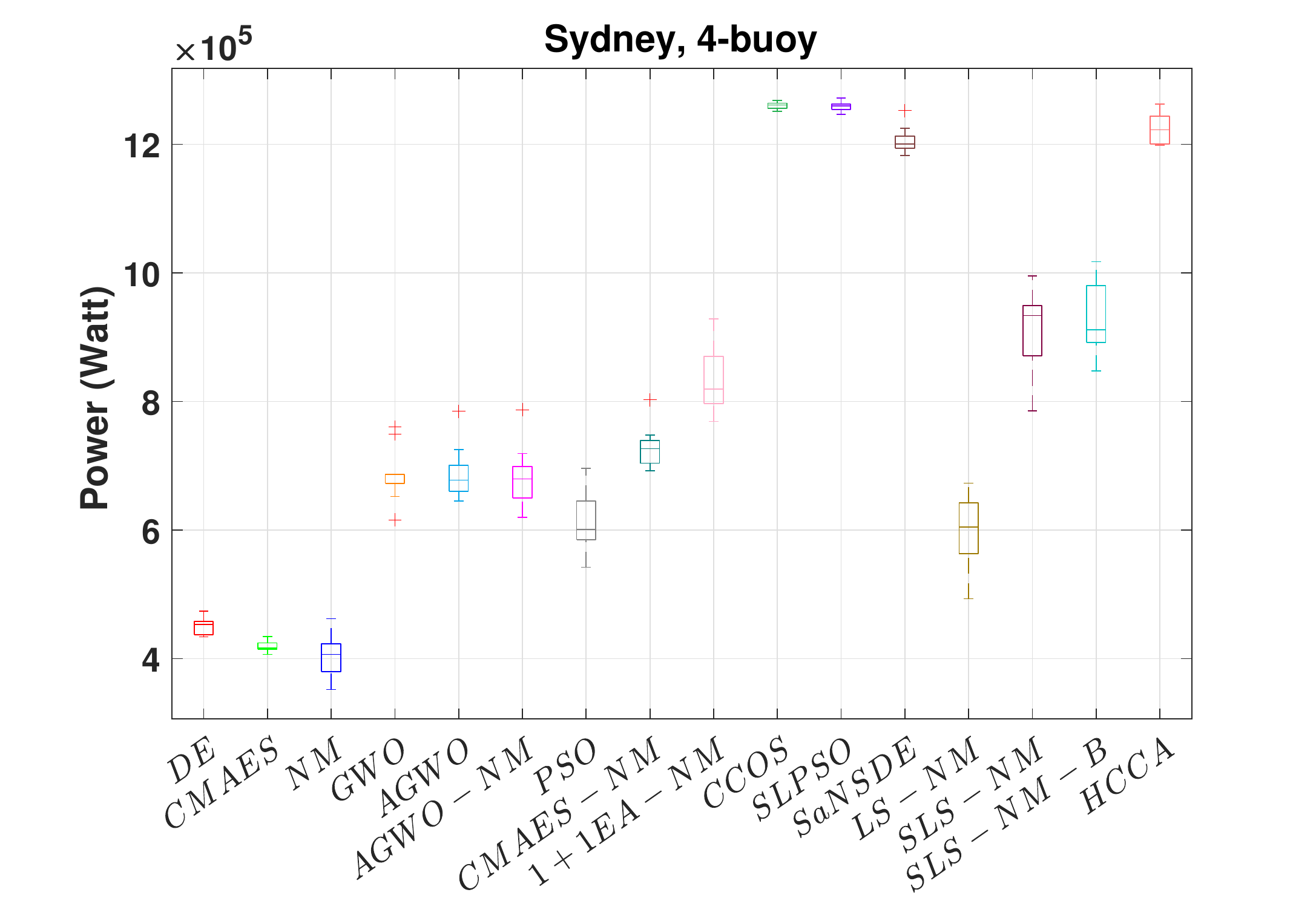}}
\caption{The comparison of the proposed algorithms' performance for 4-buoy layouts in four real wave models. The optimisation results present the best solution per experiment. (10 independent runs per each method)}%
\label{fig:boxplot_4}%
\end{figure}

In the experiments on 4-buoy layouts (medium-scale optimisation problem), 15 representative meta-heuristic algorithms are compared with the HCCA including five well-know EAs plus a new adaptive version of GWO, six cooperative EAs and two-hybrid heuristic approaches in four real wave scenarios.  The parameter settings for these meta-heuristic variants are summarized in Table \ref{details:OP}.  The termination condition of each method is reached when three days are exhausted (with 12 cores in parallel). Figure~\ref{fig:boxplot_4} presents the box-and-whiskers plot for the best-found 4-buoy configurations which produce the maximum power output for each run for 16 search heuristics for four real wave models.  It can be seen that the performance of cooperative co-evolution strategies (CCOS,  SLPSO$_{II}$ and SaNSDE$_{II}$ and the new hybrid method are considerably better than other applied meta-heuristic algorithms. The next best performances are exhibited by both SLS-NM and SLS-NM-B; however, the average absorbed power by these methods is less than the CC and HCCA approaches by $25\%$. 

Looking more closely at Table~\ref{table:allresults4}, we can observe the highest absorbed power of 4-buoy layouts are  found by SLPSO$_{II}$, CCOS and HCCA, respectively. These optimisation results are closely followed by the SaNSDE$_{II}$ algorithm. These competitive performances are supported by the statistical test results, ranked using the non-parametric Friedman test, shown in Table \ref{table:Friedman}.  It is noteworthy to note that among the five optimisation methods in the all-at-once strategy, the GWO and AGWO performances are substantially better than others.  

\begin{figure}
\centering
\subfloat[]{
\includegraphics[clip,width=0.49\columnwidth]{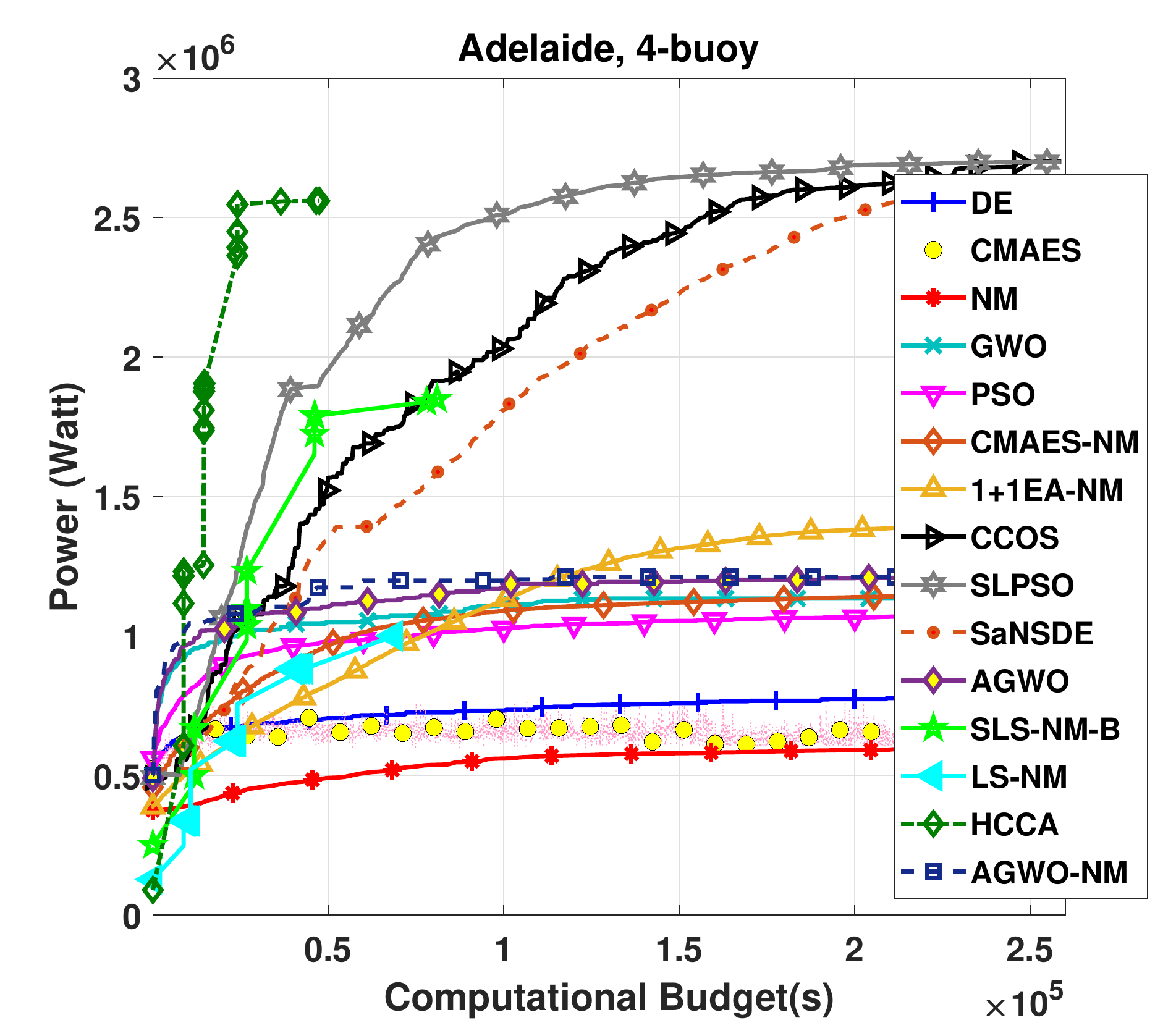}}
\subfloat[]{
\includegraphics[clip,width=0.49\columnwidth]{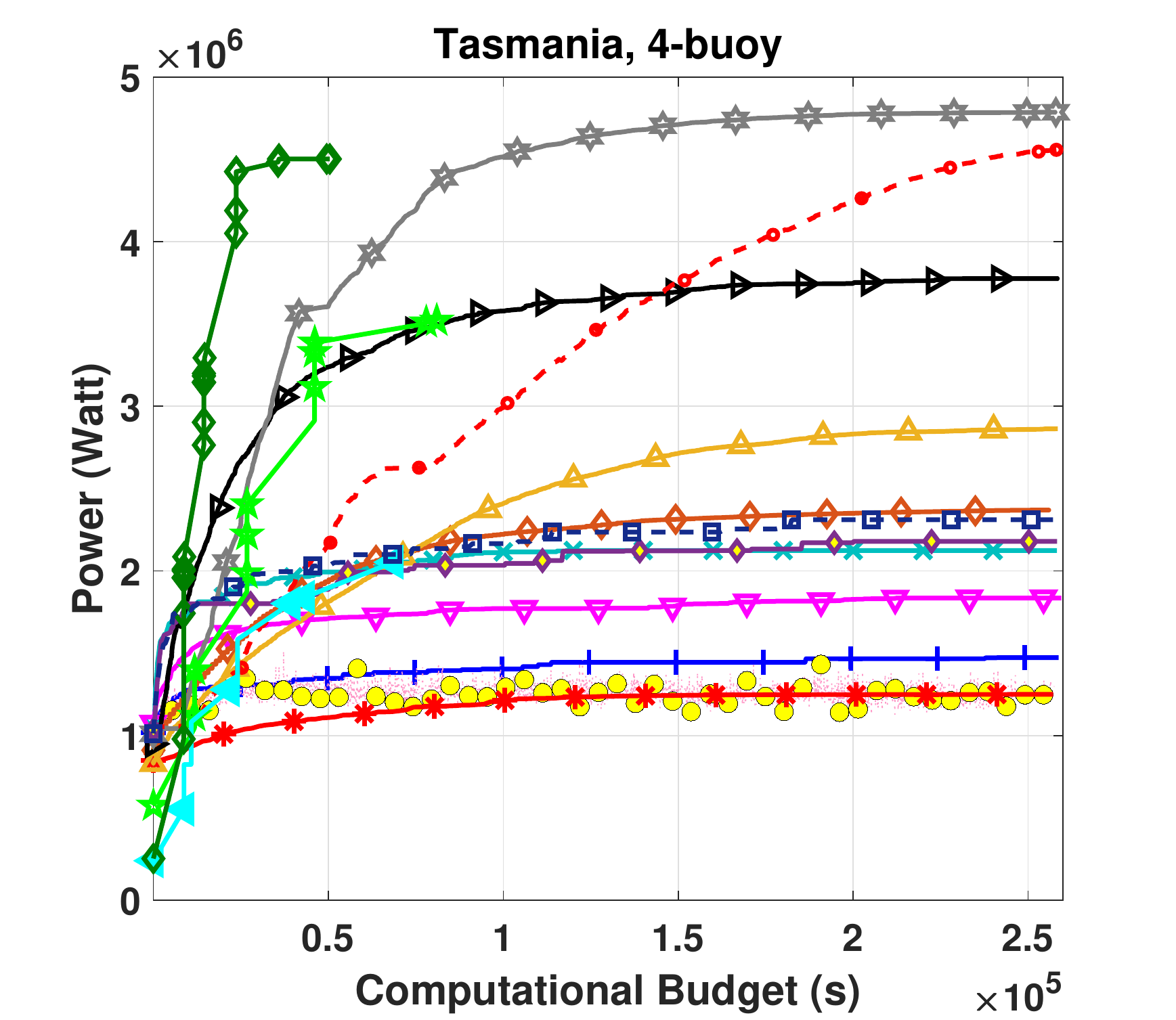}}\\
\subfloat[]{
\includegraphics[clip,width=0.49\columnwidth]{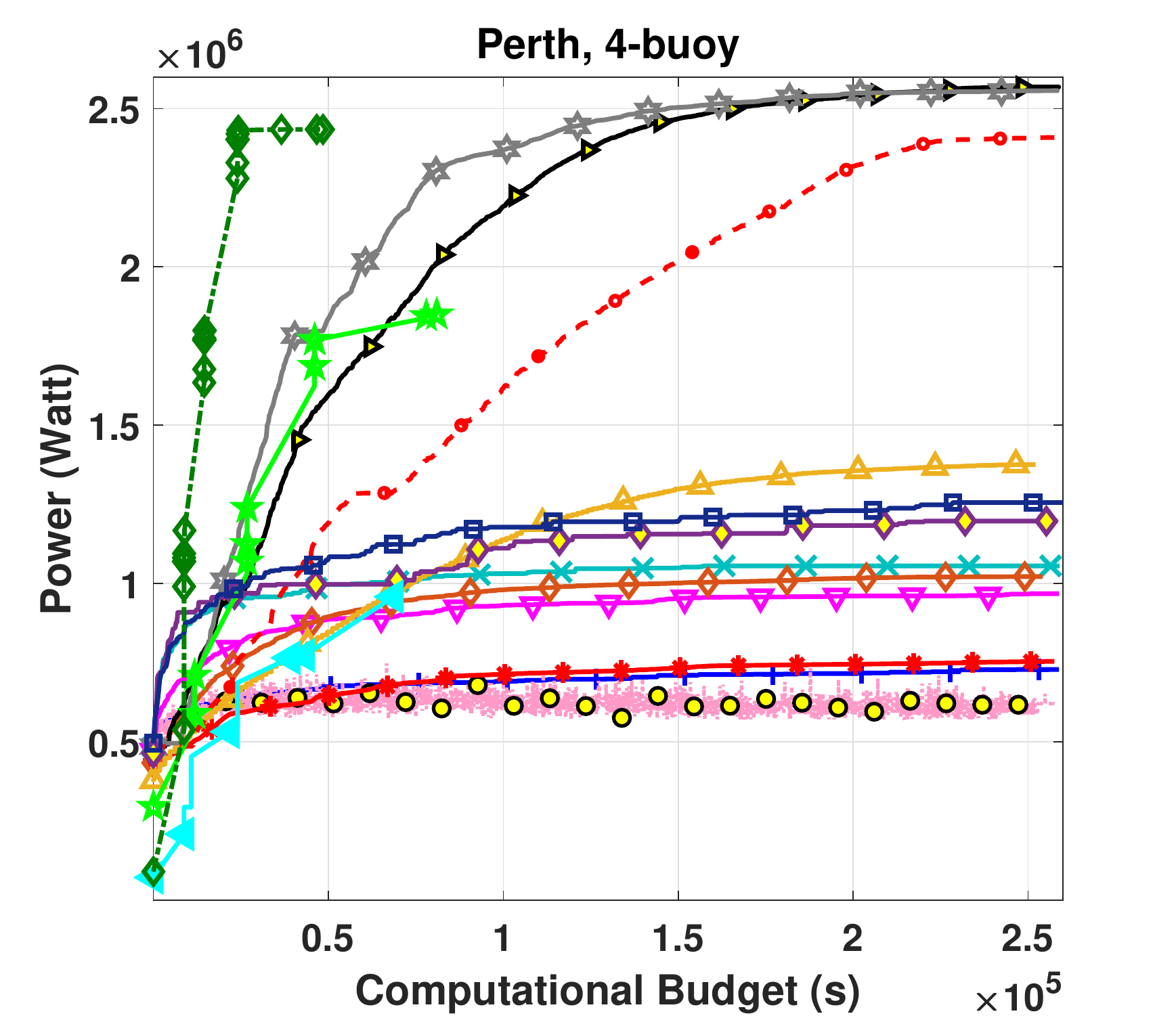}}
\subfloat[]{
\includegraphics[clip,width=0.49\columnwidth]{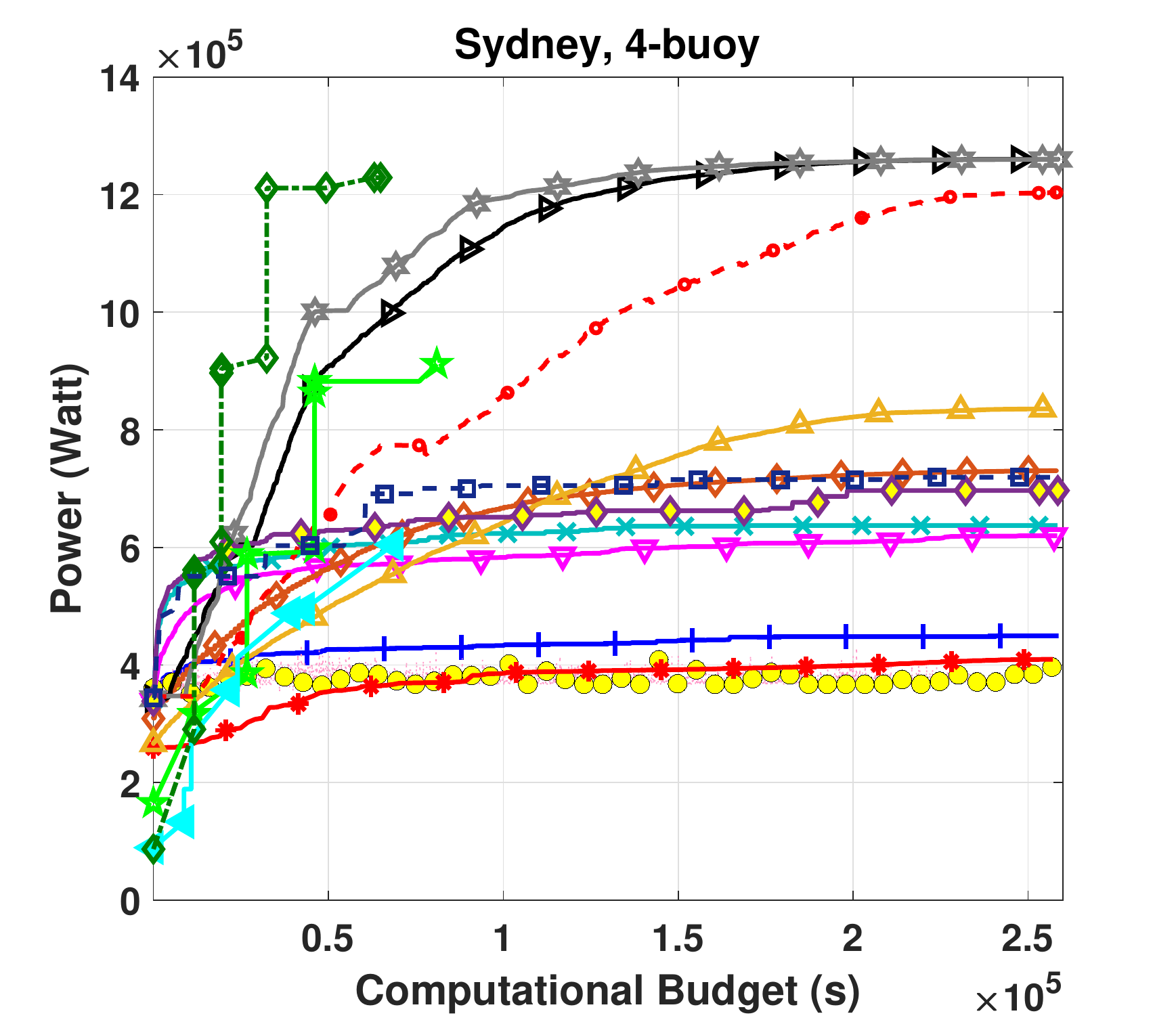}}\\

\caption{Comparison of algorithms' effectiveness and convergence rate for 4-buoy layouts in four real wave scenarios. }%
\label{fig:con_4_plot}%
\end{figure}

Viewing  the convergence curves (Figure \ref{fig:con_4_plot}) from this experiment (4-buoy) in four real wave models, it is clear that the HCCA framework converges faster for the 4-buoy layout than other search methods. In fact, HCCA improves beyond the power outputs achieved by other methods when it has consumed just $20\%$ of its 3-day computational budget. 


\subsubsection{16-buoy layout results}
As evaluating one 16-buoy layout is ten times more expensive than a 4-buoy layout, optimizing such large wave farms is a challenging problem. According to the statistical results of Table \ref{table:allresults16}, we see that the average 16-buoy layouts power output which are found by HCCA  is increased substantially to $80\%$ more than previous research outcomes (SLS-NM-B \cite{Neshat:2019:HEA:3321707.3321806}) in all wave scenarios. 
Table \ref{table:Friedman} presents the average rank of all heuristic methods for 16-buoy experiments, that HCCA, SLS-NM-B and SLS-NM have the highest rank, respectively. 

\begin{figure}
\centering
\subfloat[]{
\includegraphics[clip,width=0.49\columnwidth]{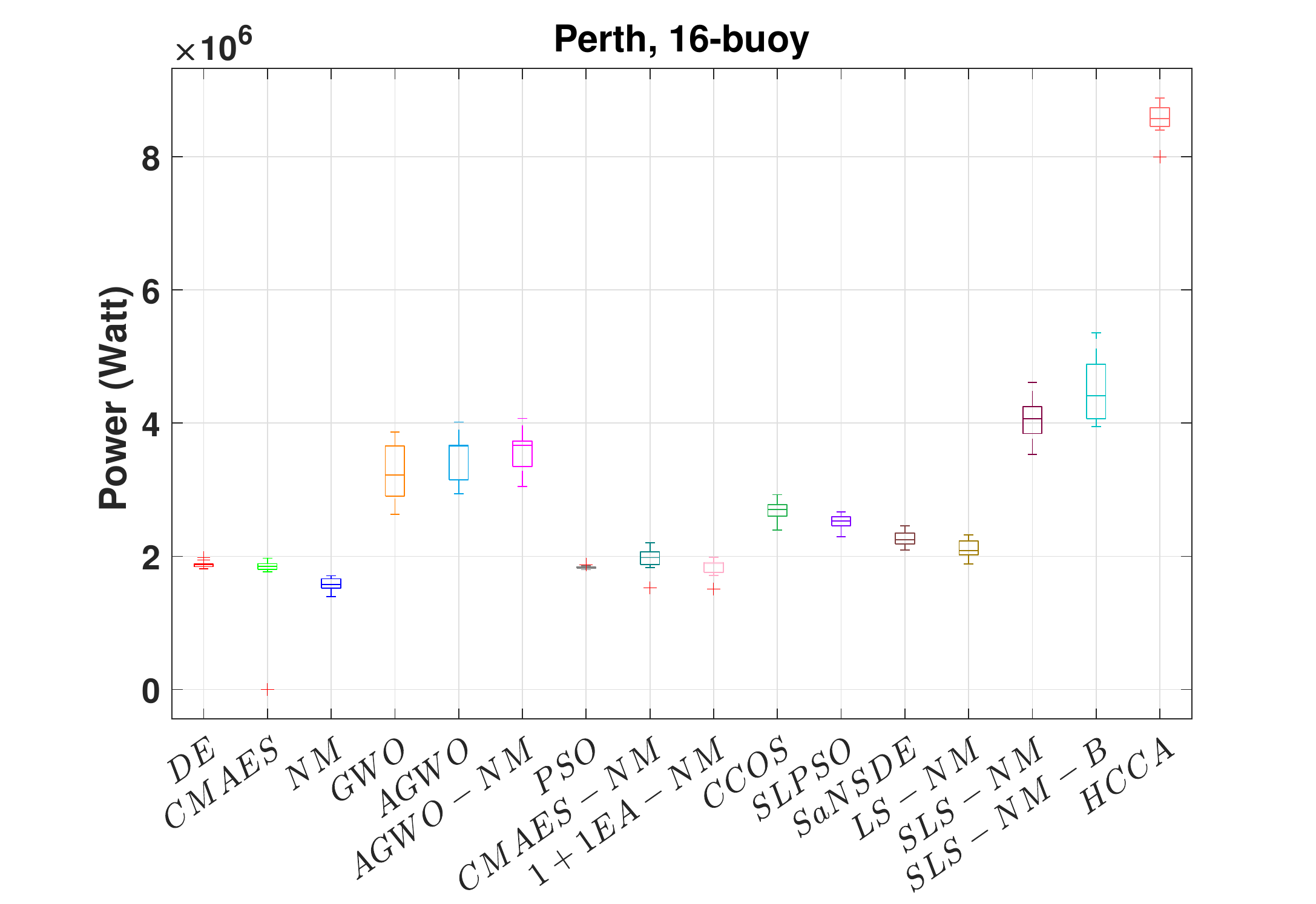}}
\subfloat[]{
\includegraphics[clip,width=0.49\columnwidth]{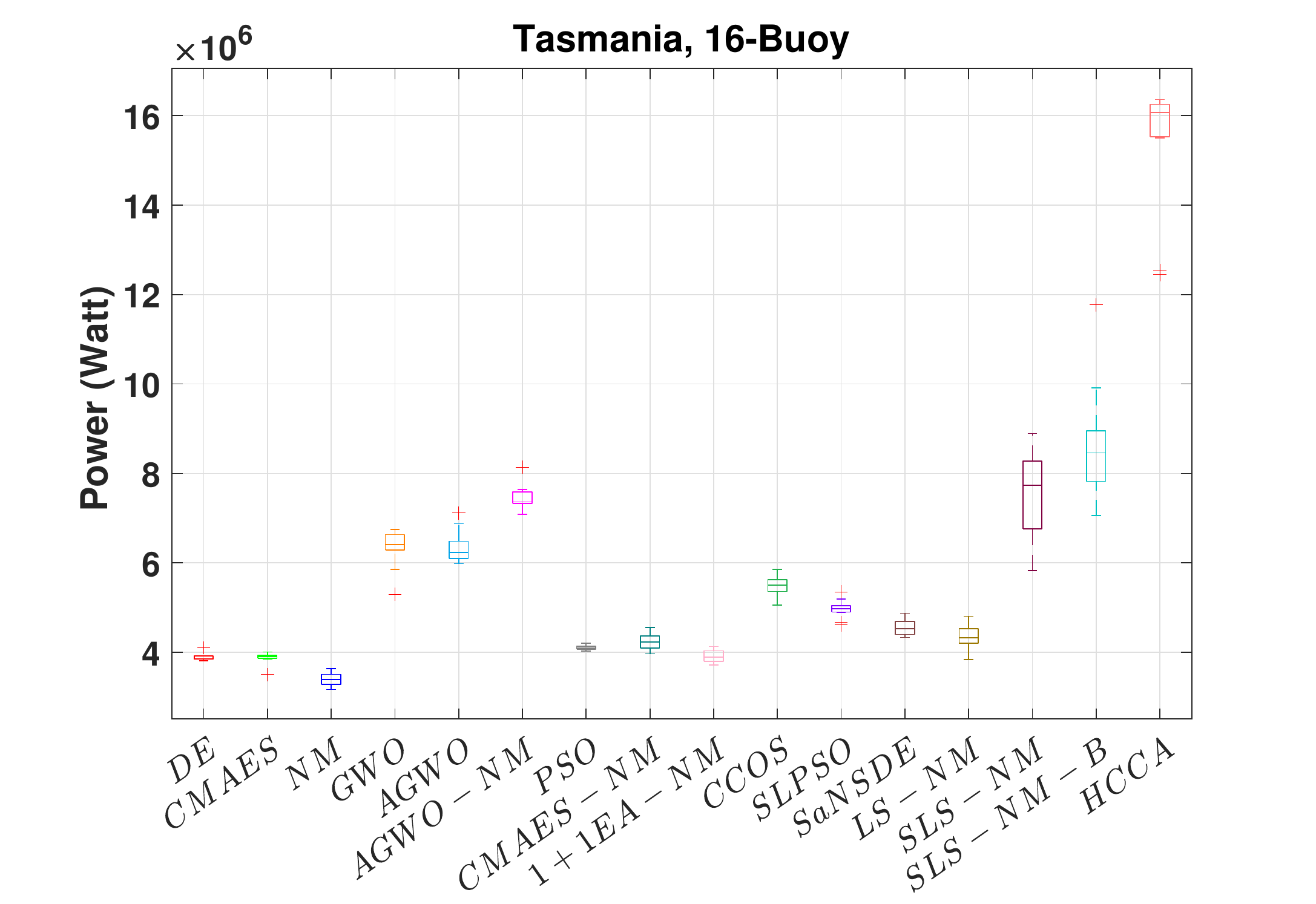}}\\
\subfloat[]{
\includegraphics[clip,width=0.49\columnwidth]{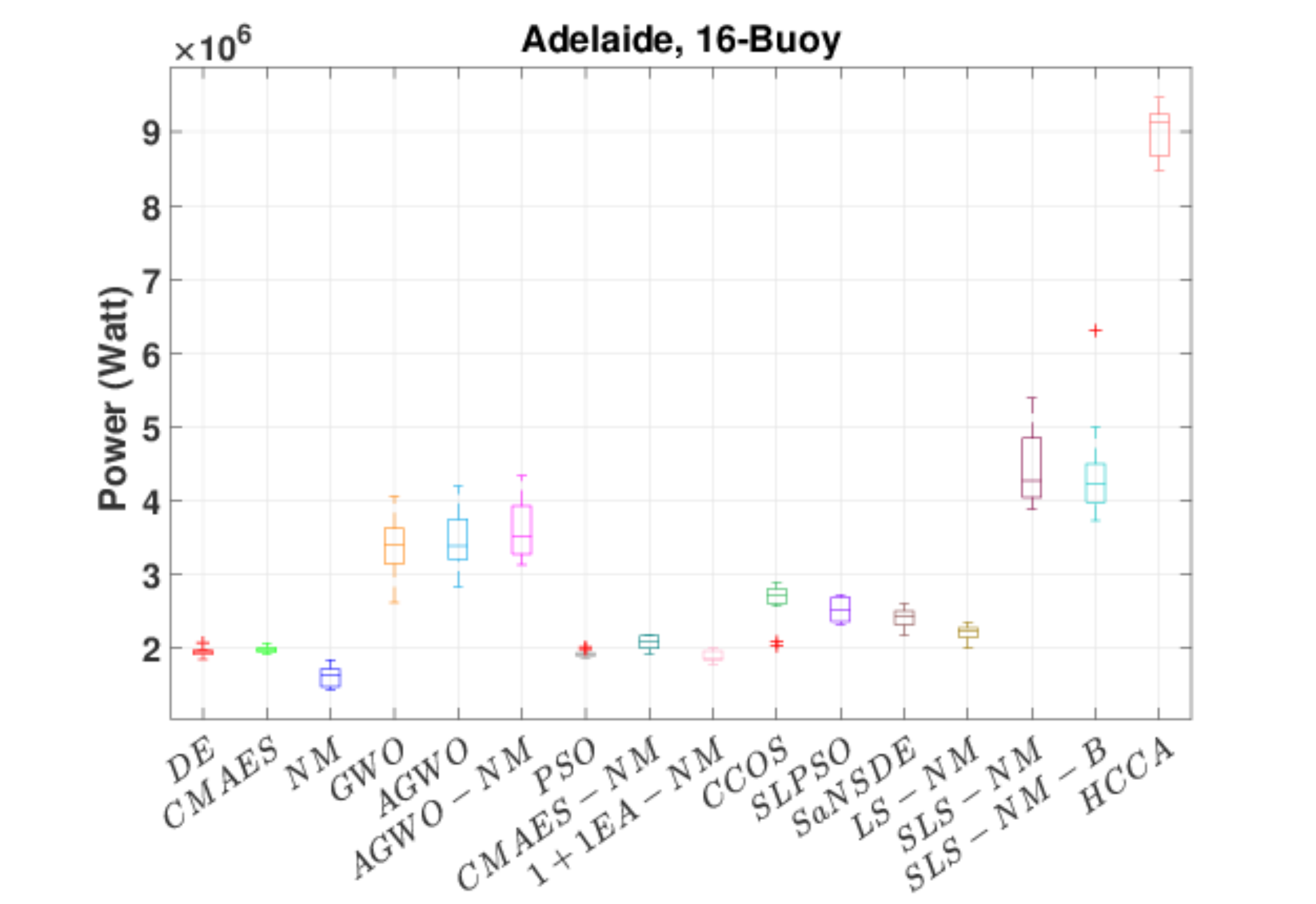}}
\subfloat[]{
\includegraphics[clip,width=0.49\columnwidth]{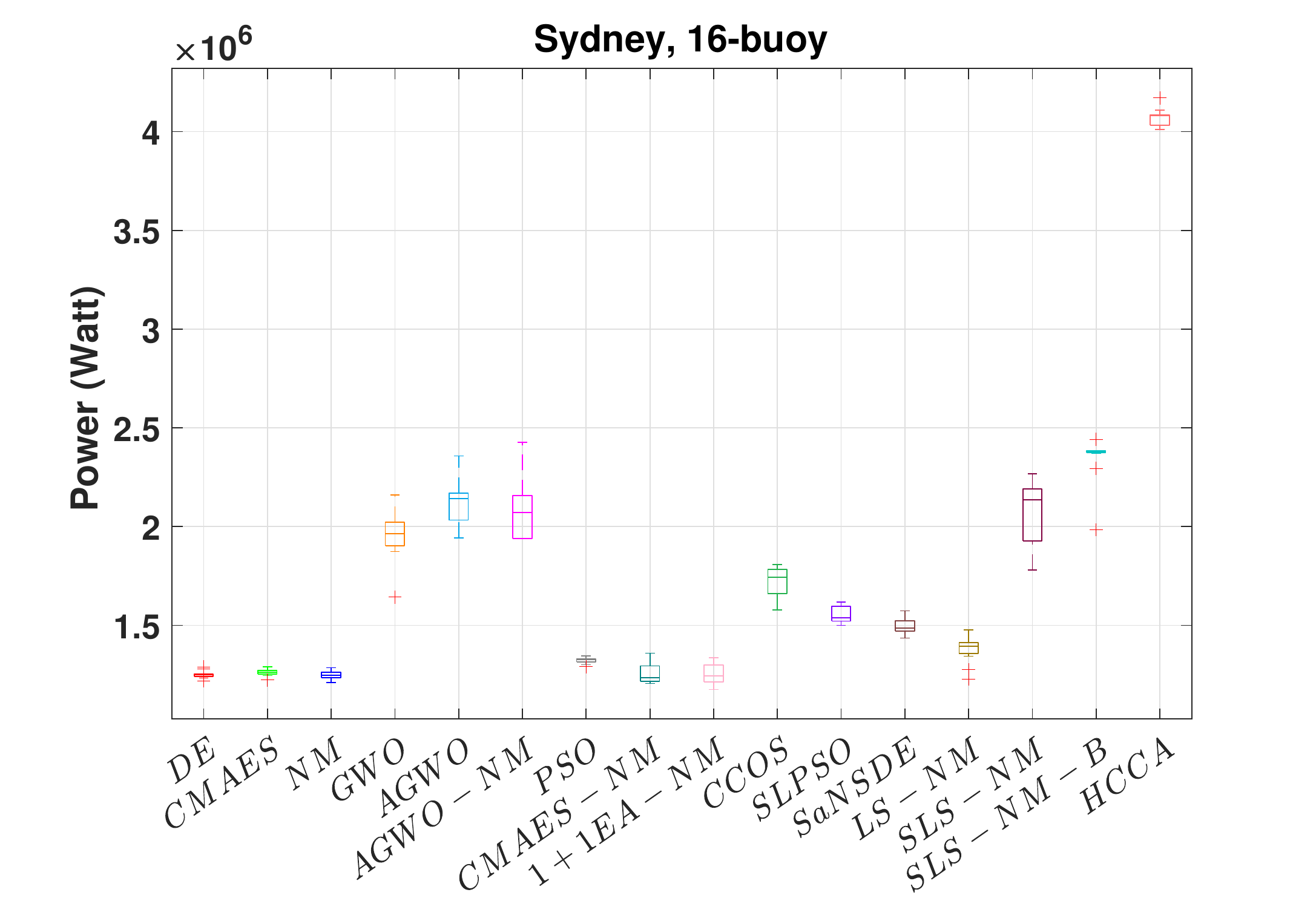}}\\
\caption{The comparison of the proposed algorithms' performance for 16-buoy layouts in four real wave model. The optimisation results present the best solution per experiment. (10 independent runs per each method)}%
\label{fig:boxplot_16}%
\end{figure}

The best-found 16-buoy layouts power output for each run of all heuristic methods are plotted as a box plot by Figure~\ref{fig:boxplot_16}. Figure~\ref{fig:boxplot_16} shows that HCCA performs much better than other optimisation algorithms, as mentioned before. After SLS-NM and SLS-NM-B, The efficiency of the AGWO-NM, AGWO and GWO are competitive compared with other cooperative and generic EAs. The primary reason is derived from the robust exploitation and exploration capability of GWO for PTO parameter optimisation and having good performance for high-dimensional problems. 
It is noteworthy that the CC frameworks (CCOS, SLPSO$_{II}$ and SaNSDE$_{II}$) are not shown to be
highly effective. This may be because the CC framework is not equipped with the systematic position optimisation mechanism (SLS) used by some of the one-at-a-time placement algorithms.

\begin{figure}
\centering
\subfloat[]{
\includegraphics[clip,width=0.49\columnwidth]{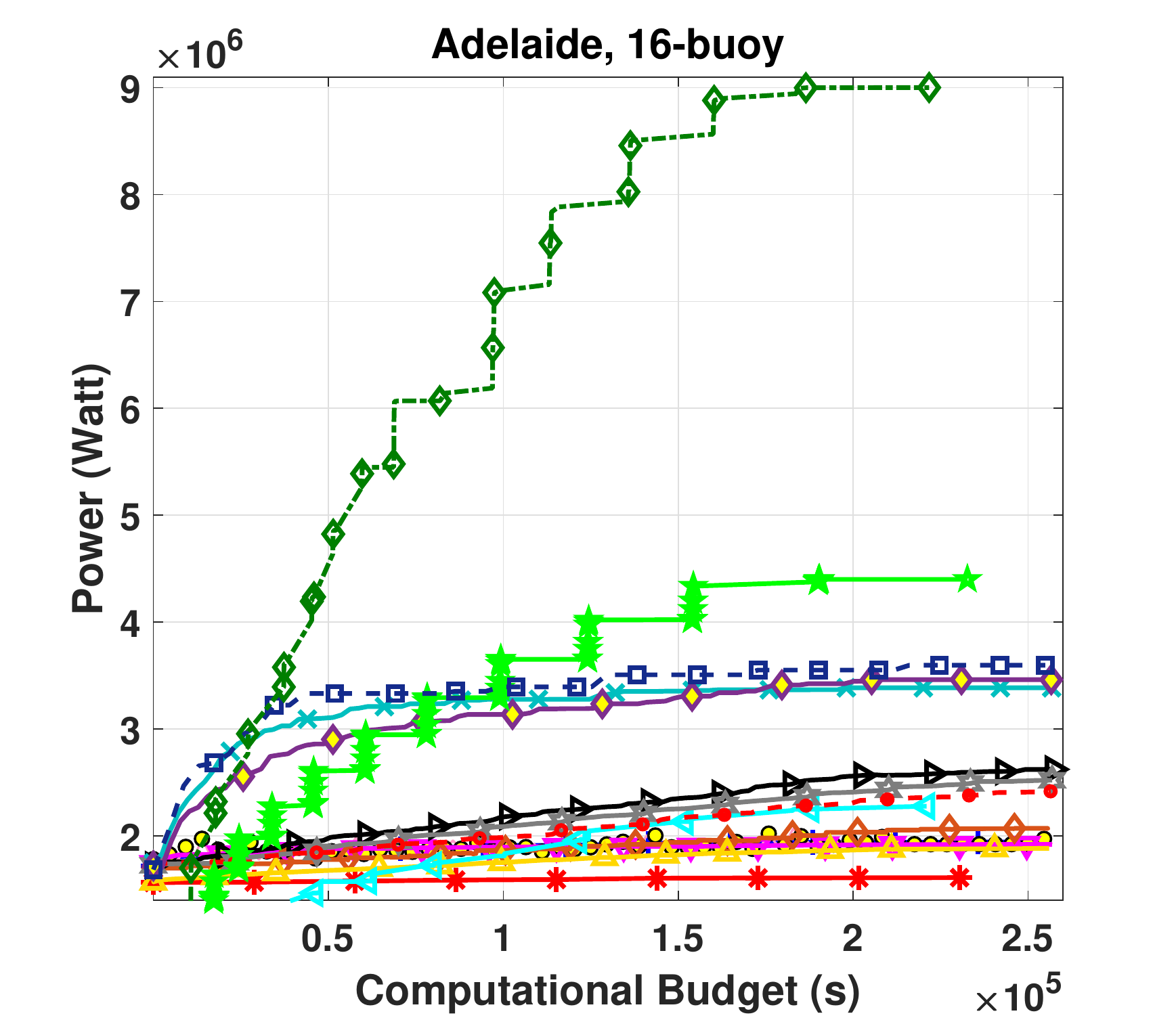}}
\subfloat[]{
\includegraphics[clip,width=0.49\columnwidth]{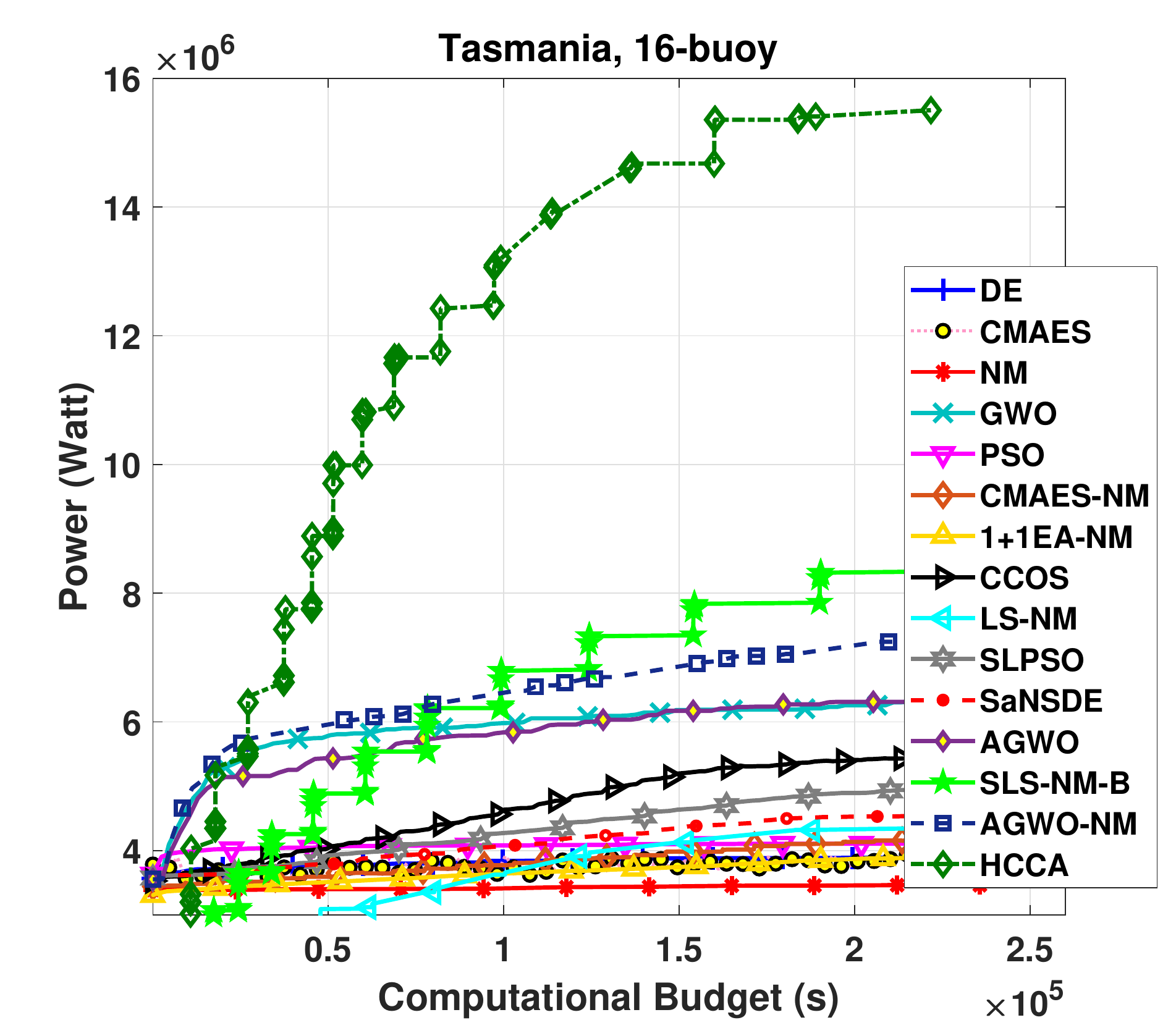}}\\
\subfloat[]{
\includegraphics[clip,width=0.49\columnwidth]{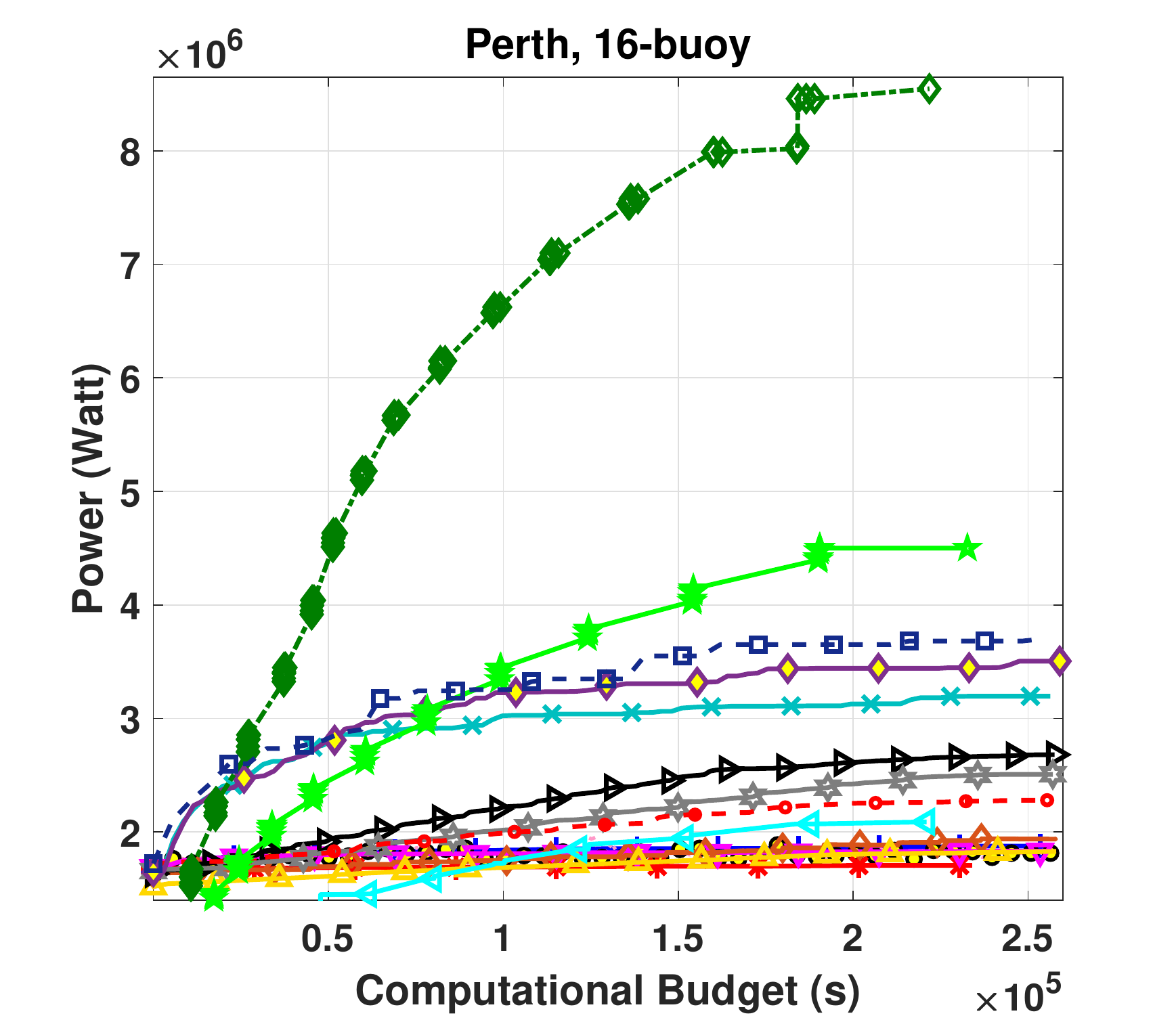}}
\subfloat[]{
\includegraphics[clip,width=0.49\columnwidth]{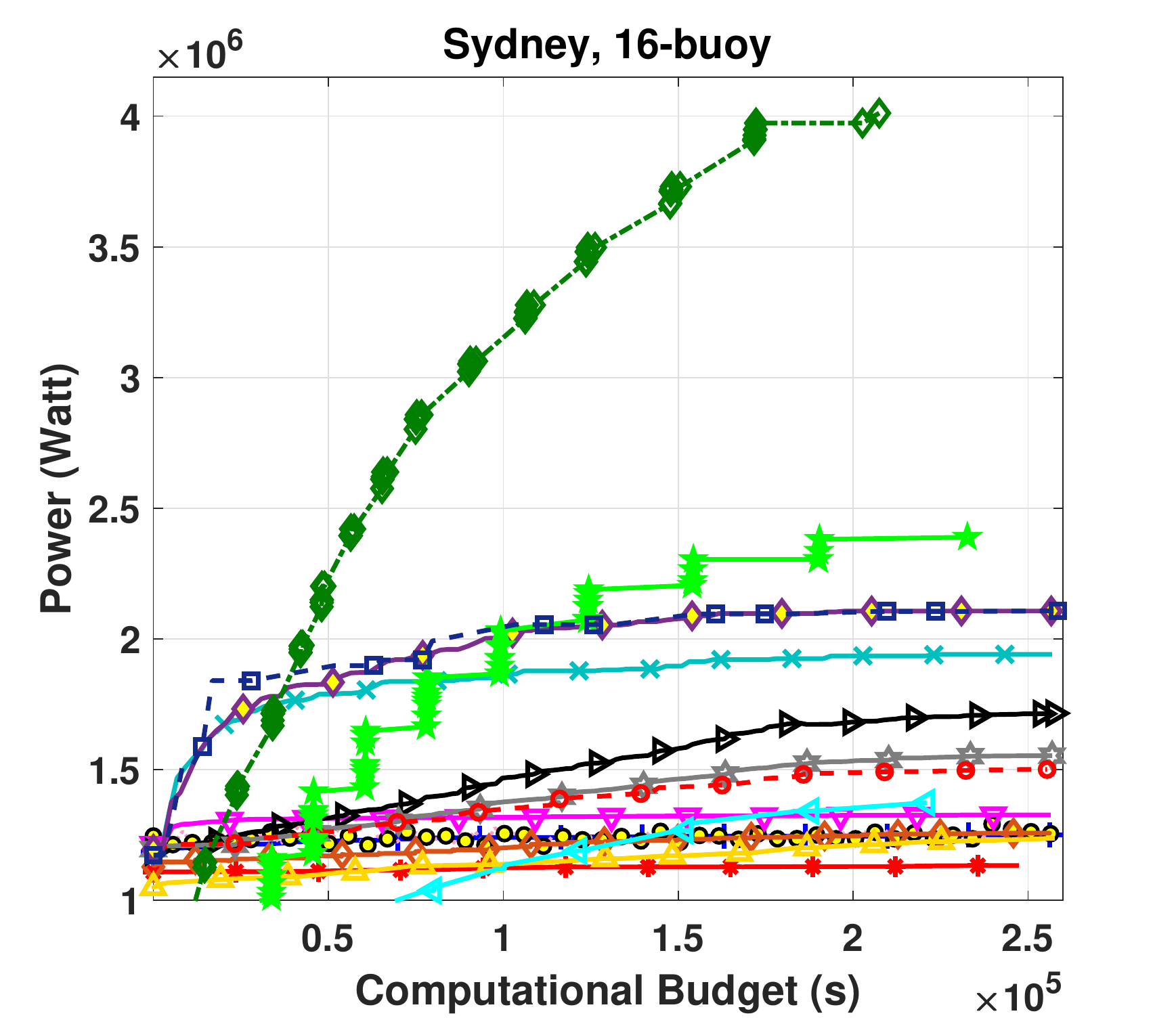}}\\

\caption{Comparison of  algorithms' performance and convergence rate for 16-buoy layouts in four real wave scenarios. }%
\label{fig:con_16_plot}%
\end{figure}

Figure \ref{fig:con_16_plot} illustrates the convergence rate of proposed methods experiments during the three-day runtime budget. It is observed that in the initial hours, GWO and their modified versions rapidly converge to effective configurations; however, they could not keep this upward trend and converge toward locally optimal settings. HCCA, clearly, has the fastest convergence speed in the four-wave models. It can be observed that SLS-NM-B is able to converge to a reasonable configuration, but that is not comparable to HCCA's achievements because of the low efficiency of Nelder-Mean optimizing the large-scale PTO parameter part of the problem. Another important observation is that the CC approaches seem to suffer from premature convergence. In addition, they also appear to be not fast enough for such expensive optimisation problems which allow just a few thousand full evaluations ($3\times10^3$). 
The best 16-buoy layouts of the nominated five methods among all heuristics can be shown in Figure~\ref{fig:allbest}, including HCCA, CMA-ES, AGWO-NM, SLPSO$_{II}$ and LS-NM in four real wave scenarios. In terms of position optimisation, it is clearly observed that HCCA is able to adjust the position of each generator successfully, which leads to a distinctive pattern of one or more rows, roughly aligned with the norm of the dominant wave direction, for placing the 16 buoys. This pattern is associated with a fast position optimisation mechanism of SLS-NM \cite{Neshat:2019:HEA:3321707.3321806} that plays the role of one component in HCCA. In contrast, the other best layouts are relatively disordered.    

\begin{sidewaystable}
\centering
\caption{The average ranking of the proposed methods by non-parametric statistical test (Friedman test).}
\label{table:Friedman}
\scalebox{0.7}{
\begin{tabular}{l|l|l|l|l|l|l|l|l|l}
\hlineB{4} 
\multicolumn{5}{ c }{\textbf{\begin{large}4-buoy\end{large}}}&
\multicolumn{5}{ c }{\textbf{\begin{large}16-buoy\end{large}}}\\ 
\hlineB{4}
\textbf{Rank}   & \textbf{Perth}            & \textbf{Adelaide}        & \textbf{Sydney}          & \textbf{Tasmania}        & \textbf{Rank}    & \textbf{Perth}            & \textbf{Adelaide}         & \textbf{Sydney}           & \textbf{Tasmania}         \\
\hlineB{2} 
1      & CCOS (1.50)      & SLPSO (1.33)    & CCOS (1.58)     & SLPSO (1.00)    & 1       & HCCA (1.00)      & HCCA (1.00)      & HCCA (1.00)      & HCCA (1.00)      \\
\hlineB{2} 
2      & SLPSO (1.83)     & CCOS (1.66)     & SLPSO (1.83)    & SaNSDE (2.25)   & 2       & SLS-NM-B (2.16)  & SLS-NM-B (2.58)  & SLS-NM-B (2.33)  & SLS-NM-B (2.50)  \\
\hlineB{2} 
3      & HCCA (2.83)      & SaNSDE (3.41)   & HCCA (3.00)     & HCCA (2.83)     & 3       & SLS-NM (3.16)    & SLS-NM (2.66)    & AGWO (3.75)     & AGWO-NM (3.33)  \\
\hlineB{2} 
4      & SaNSDE (3.83)    & HCCA (3.58)     & SaNSDE (3.58)   & CCOS (4.33)     & 4       & AGWO-NM (4.58)  & AGWO-NM (4.66)  & SLS-NM (4.33)    & SLS-NM (3.58)    \\
\hlineB{2} 
5      & SLS-NM-B (5.42)  & SLS-NM-B (5.25) & SLS-NM-B (5.58) & SLS-NM-B (5.08) & 5       & AGWO (4.75)     & AGWO (4.75)     & AGWO-NM (4.58)  & GWO (5.25)       \\
\hlineB{2} 
6      & SLS-NM (5.58)    & SLS-NM (5.75)   & SLS-NM (5.75)   & SLS-NM (5.5)    & 6       & GWO (5.41)       & GWO (5.33)       & GWO (5.16)       & AGWO (5.41)     \\
\hlineB{2} 
7      & 1+1EA-NM (7.25)  & 1+1EA-NM (7.50) & 1+1EA-NM (6.75) & 1+1EA-NM (7.00) & 7       & CCOS (7.00)      & CCOS (7.66)      & CCOS (7.00)      & CCOS (6.91)      \\
\hlineB{2} 
8      & AGWO-NM (9.08)  & AGWO (8.75)    & CMAES-NM (8.50) & CMAES-NM (8.66) & 8       & SLPSO (8.08)     & SLPSO (8.16)     & SLPSO (8.00)     & SLPSO (8.08)     \\
\hlineB{2} 
9      & AGWO (9.33)     & AGWO-NM (9.50) & AGWO-NM (9.91) & AGWO-NM (9.00) & 9       & SaNSDE (9.08)    & SaNSDE (8.50)    & SaNSDE (8.83)    & SaNSDE (9.16)    \\
\hlineB{2} 
10     & GWO (10.33)      & GWO (10.81)     & AGWO (9.91)    & AGWO (10.58)   & 10      & LS-NM (10.16)    & LS-NM (9.91)     & LS-NM (10.58)    & LS-NM (10.50)    \\
\hlineB{2} 
11     & CMAES-NM (10.75) & CMAES-NM (9.83) & GWO (10.08)     & GWO (10.75)     & 11      & CMAES-NM (11.66) & CMAES-NM (11.25) & PSO (11.16)      & CMAES-NM (11.16) \\
\hlineB{2} 
12     & PSO (11.91)      & PSO (11.00)     & LS-NM (12.25)   & LS-NM (11.25)   & 12      & DE (12.83)       & CMA-ES (12.58)   & CMA-ES (13.16)   & PSO (11.91)      \\
\hlineB{2} 
13     & LS-NM (11.91)    & LS-NM (12.41)   & PSO (12.25)     & PSO (12.75)     & 13      & CMA-ES (13.00)   & DE (13.33)       & CMAES-NM (13.91) & 1+1EA-NM (13.66) \\
\hlineB{2} 
14     & DE (14.25)       & DE (14.08)      & DE (14.16)      & DE (14.25)      & 14      & 1+1EA-NM (13.41) & PSO (13.58)      & 1+1EA-NM (14.00) & DE (13.66)       \\
\hlineB{2} 
15     & CMA-ES (14.91)   & CMA-ES (14.91)  & NM (15.33)      & CMA-ES (14.91)  & 15      & PSO (13.75)      & 1+1EA-NM (14.00) & DE (14.08)       & CMA-ES (13.83)   \\
\hlineB{2} 
16     & NM (15.25)       & NM (16.00)      & CMA-ES (15.50)  & NM (15.83)      & 16      & NM (15.91)       & NM (16.00)       & NM (14.08)       & NM (16.00)      \\
\hlineB{5} 

\end{tabular}
}
\end{sidewaystable}

\begin{sidewaystable}
\centering

\scalebox{0.7}{

\begin{tabular}{l|l|l|l|l|l|l|l|l|l|l|l|l|l|l|l|l}

\hlineB{5}
\multicolumn{16}{ c }{\textbf{\begin{large}Perth wave scenario (16-buoy)\end{large}}}\\ \\
\hlineB{4}
& \textbf{DE}&
\textbf{CMA-ES}&
\textbf{NM}&
\textbf{GWO}& 
  \textbf{AGWO}&
\textbf{PSO}&
\textbf{CMAES-NM} &
\textbf{1+1EA-NM} & 
\textbf{AGWO-NM}&
\textbf{CCOS}&
\textbf{SLPSO$_{II}$}&
\textbf{SaNSDE$_{II}$}&
\textbf{LS-NM }&
\textbf{SLS-NM}&
\textbf{SLS-NM-B}&
\textbf{HCCA} \\

                  \hlineB{4}
\texttt{\textbf{Max}} & 1978004 & 1972042 & 1707140 & 3869189 & 4017436 & 1874697  & 2205390  & 1985315 &  4071870    & 2926335 & 2666886 & 2460983 & 2316240     &4613064  & 5355093 & \textbf{8884930}    \\

\texttt{\textbf{Min}} & 1813302 & 1764271 & 1393953& 2631383 &2940571 & 1798449 & 1527216 & 1508531 & 3048410  & 2395674 &2294175 & 2092220& 1887327 & 3530473 & 3949997 & 8000897    \\

\texttt{\textbf{Mean}} & 1876041  & 1855243 & 1585395 & 3258014&3517185 & 1830755 & 1956295 & 1830649 & 3608593 & 2683725&2507576 & 2279899&2106215 &4010725  & 4497538 & \textbf{8561839}   \\

\texttt{\textbf{Median}}& 1876824& 1847380 & 1578621 & 3218468& 3664482 & 1830980 & 1981432 & 1889295 & 3693526 & 2706138 &2528358 & 2258676& 2078183 & 4010689 &   4413129& \textbf{8571208}    \\

\texttt{\textbf{STD}} & 46761 & 57806 & 91733 & 428449 &322679  & 21004 & 172404& 81722 &  312569 & 145881& 115616 & 106092 & 146586 & 320417 &  481974 & 229971    \\
\hlineB{4}
\multicolumn{16}{ c }{\textbf{\begin{large}Sydney wave scenario (16-buoy)\end{large}}}\\ \\
\hlineB{4}
                 & \textbf{DE}& 
                 \textbf{CMA-ES}&
                 \textbf{NM}& 
                 \textbf{GWO}&
                   \textbf{AGWO}&
                 \textbf{PSO}&
                 \textbf{CMAES-NM} &
                 \textbf{1+1EA-NM} &
                 \textbf{AGWO-NM}&
                 \textbf{CCOS}&
                 \textbf{SLPSO$_{II}$}&
                   \textbf{SaNSDE$_{II}$}&
                 \textbf{LS-NM }&
                 \textbf{SLS-NM}&
                 \textbf{SLS-NM-B}&
                 \textbf{HCCA} \\
                  \hlineB{4}

\texttt{\textbf{Max}} & 1288869 & 1290891 & 1285570 & 2160341&2357618 & 1345119& 1359025 & 1334937 & 2427610 & 1808910& 1617776&1573388 & 1476492 & 2266424 & 2441157 & \textbf{4170868}   \\

\texttt{\textbf{Min}} & 1218526 & 1222920 & 1210160 & 1643183& 1942160 & 1290986 & 1206080 & 1175279& 1938793 & 1578082 &1500364 & 1436004& 1228500  & 1779821  &1983922  & 4011595   \\

\texttt{\textbf{Mean}}    & 1250086 & 1261968 & 124989 & 1961163& 2116037 & 1320898 & 1259204 & 1254215 &2101718  & 1719665&1552703 &1499720 & 1374478  &2089419  & 2345455 & \textbf{4075718}    \\

\texttt{\textbf{Median}}& 1249281 & 1261820 & 1249830 & 1965235&2141568  & 1326489 & 1234540 & 1244844 &2116880  & 1743946 &1538400 &1487439 & 1389960 &2136654  & 2381567 & \textbf{4079520}   \\

\texttt{\textbf{STD}}     & 18793 & 18397 & 22330 & 149244&110808  &14393 & 53006 & 52187 & 150286 & 78724 & 42453 &45235 & 69691 & 158182 & 118228 & 61356   \\

\hlineB{4}
\multicolumn{16}{ c }{\textbf{\begin{large}Adelaide wave scenario (16-buoy)\end{large}}}\\ \\
\hlineB{4}
                  & \textbf{DE}&
                  \textbf{CMA-ES}&
                  \textbf{NM}&
                  \textbf{GWO}&
                    \textbf{AGWO}&
                  \textbf{PSO}&
                  \textbf{CMAES-NM} &
                  \textbf{1+1EA-NM} &
                  \textbf{AGWO-NM}&
                  \textbf{CCOS}& 
                 \textbf{SLPSO$_{II}$}&
                   \textbf{SaNSDE$_{II}$}&
                  \textbf{LS-NM }&
                  \textbf{SLS-NM}&
                  \textbf{SLS-NM-B}& 
                  \textbf{HCCA} \\
                  \hlineB{4}

\texttt{\textbf{Max}} & 2073818 & 2066628 & 1841800 & 4055518& 4198980 & 2013721 & 2182902 & 1996107 & 4331090 & 2890630&2724026 &2600827 & 2353122 & 5389570 & 6305950 & \textbf{9470521}   \\

\texttt{\textbf{Min}}  & 1847975  & 1925787 & 1442820 & 2615454& 2828922 & 1852867 & 1918926 & 1782396 & 3129079 & 2027935& 2324849 &2177553 & 1998174&3886630  & 3731321 & 8482713    \\

\texttt{\textbf{Mean}}  & 1942373 & 1980279 & 1629929 & 3386471& 3475527 & 1921268 & 2076939 & 1884884 & 3588090 & 2629506& 2521642 &2417450 & 2203759 & 4424488 & 4400128 & \textbf{9001224}   \\

\texttt{\textbf{Median}}  & 1947228 & 1976812 & 1637145 & 3397133& 3385324  & 1915633 & 2082066 & 1863392 & 3521503 & 2725506& 2512711&2476026 & 2225386&4271271  &4233962  & \textbf{9124786}    \\

\texttt{\textbf{STD}}     & 57795 & 40254 & 146374 & 425179& 393443 &42856 & 96119 & 71047 & 379587  & 282542&158706 &130958 & 102117 & 501941 &696882  & 343378   \\

\hlineB{4}
\multicolumn{16}{ c }{\textbf{\begin{large}Tasmania wave scenario (16-buoy)\end{large}}}\\ \\
\hlineB{4}
                  & \textbf{DE}& 
                  \textbf{CMA-ES}& 
                  \textbf{NM}&
                  \textbf{GWO}&
                  \textbf{AGWO}&
                  \textbf{PSO}&
                  \textbf{CMAES-NM} & 
                  \textbf{1+1EA-NM} & 
                  \textbf{AGWO-NM}& 
                  \textbf{CCOS}&
                  \textbf{SLPSO$_{II}$}&
                   \textbf{SaNSDE$_{II}$}&
                  \textbf{LS-NM }&
                  \textbf{SLS-NM}&
                  \textbf{SLS-NM-B}&
                  \textbf{HCCA} \\
                  \hlineB{4}
\texttt{\textbf{Max}} & 4102931 & 4010201 & 3636190     & 6752582& 7117766 & 4202333 & 4555196 & 4130826   & 8143490 & 5852075&5351325 &4874586  & 4807895 &8891329   & 11771018 & \textbf{16357582}   \\

\texttt{\textbf{Min}}     & 3815606 & 3505586 & 3165160 & 5297795&5982637   & 4031614 & 3966642  & 3713873 & 7093660 & 5054946& 4615768 & 4331868& 3842791 &  5830263& 7054758  & 15503720   \\

\texttt{\textbf{Mean}}    & 3889232 & 3870418 & 3388250 & 6344121& 6364880 & 4105837  & 4237851 & 3915490 & 7456337 & 5477239& 4970179 &4552102 & 4332727 &  7549235& 8606078 & \textbf{15952780}   \\

\texttt{\textbf{Median}}  & 3860710  & 3904263 & 3393264 & 6411575&6249592  & 4097199 & 4230567 & 3896565 & 7374770 & 5499668&4978008 & 4532614 & 4295434 &  7733837& 8458468 & \textbf{16070584}    \\

\texttt{\textbf{STD}}     & 76482 & 136384 & 146906  & 414505&362093  & 48965 & 178946  & 141545 & 287090 & 232319& 199092&173880 & 284993 & 964014  &1279960  & 341515 \\
\hlineB{5}
\end{tabular}
}
\caption{Performance comparison of various heuristics for the 16-buoy case, based on maximum, median and mean power output layout of the best solution per experiment.}
\label{table:allresults16}
\end{sidewaystable}

\begin{sidewaystable}
\centering
\scalebox{0.7}{
\begin{tabular}{l|l|l|l|l|l|l|l|l|l|l|l|l|l|l|l|l}
\hlineB{5}
\multicolumn{16}{ c }{\textbf{\begin{large}Perth wave scenario (4-buoy)\end{large}}}\\ \\
\hlineB{4}
 & \textbf{DE}&
 \textbf{CMA-ES}&
 \textbf{NM}&
 \textbf{GWO}&
 \textbf{AGWO}&
 \textbf{PSO}&
 \textbf{CMAES-NM} &
 \textbf{1+1EA-NM} &
 \textbf{AGWO-NM}&
 \textbf{CCOS}&
 \textbf{SLPSO$_{II}$}&
\textbf{SaNSDE$_{II}$}&
 \textbf{LS-NM }&
 \textbf{SLS-NM}& 
 \textbf{SLS-NM-B}& 
 \textbf{HCCA} \\ 
   \hlineB{4}

\texttt{\textbf{Max}} & 757152 & 747703 & 937666& 1203583& 1301909 &1096161 & 1181345 & 1547929 & 1490604 & \textbf{2589217}  & 2585665 & 2496828 &1056778 & 1955382  & 2081568 & 2571781    \\

\texttt{\textbf{Min}} & 679514 & 684384 & 522987 & 862094& 980777 &898365 & 664390 & 1213837 & 952285 & 2544542& 2505748&2340314 & 748581 & 1498860 & 1513747 & 2369251    \\

\texttt{\textbf{Mean}} & 728521 & 710025 & 659399    & 1055699& 1165472  & 993932 & 1021743  & 1376987 & 1207704 & \textbf{2568664} & 2556791 & 2410511 &954236  &  1701684 &  1780217 & 2462307    \\

\texttt{\textbf{Median}} & 729269 & 709265  & 612187& 1089776&1166276 & 986457  & 1092366 & 1369274&1196508 & \textbf{2572645} & 2563001 & 2408779&970925 & 1677414& 1779828& 2460435    \\

\texttt{\textbf{STD}} & 23057 & 18345 & 123361 & 108562 & 97251 & 66499 & 186353 & 81722 &  166147 & 15795 &25069 &45826 &86612  & 116545 & 164896 &  62600   \\

\hlineB{4}

\multicolumn{16}{ c }{\textbf{\begin{large}Sydney wave scenario (4-buoy)\end{large}}}\\ \\

\hlineB{4}

& \textbf{DE}& 
\textbf{CMA-ES}& 
\textbf{NM}&
\textbf{GWO}&
\textbf{AGWO}&
\textbf{PSO}&
\textbf{CMAES-NM} &
\textbf{1+1EA-NM} &
\textbf{AGWO-NM}&
\textbf{CCOS}&
\textbf{SLPSO$_{II}$}&
\textbf{SaNSDE$_{II}$}&
\textbf{LS-NM }&
\textbf{SLS-NM}&
\textbf{SLS-NM-B}&
\textbf{HCCA} \\ 

\hlineB{4}

\texttt{\textbf{Max}} & 473972 & 434564 & 462284& 760479& 784675  &696357& 802578& 928381&787006 & 1268449 & \textbf{1272316} &1252661 & 672968 &995424 & 1017464 & 1262763    \\

\texttt{\textbf{Min}} & 434121 & 406658& 352144& 616066& 645036   &542108  & 692273& 768977& 620102 & 1251253&1246321 &1182527 & 493190 &  785335& 847721  & 1198794   \\

\texttt{\textbf{Mean}}  & 450864 & 419080 & 404325 & 686588&  689951  &612739 & 730352 & 835518 & 681253 & \textbf{1260117} & 1259924 & 1205699 & 595972 & 910660 & 933515  & 1228633   \\

\texttt{\textbf{Median}} & 453002 & 416255 & 409289 & 683622& 681652 &601032 & 726847 & 819626 & 679740 & \textbf{1261359} & 1260302 &1200709 & 604652 & 933633 &911955  & 1225774   \\

\texttt{\textbf{STD}} & 12739 & 8043 & 34621 & 44746& 41462   & 45143 & 29601 & 49786 & 44509 & 5192 &7322 & 18938 & 61123 &  62389 & 58217 & 27489   \\

\hlineB{4}

\multicolumn{16}{ c }{\textbf{\begin{large}Adelaide wave scenario (4-buoy)\end{large}}}\\ \\

\hlineB{4}

& \textbf{DE}&
\textbf{CMA-ES}&
\textbf{NM}&
\textbf{GWO}&
\textbf{AGWO}&
\textbf{PSO}&
\textbf{CMAES-NM} &
\textbf{1+1EA-NM} &
\textbf{AGWO-NM}&
\textbf{CCOS}&
\textbf{SLPSO$_{II}$}&
\textbf{SaNSDE$_{II}$}&
\textbf{LS-NM }&
\textbf{SLS-NM}& 
\textbf{SLS-NM-B}&
\textbf{HCCA} \\

 \hlineB{4}

\texttt{\textbf{Max}} & 828677 & 764148 & 722935 & 1367177& 1383420  & 1217127& 1147835& 1497037&1439554 & 2720283 & \textbf{2732298} & 2679951& 1161869 & 2036738 & 2205885& 2711811    \\

\texttt{\textbf{Min}} & 706780& 730629& 520054 & 984928& 1150197 &986746 & 1058355& 1215677  & 978806& 2652191& 2657184 & 2514665 &903350 &1530568  & 1576432& 2303003    \\

\texttt{\textbf{Mean}} & 784906 & 749347 & 629736 & 1119675& 1218747   &1088023 & 1147835 & 1403663  & 1211560 & 2692888 & \textbf{2699379} & 2603358&1001791  & 1724642 &  1818722& 2541981   \\

\texttt{\textbf{Median}} & 786566 & 751533 & 635767 & 1103790&  1205228 & 1098536 & 1151581& 1436140 & 1193950 & \textbf{2700573} & 2699965 & 2617516 & 970319 &  1714104& 1844716 & 2574643    \\

\texttt{\textbf{STD}}  & 33305 & 11744   & 78249 & 123038 & 73511  & 72345 & 55729 & 80538 & 157434 & 21741& 23531 &56901 & 82646 & 156221 & 178678 & 124799    \\

\hlineB{4}

\multicolumn{16}{ c }{\textbf{\begin{large}Tasmania wave scenario (4-buoy)\end{large}}}\\ \\

\hlineB{4}

& \textbf{DE}&
\textbf{CMA-ES}&
\textbf{NM}& 
\textbf{GWO}&
\textbf{AGWO}&
\textbf{PSO}&
\textbf{CMAES-NM} &
\textbf{1+1EA-NM} &
\textbf{AGWO-NM}&
\textbf{CCOS}&
\textbf{SLPSO$_{II}$}&
\textbf{SaNSDE$_{II}$}&
\textbf{LS-NM }&
\textbf{SLS-NM}& 
\textbf{SLS-NM-B}&
\textbf{HCCA} \\

                  \hlineB{4}
 \texttt{\textbf{Max}} & 1520332 & 1514709 & 1470010& 2274781& 2384846  &2069514 & 2369299& 3221716 &2533510 & 3775336 & \textbf{4808902} & 4596171 & 2454579& 3842230 &  3991016 & 4691217   \\

\texttt{\textbf{Min}} & 1411485& 1383235 & 1125670 & 1963594& 2044584  & 1564277 & 2122243 & 2612539 & 2098842 & 3312687&4743330 &4511478 & 1875692& 3028176& 3104568& 4314911    \\

\texttt{\textbf{Mean}} & 1472476& 1426742 & 1272643 & 2106575& 2173009  & 1834935 & 2369299 & 2863883 & 2310503 & 3775336 & \textbf{4786660} &4557535 & 2094040 &3427671  &  3494835 & 4512121    \\

\texttt{\textbf{Median}} & 1473481  & 1430254 & 1236877  & 2098971& 2150044 & 1838107 & 2398747 & 2864481 & 2362061 & 3836696& \textbf{4792621}& 4563350& 2040533 & 3357045 & 3450918 & 4507503    \\

\texttt{\textbf{STD}} & 46831 & 37946 & 104560 & 111927& 89905 & 162102 & 100318 & 185941 & 147406 & 351782 & 20583 &33529 & 185895 & 271382 & 299053 & 95361  \\

\hlineB{5}
\end{tabular}
}
\caption{Performance comparison of various heuristics for the 4-buoy case, based on maximum, median and mean power output layout of the best solution per experiment.}
\label{table:allresults4}
\end{sidewaystable}

\section{Conclusions}\label{sec:conc}
Optimising a combination of positions and Power Take-off parameters of a wave farm with a large generator number creates a computationally expensive, multi-modal, large-scale and complicated optimisation problem. These challenges are the foremost motivation for discovering faster and smarter optimisation techniques. In this article, we propose a new hybrid cooperative co-evolution method (HCCA) which is composed of a fast strategy for optimising the WEC positions and an effective cooperative strategy (three optimisers) for tuning the PTOs configurations in four real wave scenarios. Moreover, we propose a new adaptive mechanism for improving  GWO and the idea is evaluated using ten variants of chaotic maps. To systematically compare the performance of the new search frameworks, we discuss and apply 15 state-of-the-art evolutionary, swarm, alternating (cooperative) and hybrid optimisation algorithms. According to the experimental results, HCCA is able to out-perform other heuristics search methods in terms of convergence speed (5 times faster than the best previous algorithm) and the quality of layouts ($80\%$ improvement of the sustained energy output in 16-buoy experiments). In the future, we would like to combine HCCA with a deep neuro-surrogate model which is trained using a minimum number of samples for speeding up the evaluation time of the large wave farm. The neuro-surrogate will aim to estimate the total power output based on the wave scenario's characteristics.

\begin{algorithm}
\small
\caption{$\mathit{Backtracking\, optimisation\, Algorithm\, (BOA)}$}\label{alg:B}
\begin{algorithmic}[1]
\Procedure{BOA ($\mathbb{S},\mathbb{A}$ )}{}\\
 \textbf{Initialization}
  
\\ $\mathit{energy}=([E_{1},E_{2},\ldots,E_{N}])=\mathit{Eval(\mathbb{S})}$ \Comment{Evaluate layout}
\\ $N_w=\mathit{round}(N/4)$ \Comment{Buoy number need to be improved}
\\$\langle\mathit{WIndex}\rangle$={\em{FindWorst}}$(\mathit{energy},N_w)$ 
\Comment{Find worst buoys power}
 \For{ $i$ in $[1,..,N_w]$ }
       \State \textbf{\em{Position optimisation}}
  \State $(\mathbb{S}_{\mathit{WIndex}(i)}^{\mathit{Position}},\mathit{energy}_{\mathit{WIndex}(i)})$={\em{Nelder-Mead}}$(\mathit{\mathbb{S}_{\mathit{WIndex}(i)}^{\mathit{Position}}})$ 
     \EndFor 
     
      \State \textbf{\em{PTO global optimisation}}
  \State $(\mathbb{S}_\mathit{PTOs},\mathit{energy})$=\em{optimise}$(\mathbb{S}_\mathit{PTOs} ,\mathbb{A}_\mathit{BestIndex})$ 
  
\State \textbf{return} $\mathbb{S},\mathit{energy}$  \Comment{Final Layout}
\EndProcedure
\end{algorithmic}
\end{algorithm}


\bibliographystyle{unsrt}  
\bibliography{sample-bibliography}
\end{document}